\documentclass{article}

\usepackage[final]{neurips_2022}

\usepackage[utf8]{inputenc} 
\usepackage[T1]{fontenc}    
\usepackage[hidelinks]{hyperref}       
\usepackage{url}            
\usepackage{booktabs}       
\usepackage{amsfonts}       
\usepackage{nicefrac}       
\usepackage{microtype}      
\usepackage{xcolor}         

\usepackage{microtype}
\usepackage{graphicx}
\usepackage{booktabs} 
\usepackage{caption}
\usepackage{subcaption}
\usepackage{mwe}
\usepackage{algorithm}
\usepackage{algorithmicx,algpseudocode}
\usepackage{ulem}
\usepackage{xcolor}
\usepackage{amsthm}
\usepackage{xr}
\usepackage{amssymb}
\usepackage{mathtools}
\usepackage{thmtools}
\usepackage{thm-restate}
\usepackage{wrapfig}

\title{Faster Deep Reinforcement Learning with \\ Slower Online Network}

\author{%
  Kavosh Asadi\\
  Amazon Web Services\\
 \And
  Rasool Fakoor \\
  Amazon Web Services \\
 \And
  Omer Gottesman \\
  Brown University \\
 \And
  Taesup Kim \\
  Seoul National University \\
 \And
  Michael L. Littman \\
  Brown University \\
  \And
  Alexander J. Smola \\
  Amazon Web Services \\
}

\begin{document}

\maketitle

\newcommand{\prox}[2]{\textrm{prox}_{#1}(#2)}
\newcommand{\proxNoParanthesis}[1]{\textrm{prox}_{#1}}
\newcommand{\norm}[3]{\left\lVert#1\right\rVert_{#2}^{#3}}
\newcommand{\empiricalExpectation}[1]{\widehat{\text E}_{#1}}
\newcommand{\expectation}[1]{\textup E_{#1}}
\newcommand{\dotProduct}[2]{\langle #1,#2 \rangle}
\newcommand{\variance}[0]{\textbf{Var}}

\newcommand{\algname}{DQNPro}
\newcommand{\algfullname}{Deep Q-Network with Proximal Iteration}
\newcommand{\E}{\mathbb{E}}
\newcommand\myeq{\vcentcolon=}
\begin{abstract}
Deep reinforcement learning algorithms often use two networks for value function optimization: an online network, and a target network that tracks the online network with some delay. Using two separate networks enables the agent to hedge against issues that arise when performing bootstrapping. In this paper we endow two popular deep reinforcement learning algorithms, namely DQN and Rainbow, with updates that incentivize the online network to remain in the proximity of the target network. This improves the robustness of deep reinforcement learning in presence of noisy updates. The resultant agents, called DQN Pro and Rainbow Pro, exhibit significant performance improvements over their original counterparts on the Atari benchmark demonstrating the effectiveness of this simple idea in deep reinforcement learning. The code for our paper is available here: Github.com/amazon-research/fast-rl-with-slow-updates.
\end{abstract}
\section{Introduction}
An important competency of reinforcement-learning (RL) agents is learning in environments with large state spaces like those found in robotics~\citep{kober_robotics}, dialog systems~\citep{williams2017hybrid}, and games~\citep{td_gammon, silver2017mastering}. Recent breakthroughs in deep RL have demonstrated that simple approaches such as Q-learning~\citep{q_learning} can surpass human-level performance in challenging environments when equipped with deep neural networks for function approximation~\citep{mnih2015human}.

Two components of a gradient-based deep RL agent are its objective function and optimization procedure. The optimization procedure takes estimates of the gradient of the objective with respect to network parameters and updates the parameters accordingly. In DQN~\citep{mnih2015human}, for example, the objective function is the empirical expectation of the temporal difference (TD) error~\citep{sutton_td} on a buffered set of environmental interactions~\citep{lin1992self}, and variants of stochastic gradient descent are employed to best minimize this objective function.

A fundamental difficulty in this context stems from the use of bootstrapping. Here, bootstrapping refers to the dependence of the target of updates on the parameters of the neural network, which is itself continuously updated during training. Employing bootstrapping in RL stands in contrast to supervised-learning techniques and Monte-Carlo RL~\citep{rl_book}, where the target of our gradient updates does not depend on the parameters of the neural network. 

\citet{mnih2015human} proposed a simple approach to hedging against issues that arise when using bootstrapping, namely to use a \textit{target network} in value-function optimization. The target network is updated periodically, and tracks the online network with some delay. While this modification constituted a major step towards combating misbehavior in Q-learning~\citep{target_based_TD, kim2019deepmellow, zhang_breaking_deadly_triad}, optimization instability is still prevalent~\citep{van_hasselt_deadly_triad}.

Our primary contribution is to endow DQN and Rainbow~\citep{rainbow} with a term that ensures the parameters of the online-network component remain in the proximity of the parameters of the target network. Our theoretical and empirical results show that our simple proximal updates can remarkably increase robustness to noise without incurring additional computational or memory costs. In particular, we present comprehensive experiments on the Atari benchmark~\citep{bellemare_atari} where proximal updates yield significant improvements, thus revealing the benefits of using this simple technique for deep RL.
\section{Background and Notation}
RL is the study of the interaction between an environment and an agent that learns to maximize reward through experience. The Markov Decision Process~\citep{PutermanMDP1994}, or MDP, is used to mathematically define the RL problem. An MDP is specified by the tuple $\langle \mathcal{S,A,R,P,\gamma}
\rangle$, where $\mathcal{S}$ is the set of states and
$\mathcal{A}$ is the set of actions. The functions $\mathcal{R:S\times A\rightarrow}\ \mathbb R$ and
$\mathcal{P:S\times A \times S\rightarrow}\ [0,1]$ denote the reward and transition dynamics of the
MDP. Finally, a discounting factor $\gamma$ is used to formalize the intuition that short-term rewards are more valuable than those received later. 

The goal in the RL problem is to learn a policy, a mapping from states to a probability distribution over actions, ${\pi:\mathcal{S}\rightarrow \mathcal{P}(\mathcal{A})}$, that obtains high sums of future discounted rewards. An important concept in RL is the state value
function. Formally, it denotes the expected discounted sum of future rewards when committing to a policy $\pi$ in a state $s$: $v^{\pi}(s)\myeq \E\big[\sum_{t=0}^{\infty}\gamma^{t} R_t\big |S_0=s, \pi \big]\ .$
We define the Bellman operator $\mathcal{T^{\pi}}$ as follows:
\begin{equation*}
\label{eq:Bellman}
\big[\mathcal{T^{\pi}}v\big](s):=\!\sum_{a\in\mathcal{A}}\pi(a\mid s)\big(\mathcal{R}(s,a)+\!\sum_{s'\in \mathcal{S}}\!\gamma\ \mathcal{P}(s,a,s') v(s')\big)\ ,
\end{equation*}
which we can write compactly as: $\mathcal{T}^{\pi}v \myeq R^{\pi} + \gamma P^{\pi} v\ ,$ 
where $\big[R^{\pi}\big](s)= \sum_{a\in\mathcal{A}}\pi(a| s)\mathcal{R}(s,a)$ and $\big[P^{\pi} v\big](s)= \sum_{a\in\mathcal{A}}\pi(a\mid s)\sum_{s'\in \mathcal{S}}\!\mathcal{P}(s,a,s') v(s')$. We also denote:
 $(\mathcal{T}^{\pi})^{n}v\myeq\underbrace{\mathcal{T}^{\pi}\cdots\mathcal{T}^{\pi}}_{n\ \textrm{compositions}}v \ .$ Notice that $v^{\pi}$ is the unique fixed-point of $(\mathcal{T}^{\pi})^n$ for all natural numbers $n$, meaning that $ v^{\pi}=(\mathcal{T}^{\pi})^n v^{\pi}\ $, for all $n$. Define $v^{\star}$ as the optimal value of a state, namely: ${v^{\star}(s)\myeq \max_{\pi} v^{\pi}(s)} $, and $\pi^{\star}$ as a policy that achieves $v^{\star}(s)$ for all states. We define the Bellman Optimality Operator $\mathcal{T^{\star}}$:
\begin{equation*}
\label{eq:Bellman_Optimality}
\big[\mathcal{T^{\star}}v\big](s):=\!\max_{a\in\mathcal{A}}\mathcal{R}(s,a)+\!\sum_{s'\in \mathcal{S}}\!\gamma\ \mathcal{P}(s,a,s') v(s')\ ,
\end{equation*}
whose fixed point is $v^{\star}$. These operators are at the heart of many planning and RL algorithms including Value Iteration~\citep{bellamn1957} and Policy Iteration~\citep{howard1960dynamic}.
\section{Proximal Bellman Operator}
In this section, we introduce a new class of Bellman operators that ensure that the next iterate in planning and RL remain in the vicinity of the previous iterate. To this end, we define the Bregman Divergence generated by a convex function $f$:
$$
D_{f}(v',v) \myeq f(v') - f(v) - \langle \nabla f(v), v'-v \rangle\ .
$$
Examples include the $l_{p}$ norm generated by $f(v)=\frac{1}{2}\norm{v}{p}{2}$ and the Mahalanobis Distance generated by $f(v)=\frac{1}{2}\langle v,Q v \rangle$ for a positive semi-definite matrix $Q$.

We now define the Proximal Bellman Operator $(\mathcal{T}^{\pi}_{c, f})^n$:
\begin{equation}
    (\mathcal{T}^{\pi}_{c, f})^n v \myeq\arg\min_{v'} ||v'- (\mathcal{T}^{\pi})^n v||^{2}_{2} +\frac{1}{c} D_{f}(v',v)\ ,
    \label{eq:proximal_bellman_operator}
\end{equation}

where $c\in(0,\infty)$. Intuitively, this operator encourages the next iterate to be in the proximity of the previous iterate, while also having a small difference relative to the point recommended by the original Bellman Operator. The parameter $c$ could, therefore, be thought of as a knob that controls the degree of gravitation towards the previous iterate.

Our goal is to understand the behavior of Proximal Bellman Operator when used in conjunction with the Modified Policy Iteration (MPI) algorithm~\citep{PutermanMDP1994, ampi}. Define $\mathcal{G}v$ as the greedy policy with respect to $v$.
At a certain iteration $k$, Proximal Modified Policy Iteration (PMPI) proceeds as follows:
\begin{eqnarray}
\pi_{k} &\leftarrow& \mathcal{G} v_{k-1}\ \label{eq:pmpi_1} ,\\
v_{k} &\leftarrow& (\mathcal{T}^{\pi_{k}}_{c, f})^{n} v_{k-1} \label{eq:pmpi_2}\ .
\end{eqnarray}
The pair of updates above generalize existing algorithms. Notably, with $c\rightarrow\infty$ and general $n$ we get MPI, with $c\rightarrow\infty$ and $n=1$ the algorithm reduces to Value Iteration, and with $c\rightarrow\infty$ and $n=\infty$ we have a reduction to Policy Iteration. For finite $c$, the two extremes of $n$, namely $n=1$ and $n=\infty$, could be thought of as the proximal versions of Value Iteration and Policy Iteration, respectively.

To analyze this approach,
it is first natural to ask if each iteration of PMPI could be thought of as a contraction so we can get sound and convergent behavior in planning and learning. For $n> 1$, \citet{ampi} constructed a contrived MDP demonstrating that one iteration of MPI can unfortunately expand. As PMPI is just a generalization of MPI, the same example from \citet{ampi} shows that PMPI can expand. In the case of $n=1$, we can rewrite the pair of equations (\ref{eq:pmpi_1}) and (\ref{eq:pmpi_2}) in a single update as follows:
$v_k \leftarrow \mathcal{T}^{\star}_{c, f} v_{k-1}\ .$
When $c\rightarrow\infty$, standard proofs can be employed to show that the operator is a contraction~\citep{littman_csaba}.
We now show that $\mathcal{T}^{\star}_{c, f}$ is a contraction for finite values of $c$. See our appendix for proofs.
\begin{restatable}[]{thm}{contraction}
\label{thm:goldbach}
The Proximal Bellman Optimality Operator $\mathcal{T}^{\star}_{c, f}$ is a contraction with fixed point $v^{\star}$.
\end{restatable}
Therefore, we get convergent behavior when using $\mathcal{T}^{\star}_{c, f}$ in planning and RL. The addition of the proximal term is fortunately not changing the fixed point, thus not negatively affecting the final solution.
This could be thought of as a form of regularization that vanishes in the limit; the algorithm converges to $v^{\star}$ even without decaying $1/c$.

Going back to the general $n\geq 1$ case, we cannot show contraction, but following previous work~\citep{bertsekas1996neuro, ampi}, we study error propagation in PMPI in presence of additive noise where we get a noisy sample of the original Bellman Operator $(\mathcal{T}^{\pi_k})^n v_{k-1}+\epsilon_{k}$. The noise can stem from a variety of reasons, such as approximation or estimation error. For simplicity, we restrict the analysis to $D_f(v',v)= ||v'-v||^{2}_{2}$, so we rewrite update (\ref{eq:pmpi_2}) as:
$$v_{k}\leftarrow \arg\min_{v'} ||v'- \big((\mathcal{T}^{\pi_k})^n v_{k-1}+\epsilon_{k}\big)||^{2}_{2} +\frac{1}{c} \norm{v'-v_{k-1}}{2}{2}$$
which can further be simplified to:
$$v_{k}\leftarrow \underbrace{(1-\beta)(\mathcal T^{\pi_k})^{n}v_{k-1}+ \beta v_{k-1}}_{:=(\mathcal{T}^{\pi_k}_{\beta})^n v_{k-1}} +(1-\beta)\epsilon_{k} , $$
where $\beta=\frac{1}{1+c}$. This operator is a generalization of the operator proposed by~\citet{smirnova2020convergence} who focused on the case of $n=1$. To build some intuition, notice that the update is multiplying error $\epsilon_k$ by a term that is smaller than one, thus better hedging against large noise. While the update may slow progress when there is no noise, it is entirely conceivable that for large enough values of $\epsilon_k$, it is better to use non-zero $\beta$ values. In the following theorem we formalize this intuition. Our result leans on the theory provided by \citet{ampi} and could be thought of as a generalization of their theorem for non-zero $\beta$ values.

\begin{restatable}[]{thm}{error}
\label{thm:loss_bound}
Consider the PMPI algorithm specified by:
\begin{eqnarray}
\pi_{k} &\leftarrow& \mathcal{G}_{\epsilon'_k} v_{k-1}\ ,\\
v_{k}&\leftarrow& (\mathcal{T}^{\pi_k}_{\beta})^n v_{k-1} +(1-\beta)\epsilon_k\ .
\end{eqnarray}

Define the Bellman residual $b_k\myeq v_{k} - \mathcal{T}^{\pi_{k+1}}v_k$, and error terms $x_k := (I-\gamma P^{\pi_{k}})\epsilon_k$ and $y_k:=\gamma P^{\pi^*}\epsilon_{k}$.
After $k$ steps:
$$ v^{*} - v^{\pi_k} = \underbrace{v^{\pi^{*}}-(\mathcal{T}^{\pi_{k+1}}_{\beta})^n v_k}_{d_k} +\underbrace{(\mathcal{T}^{\pi_{k+1}}_{\beta})^n v_k - v_{\pi_k}}_{s_k} $$
\begin{itemize}
    \item where $d_{k} \leq \gamma P^{\pi^{*}}d_{k-1} -\big((1-\beta)y_{k-1} + \beta b_{k-1}\big) + (1-\beta)\sum_{j=1}^{n-1}(\gamma P^{\pi_{k}})^j b_{k-1} + \epsilon'_{k}$
    \item $s_k \leq \big((1-\beta)(\gamma P^{\pi_k})^n + \beta I\big)(I-\gamma P^{\pi_k})^{-1}b_{k-1}$
    \item $b_{k} \leq \big((1-\beta)(\gamma P^{\pi_k})^{n} + \beta I\big) b_{k-1}+ (1-\beta) x_{k} + \epsilon'_{k+1}$
\end{itemize}
\end{restatable}

The bound provides intuition as to how the proximal Bellman Operator can accelerate convergence in the presence of high noise. For simplicity, we will only analyze the effect of the $\epsilon$ noise term, and ignore the $\epsilon'$ term. We first look at the Bellman residual, $b_k$. Given the Bellman residual in iteration $k-1$, $b_{k-1}$, the only influence of the noise term $\epsilon_k$ on $b_k$ is through the $(1-\beta) x_k$, term, and we see that $b_k$ decreases linearly with larger $\beta$.

The analysis of $s_k$ is slightly more involved but follows similar logic. The bound for $s_k$ can be decomposed into a term proportional to $b_{k-1}$ and a term proportional to $\beta b_{k-1}$, where both are multiplied with positive semi-definite matrices. Since $b_{k-1}$ itself linearly decreases with $\beta$, we conclude that larger $\beta$ decreases the bound quadratically.

The effect of $\beta$ on the bound for $d_k$ is more complex. The terms $\beta y_{k-1}$ and $\sum_{j=1}^{n-1}(\gamma P^{\pi_{k}})^j b_{k-1}$ introduce a linear decrease of the bound on $d_k$ with $\beta$, while the term $\beta (I - \sum_{j=1}^{n-1}(\gamma P^{\pi_{k}})^j) b_{k-1}$ introduces a quadratic dependence whose curvature depends on $I - \sum_{j=1}^{n-1}(\gamma P_{\pi_{k}})^j$. This complex dependence on $\beta$ highlights the trade-off between noise reduction and magnitude of updates. To understand this trade-off better, we examine two extreme cases for the magnitude on the noise. When the noise is very large, we may set $\beta=1$, equivalent to an infinitely strong proximal term. It is easy to see that for $\beta=1$, the values of $d_k$ and $s_k$ remain unchanged, which is preferable to the increase they would suffer in the presence of very large noise. On the other extreme, when no noise is present, the $x_k$ and $y_k$ terms in Theorem~\ref{thm:loss_bound} vanish, and the bounds on $d_k$ and $s_k$ can be minimized by setting $\beta=0$, i.e. without noise the proximal term should not be used and the original Bellman update performed. Intermediate noise magnitudes thus require a value of $\beta$ that balances the noise reduction and update size.
\section{Deep Q-Network with Proximal Updates}
We now endow DQN-style algorithms with proximal updates. Let $\langle s,a,r,s'\rangle$ denote a buffered tuple of interaction. Define the following objective function:
\begin{equation}\label{eq:td_error}
  h(\theta,w):=\empiricalExpectation{\langle s,a,r,s'\rangle}\Big[\big(r+\gamma \max_{a'} \widehat Q(s',a';\theta)- \widehat Q(s,a;w)\big)^2\Big]\  . 
\end{equation}

Our proximal update is defined as follows:
\begin{equation}
w_{t+1} \leftarrow \arg\min_{w} h(w_t,w)+\frac{1}{2\tilde c}\norm{w-w_{t}}{2}{2}\ . \label{eq:prox_for_deep_rl}
\end{equation}

This algorithm closely resembles the standard proximal-point algorithm~\citep{RockafellarProxPaper1976,prox_parikh_boyd} with the important caveat that the function $h$ is now taking two vectors as input. At each iteration, we hold the first input constant while optimizing over the second input. 

In the optimization literature, the proximal-point algorithm is well-studied in contexts where an analytical solution to (\ref{eq:prox_for_deep_rl}) is available. With deep learning no closed-form solution exists, so we approximately solve (\ref{eq:prox_for_deep_rl}) by taking a fixed number of descent steps using stochastic gradients. Specifically, starting each iteration with $w=w_t$, we perform multiple $w$ updates ${w \leftarrow  w -\alpha\big(\nabla_{2} h(w_t,w) + \frac{1}{\tilde c} (w-w_t)\big)}$. We end the iteration by setting $w_{t+1}\leftarrow w$.
To make a connection to standard deep RL, the online weights $w$ could be thought of as the weights we maintain in the interim to solve~(\ref{eq:prox_for_deep_rl}) due to lack of a closed-form solution. Also, what is commonly referred to as the target network could better be thought of as just the previous iterate in the above proximal-point algorithm.

Observe that the update can be written as: $w\leftarrow \big(1-(\alpha/\tilde c)\big) \cdot w +(\alpha/\tilde c) \cdot w_t -\alpha \nabla_{2} h(w_t,w) \  .
\label{eq:online_update_2}
$ Notice the intuitively appealing form: we first compute a convex combination of $w_t$ and $w$, based on the hyper-parameters $\alpha$ and $\tilde c$, then add the gradient term to arrive at the next iterate of $w$. If $w_t$ and $w$ are close, the convex combination is close to $w$ itself and so this DQN with proximal update ({\sl DQN Pro}) would behave similarly to the original DQN. However, when $w$ strays too far from $w_t$, taking the convex combination ensures that $w$ gravitates towards the previous iterate $w_t$. The gradient signal from minimizing the squared TD error~(\ref{eq:td_error}) should then be strong enough to cancel this default gravitation towards $w_t$. The update includes standard DQN as a special case when $\tilde c\rightarrow \infty$. The pseudo-code for DQN is presented in the Appendix. The difference between DQN and DQN Pro is minimal (shown in gray), and corresponds with a few lines of code in our implementation.
\section{Experiments}\label{sec:experiment}

In this section, we empirically investigate the effectiveness of proximal updates in planning and reinforcement-learning algorithms. We begin by conducting experiments with PMPI in the context of approximate planning, and then move to large-scale RL experiments in Atari.

\subsection{PMPI Experiments}
We now focus on understanding the empirical impact of adding the proximal term on the performance of approximate PMPI. To this end, we use the pair of update equations:
\begin{eqnarray*}
\pi_{k} &\leftarrow& \mathcal{G} v_{k-1}\ ,\\
v_{k} &\leftarrow& (1-\beta)\big((\mathcal{T}^{\pi_{k}})^{n} v_{k-1} + \epsilon_{k}\big) +\beta v_{k-1}\ .
\end{eqnarray*}
For this experiment, we chose the toy $8\times8$ Frozen Lake environment from Open AI Gym~\citep{openaiGym}, where the transition and reward model of the environment is available to the planner. Using a small environment allows us to understand the impact of the proximal term in the simplest and most clear setting. Note also that we arranged the experiment so that the policy greedification step $\mathcal{G} v_{k-1}\ \forall k$ is error-free, so we can solely focus on the interplay between the proximal term and the error caused by imperfect policy evaluation.

We applied 100 iterations of PMPI, then measured the quality of the resultant policy $\pi\myeq\pi_{100}$ as defined by the distance between its true value and that of the optimal policy, namely $\norm{V^{\star} - V^{\pi}}{\infty}{}$. We repeated the experiment with different magnitudes of error, as well as different values of the $\beta$ parameter.

\begin{figure}[ht]
\vspace*{-0.7em}
\centering\captionsetup[subfigure]{justification=centering}
\begin{subfigure}{0.45\textwidth}
\centering 
\includegraphics[width=\textwidth]{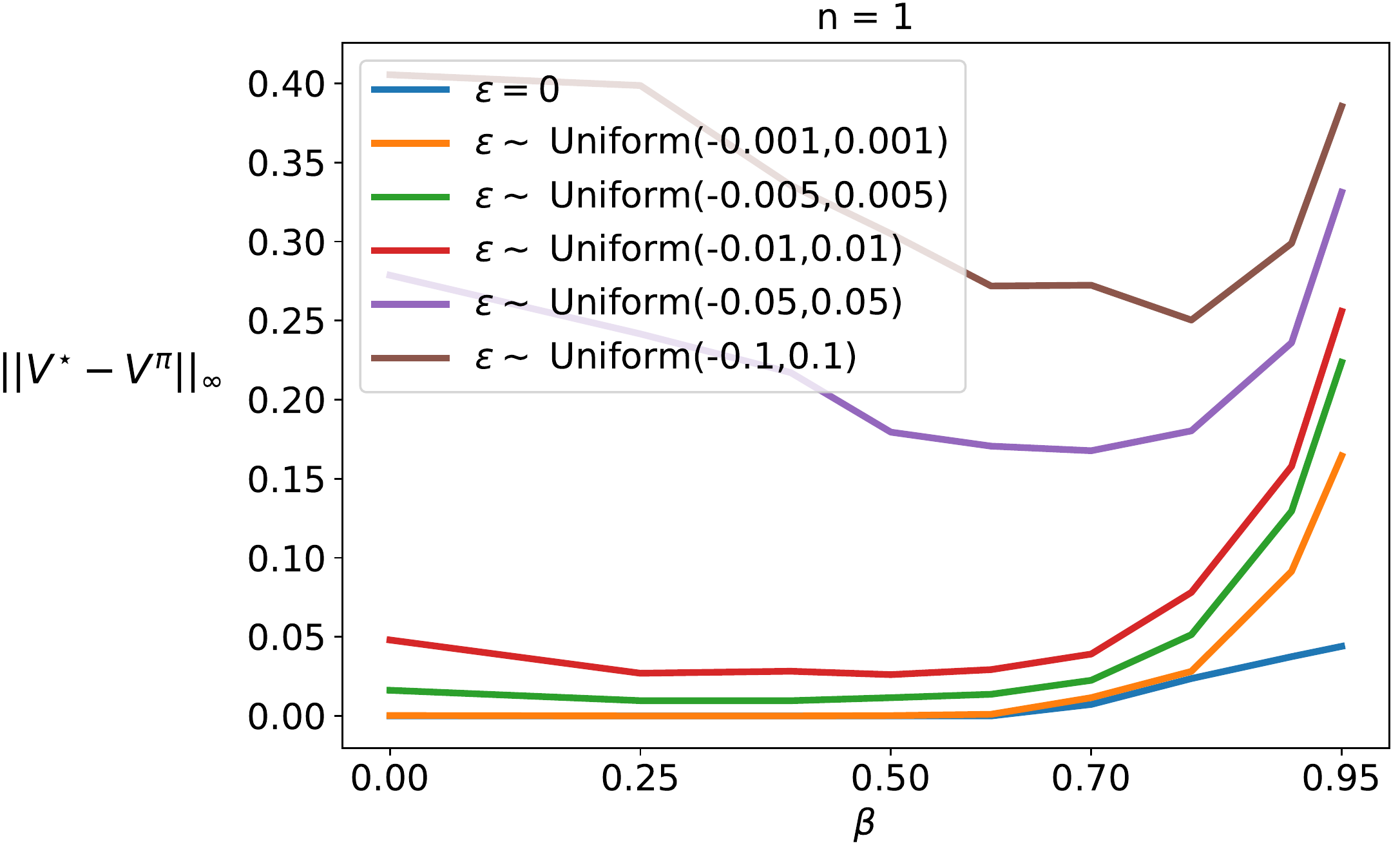} 
\end{subfigure}
\begin{subfigure}{ 0.45\textwidth} 
\centering 
\includegraphics[width=\textwidth]{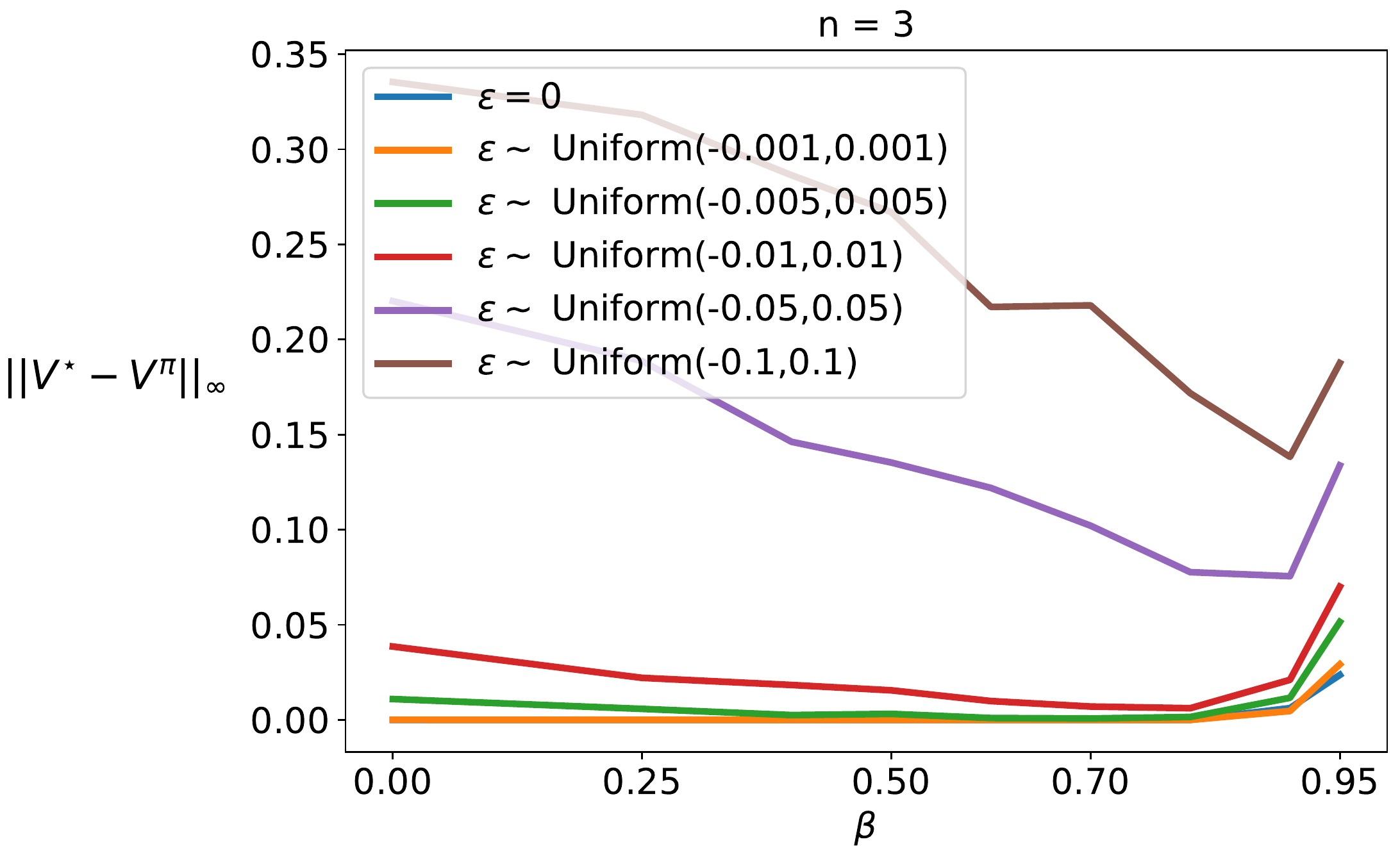} 
\end{subfigure}
\caption{ Performance of approximate PMPI has a U-shaped dependence on the parameter $\beta=\frac{1}{1+c}$. Results are averaged over 30 random seeds with $n=1$ \textbf{(left)} and $n=3$ \textbf{(right)}.}
\label{fig:MPI} 
\end{figure}

From Figure \ref{fig:MPI}, it is clear that the final performance exhibits a U-shape with respect to the parameter $\beta$. It is also noticable that the best-performing $\beta$ is shifting to the right side (larger values) as we increase the magnitude of noise. This trend makes sense, and is consistent with what is predicted by Theorem~2: As the noise level rises, we have more incentive to use larger (but not too large) $\beta$ values to hedge against it.

\subsection{Atari Experiments}

In this section, we evaluate the proximal (or Pro) agents relative to their original DQN-style counterparts on the Atari benchmark~\citep{bellemare_atari}, and show that endowing the agent with the proximal term can lead into significant improvements in the interim as well as in the final performance. We next investigate the utility of our proposed proximal term through further experiments. Please see the Appendix for a complete description of our experimental pipeline.

\subsubsection{Setup}\label{sec:expsetup}

We used 55 Atari games~\citep{bellemare_atari} to conduct our experimental evaluations. Following \citet{MachadoAracde2017} and \citet{dopamine}, we used sticky actions to inject stochasticity into the otherwise deterministic Atari emulator.

Our training and evaluation protocols and the hyper-parameter settings follow those of the Dopamine baseline~\citep{dopamine}. To report performance, we measured the undiscounted sum of rewards obtained by the learned policy during evaluation. We further report the learning curve for all experiments averaged across 5 random seeds. We reiterate that we used the exact same hyper-parameters for all agents to ensure a sound comparison. 
\begin{figure}[ht]
\vspace*{-0.7em}
\centering\captionsetup[subfigure]{justification=centering}
\begin{subfigure}{0.425\textwidth}
\centering 
\includegraphics[width=\textwidth]{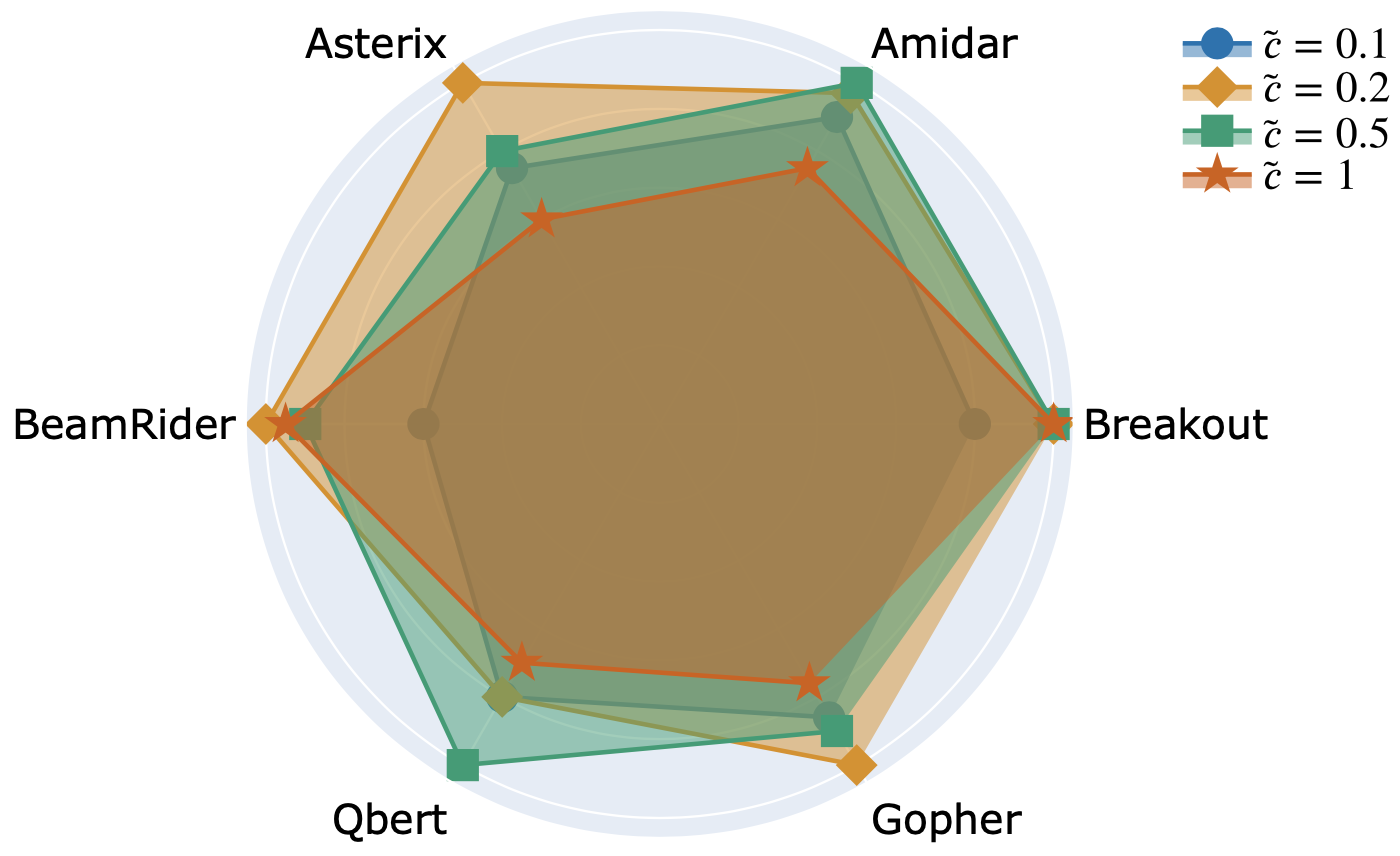} 
\end{subfigure}
\begin{subfigure}{ 0.425\textwidth} 
\centering 
\includegraphics[width=\textwidth]{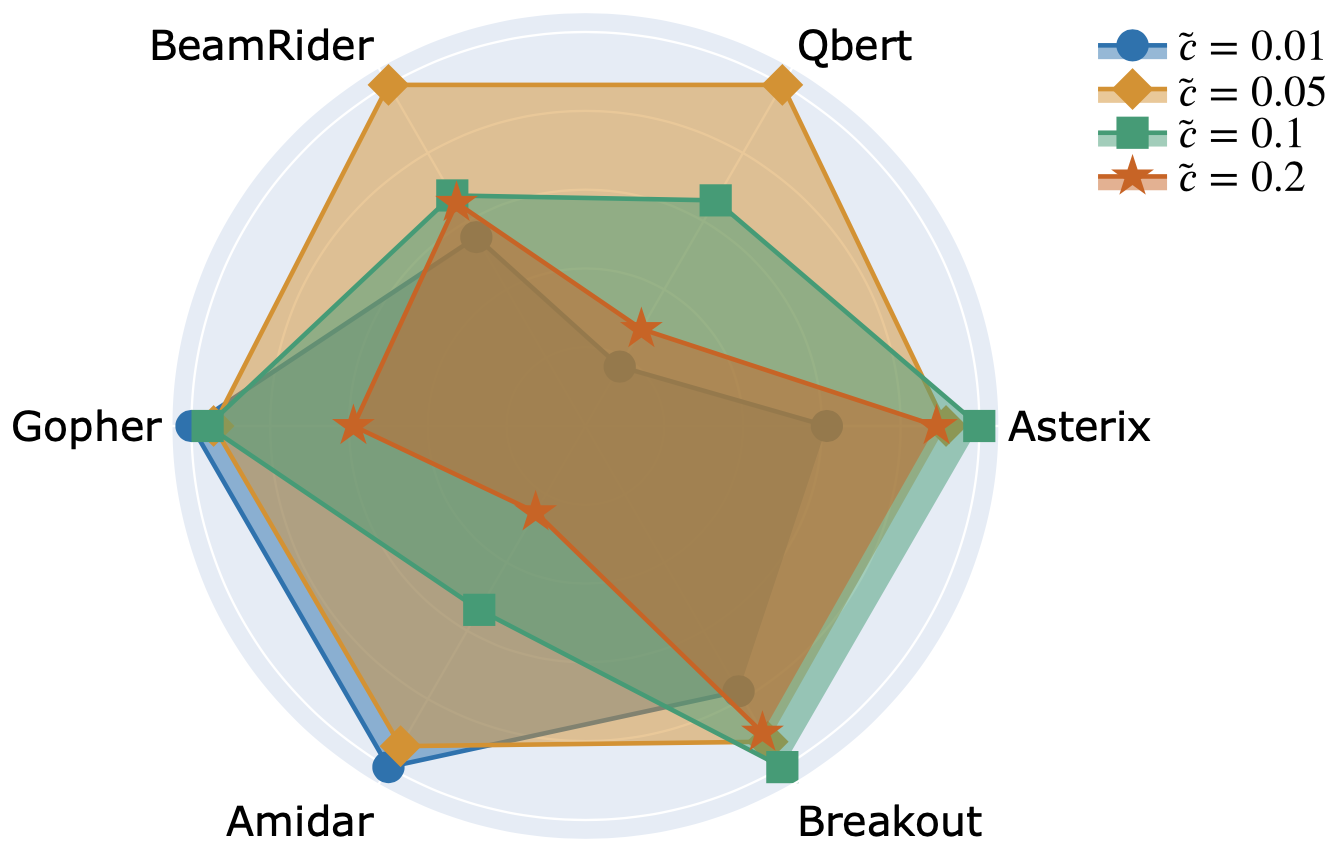} 
\end{subfigure}
\caption{ A minimal hyper-parameter tuning for $\tilde c$ in DQN Pro \textbf{(left)} and Rainbow Pro \textbf{(right)}.}
\label{fig:hyper_parameter} 
\end{figure}

Our Pro agents have a single additional hyper-parameter $\tilde c$. We did a minimal random search on 6 games to tune $\tilde c$. Figure~\ref{fig:hyper_parameter} visualizes the performance of Pro agents as a function of $\tilde c$. In light of this result, we set $\tilde c = 0.2$ for DQN Pro and $\tilde c = 0.05$ for Rainbow Pro. We used these values of $\tilde c$ for all 55 games, and note that we performed no further hyper-parameter tuning at all.

\subsubsection{Results}\label{sec:results}
The first question is whether endowing the DQN agent with the proximal term can yield significant improvements over the original DQN.
\begin{figure*}[ht]
\centering
\captionsetup{}
\begin{subfigure}{1\linewidth} 
\centering 
\includegraphics[width=\textwidth]{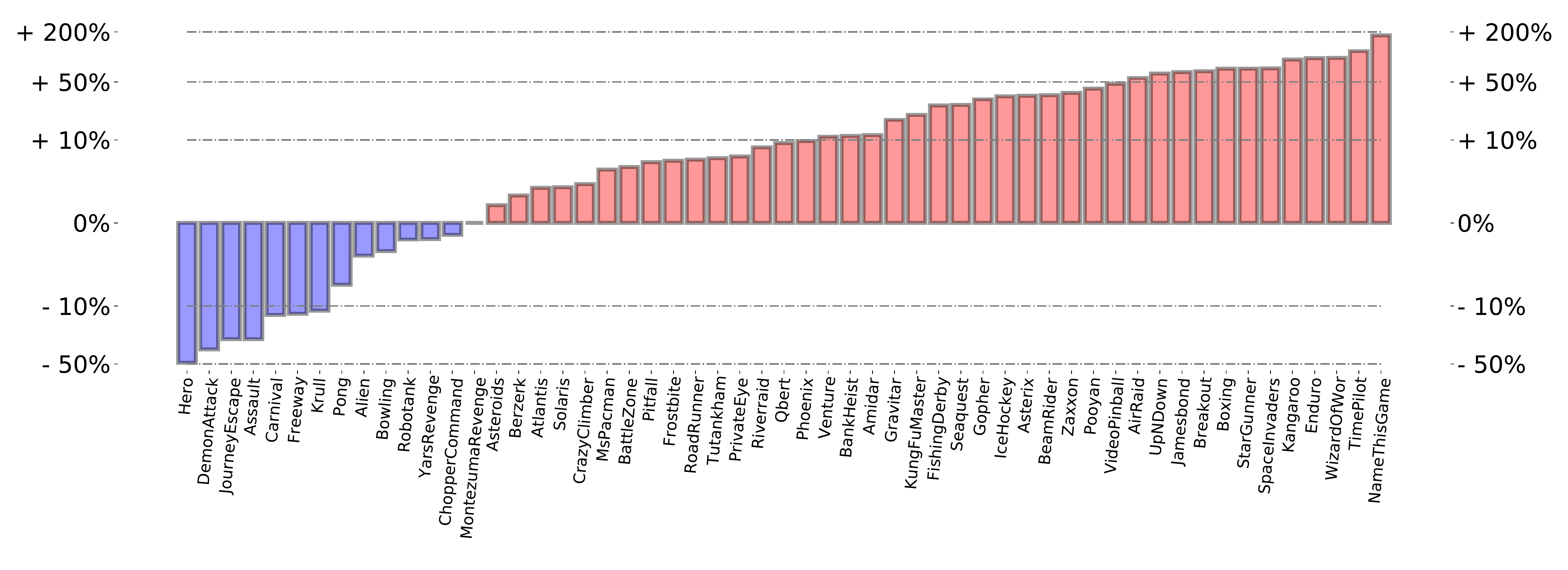} 
\end{subfigure} 
\begin{subfigure}{1\linewidth} 
\centering 
\includegraphics[width=\textwidth]{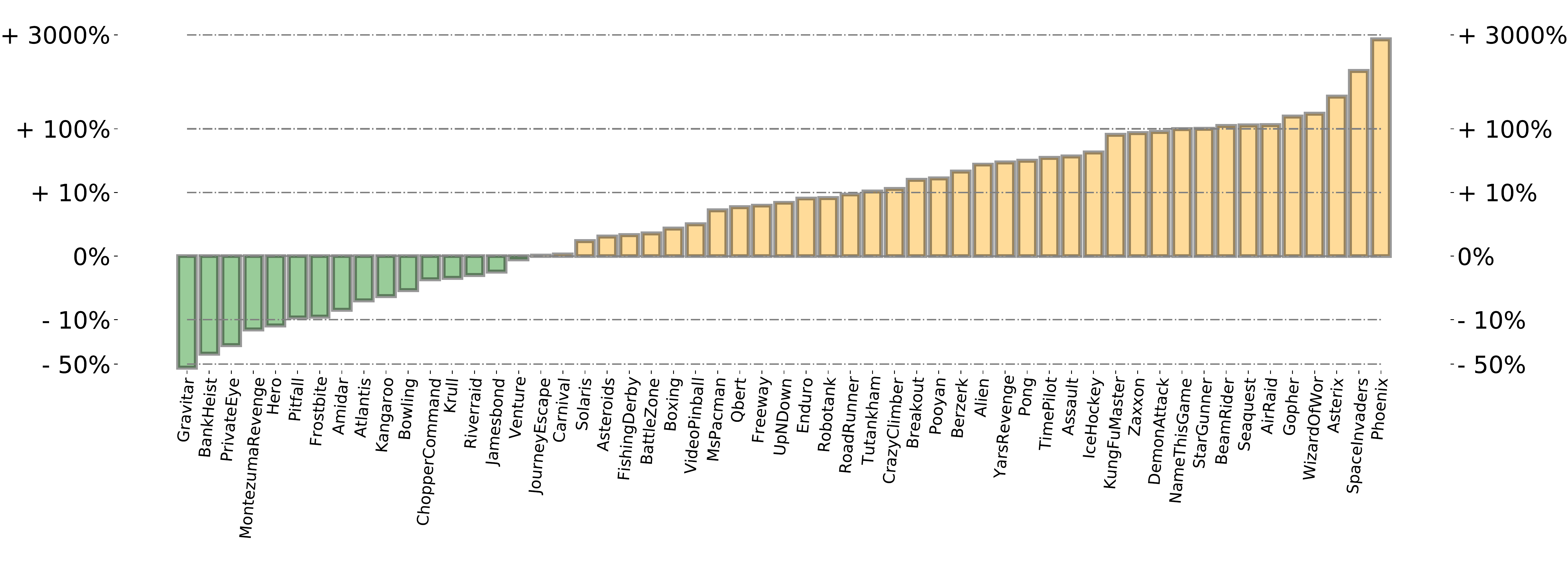}
\end{subfigure}
\caption{
Final gain for DQN Pro over DQN \textbf{(top)}, and Rainbow Pro over Rainbow \textbf{(bottom)}, averaged over 5 seeds. DQN Pro and Rainbow Pro significantly outperform their original counterparts.}
\label{fig:final_performance_atari}
\end{figure*} 
Figure \ref{fig:final_performance_atari} (top) shows a comparison between DQN and DQN Pro in terms of the final performance. In particular, following standard practice~\citep{wang2016dueling, iqn, van2019using}, for each game we compute:
$$
\frac{\textrm{Score}_{\textrm{DQN Pro}} - \textrm{Score}_{\textrm{DQN}} }{ \max(\textrm{Score}_{\textrm{DQN}},\textrm{Score}_{\textrm{Human}}) - \textrm{Score}_{\textrm{Random}} }
\ .$$

Bars shown in red indicate the games in which we observed better final performance for DQN Pro relative to DQN, and bars in blue indicate the opposite. The height of a bar denotes the magnitude of this improvement for the corresponding benchmark; notice that the y-axis is scaled logarithmically. We took human and random scores from previous work~\citep{nair2015massively, iqn}. It is clear that DQN Pro dramatically improves upon DQN. We defer to the Appendix for full learning curves on all games tested. 

Can we fruitfully combine the proximal term with some of the existing algorithmic improvements in DQN? To answer this question, we build on the Rainbow algorithm of~\citet{rainbow} who successfully combined numerous important algorithmic ideas in the value-based RL literature. We present this result in Figure \ref{fig:final_performance_atari} (bottom). Observe that the overall trend is for Rainbow Pro to yield large performance improvements over Rainbow.

\begin{figure}
  \begin{minipage}{0.55\textwidth}
    \includegraphics[width=\textwidth]{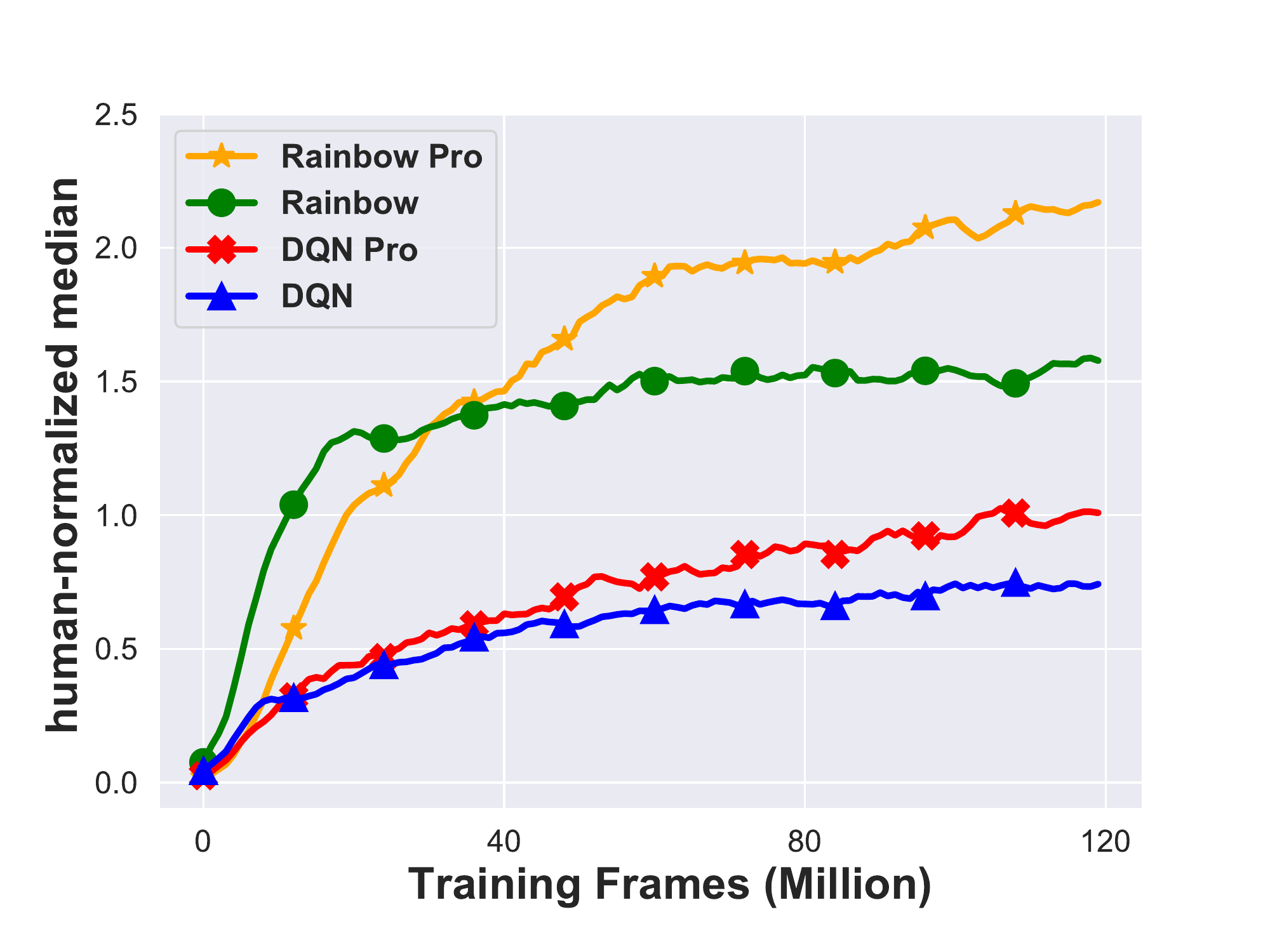}
\end{minipage}
\hspace{2pt}
\begin{minipage}{0.375\textwidth}
\includegraphics[width=.5\textwidth]{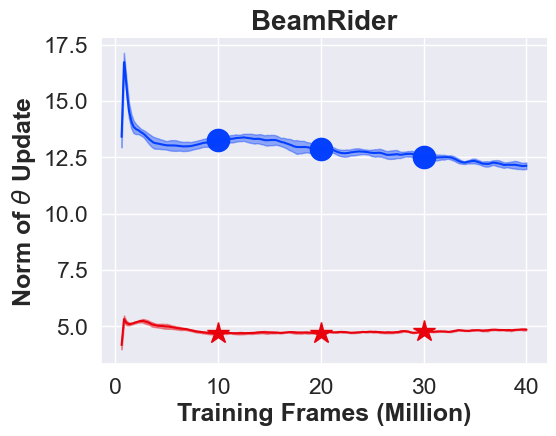}%
\includegraphics[width=.5\textwidth]{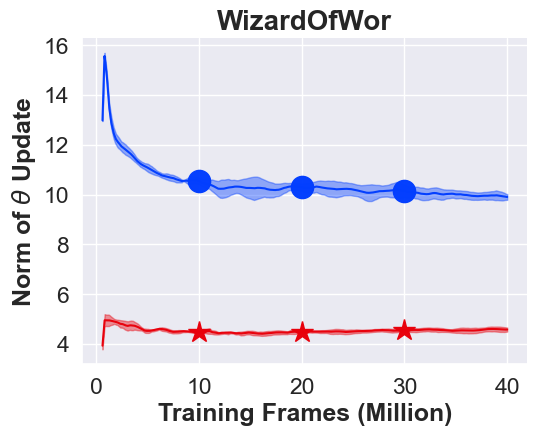} 
\includegraphics[width=.5\textwidth]{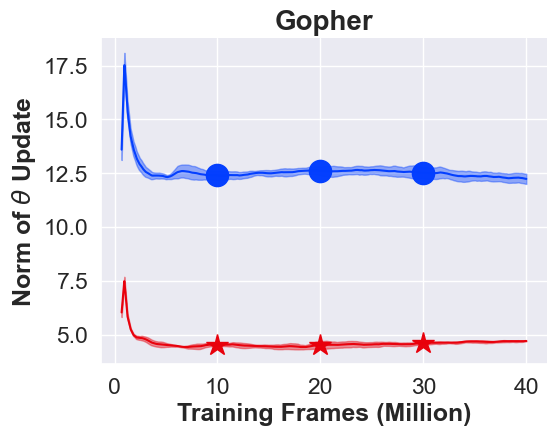}%
\includegraphics[width=.5\textwidth]{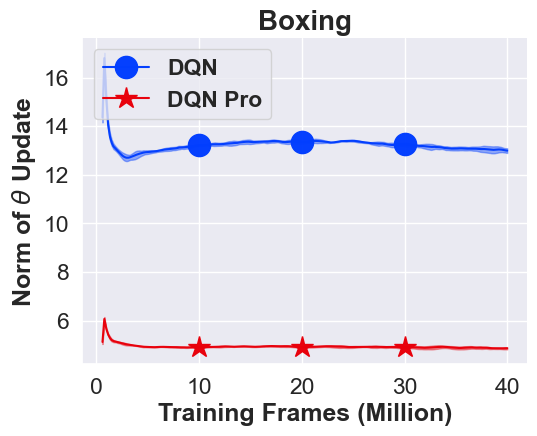} 
\label{fig:effect_of_proximal} 
\end{minipage}
\caption{\textbf{(left)}: Human-normalized median performance for DQN, Rainbow, DQN Pro, and Rainbow Pro on 55 Atari games. Results are averaged over 5 independent seeds. Our agents, Rainbow Pro (yellow) and DQN Pro (red) outperform their original counterparts Rainbow (green) and DQN (blue). \textbf{(right)}: Using the proximal term reduces the magnitude of target network updates.} 
\label{fig:main_learning_curve}
\end{figure}

Additionally, we measured the performance of our agents relative to human players. To this end, and again following previous work~\citep{wang2016dueling, iqn, van2019using}, for each agent we compute the human-normalized score:
$$
\frac{\textrm{Score}_{\textrm{Agent}} - \textrm{Score}_{\textrm{Random}} }{ \textrm{Score}_{\textrm{Human}}- \textrm{Score}_{\textrm{Random}} } .
$$
In Figure~\ref{fig:main_learning_curve} (left), we show the median of this score for all agents, which~\citet{wang2016dueling} and~\citet{rainbow} argued is a sensible quantity to track. We also show per-game learning curves with standard error in the Appendix.

We make two key observations from this figure. First, the very basic DQN Pro agent is capable of achieving human-level performance (1.0 on the y-axis) after 120 million frames. Second, the Rainbow Pro agent achieves 220 percent human-normalized score after only 120 million frames.

\subsubsection{Additional Experiments}
Our purpose in endowing the agent with the proximal term was to keep the online network in the vicinity of the target network, so it would be natural to ask if this desirable property can manifest itself in practice when using the proximal term. In Figure \ref{fig:main_learning_curve}, we answer this question affirmatively by plotting the magnitude of the update to the target network during synchronization. Notice that we periodically synchronize online and target networks, so the proximity of the online and target network should manifest itself in a low distance between two consecutive target networks. Indeed, the results demonstrate the success of the proximal term in terms of obtaining the desired proximity of online and target networks.

While using the proximal term leads to significant improvements, one may still wonder if the advantage of DQN Pro over DQN is merely stemming from a poorly-chosen \textit{period} hyper-parameter in the original DQN, as opposed to a truly more stable optimization in DQN Pro. To refute this hypothesis, we ran DQN with various settings of the \textit{period} hyper-parameter $\{2000, 4000, 8000, 12000\}$. This set included the default value of the hyper-parameter (8000) from the original paper~\citep{mnih2015human}, but also covered a wider set of settings.

Additionally, we tried an alternative update strategy for the target network, referred to as Polyak averaging, which was popularized in the context of continuous-action RL~\citep{ddpg}: $\theta \leftarrow \tau w + (1-\tau) \theta$. For this update strategy, too, we tried different settings of the $\tau$ hyper-parameter, namely $\{0.05, 0.005, 0.0005\}$, which includes the value $0.005$ used in numerous papers~\citep{ddpg,td3,rbfdqn}.
\begin{figure}
\begin{minipage}{0.425\textwidth}
\includegraphics[width=.5\textwidth]{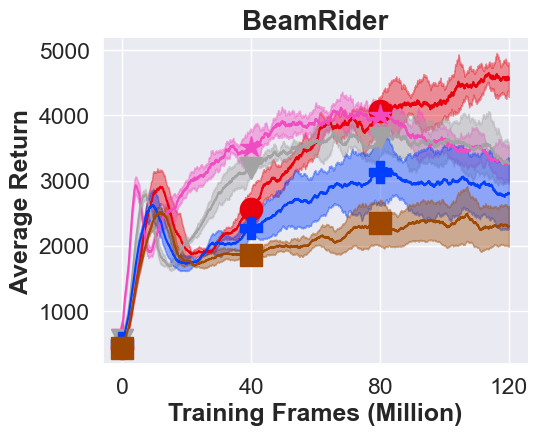}%
\includegraphics[width=.5\textwidth]{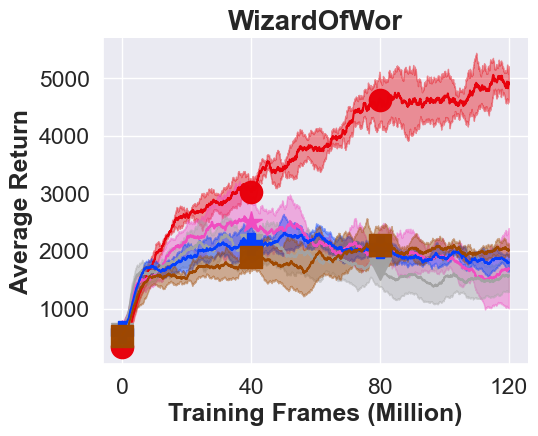}
\includegraphics[width=.5\textwidth]{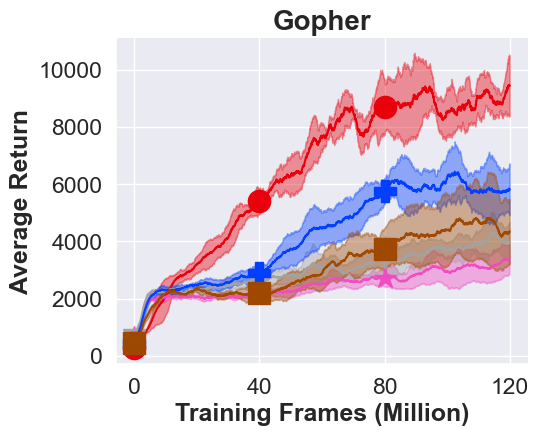}%
\includegraphics[width=.5\textwidth]{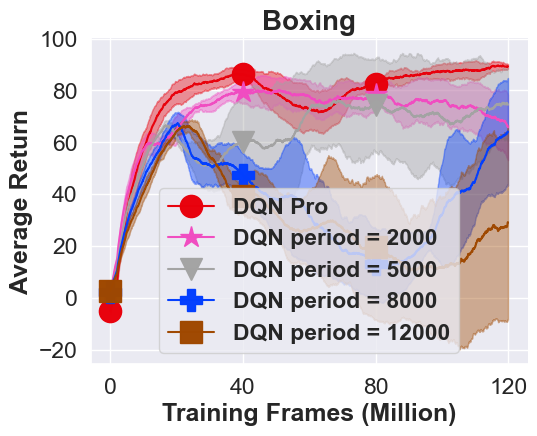}
\end{minipage}
\hspace{20pt}
\begin{minipage}{0.425\textwidth}
\includegraphics[width=.5\textwidth]{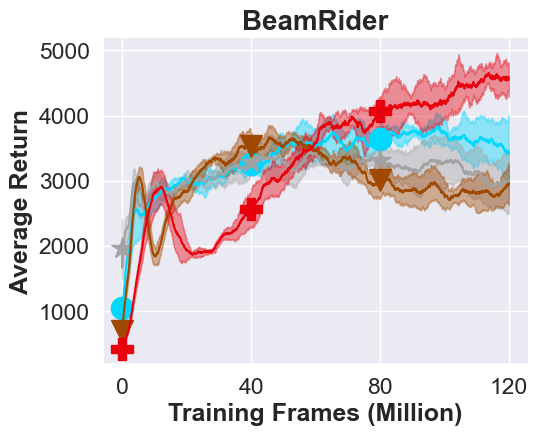}%
\includegraphics[width=.5\textwidth]{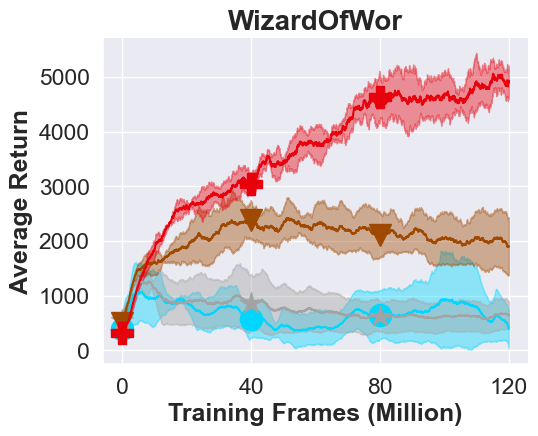}
\includegraphics[width=.5\textwidth]{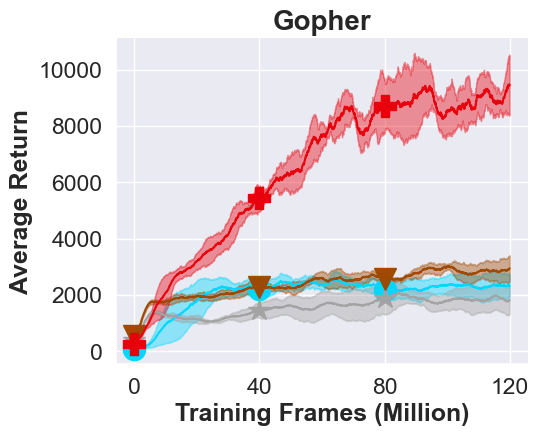}%
\includegraphics[width=.5\textwidth]{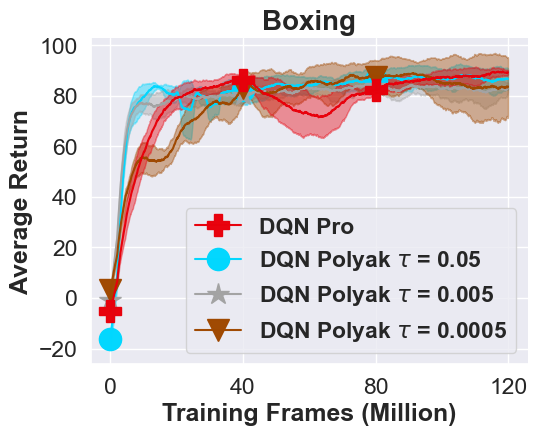}
\end{minipage}
\caption{A comparison between DQN Pro and DQN with periodic \textbf{(left)} and Polyak \textbf{(right)} updates.}
\label{fig:target_network}
\end{figure}

Figure \ref{fig:target_network} presents a comparison between DQN Pro and DQN with periodic and Polyak target updates for various hyper-parameter settings of \textit{period} and $\tau$. It is clear that DQN Pro is consistently outperforming the two alternatives regardless of the specific values of \textit{period} and $\tau$, thus clearly demonstrating that the improvement is stemming from a more stable optimization procedure leading to a better interplay between the two networks.

Finally, an alternative approach to ensuring lower distance between the online and the target network is to anneal the step size based on the number of updates performed on the online network since the last online-target synchronization. In this case we performed this experiment in 4 games where we knew proximal updates provide improvements based on our DQN Pro versus DQN resulst in Figure~\ref{fig:final_performance_atari}. In this case we linearly decreased the step size from the original DQN learning rate $\alpha$ to $\alpha'\ll \alpha$ where we tuned $\alpha'$ using random search. Annealing indeed improves DQN, but DQN Pro outperforms the improved version of DQN. Our intuition is that Pro agents only perform small updates when the target network is far from the online network, but naively decaying the learning rate can harm progress when the two networks are in vicinity of each other.
\begin{figure}
\centering\captionsetup[subfigure]{justification=centering,skip=0pt}
\begin{subfigure}[t]{0.24\textwidth} 
\centering 
\includegraphics[width=\textwidth]{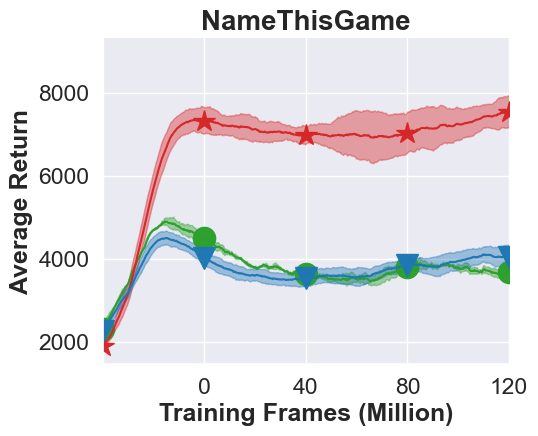} 
\label{fig:NameThisGameNoFrameskip-v0_learning_curves_4_domains.png} 
\end{subfigure}%
~ 
\begin{subfigure}[t]{ 0.24\textwidth} 
\centering 
\includegraphics[width=\textwidth]{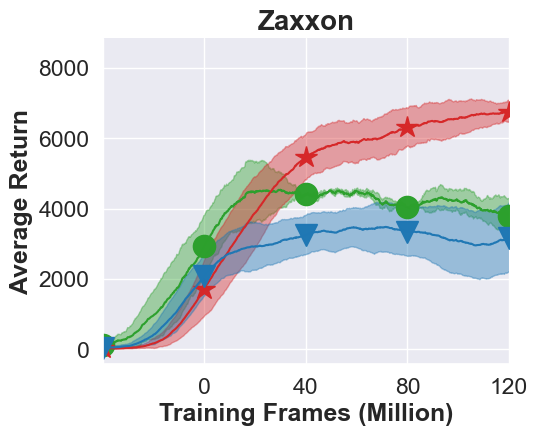} 
\label{fig:ZaxxonNoFrameskip-v0_learning_curves_4_domains.png} 
\end{subfigure}%
~ 
\begin{subfigure}[t]{ 0.24\textwidth} 
\centering 
\includegraphics[width=\textwidth]{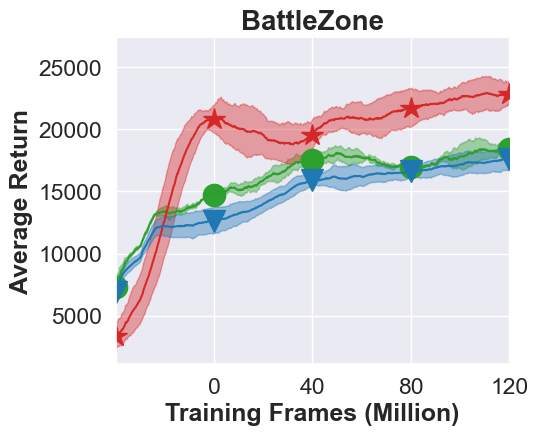} 
\label{fig:BattleZoneNoFrameskip-v0_learning_curves_4_domains.png} 
\end{subfigure}%
~ 
\begin{subfigure}[t]{ 0.24\textwidth} 
\centering 
\includegraphics[width=\textwidth]{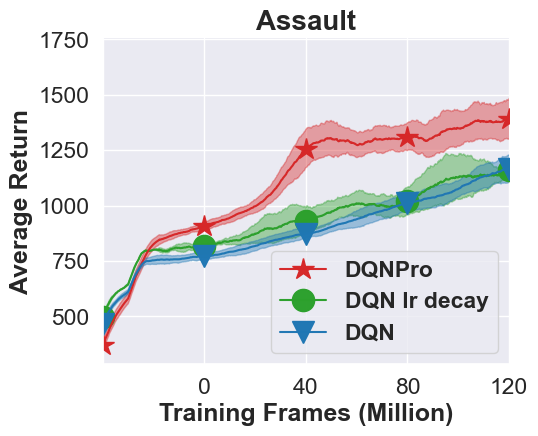} 
\label{fig:AssaultNoFrameskip-v0_learning_curves_4_domains.png} 
\end{subfigure}%
\vspace{-.5cm}
\caption{Learning-rate decay does not bridge the gap between DQN and DQN Pro.} 
\label{fig:rebut} 
\end{figure} 

\section{Discussion}
In our experience using proximal updates in the parameter space were far superior than proximal updates in the value space. We believe this is because the parameter-space definition can enforce the proximity globally, while in the value space one can only hope to obtain proximity locally and on a batch of samples. One may hope to use natural gradients to enforce value-space proximity in a more principled way, but doing so usually requires significantly more computational resources~\cite{knight2018natural}. This is in contrast to our proximal updates which add negligible computational cost in the simple form of taking a dimension-wise weighted average of two weight vectors.

In addition, for a smooth (Lipschitz) $Q$ function, performing parameter-space regularization guarantees function-space regularization. Concretely: $\forall s, \forall a \ |Q(s,a;\theta)-Q(s,a;\theta')|\leq L||\theta-\theta'||\ , $ where $L$ is the Lipschitz constant of $Q$. Moreover, deep networks are Lipschitz~\citep{neyshabur2015norm, asadi2018lipschitz}, because they are constructed using compositions of Lipschitz functions (such as ReLU, convolutions, etc) and that composition of Lipschitz functions is Lipschitz. So performing value-space updates may be an overkill. Lipschitz property of deep networks has successfully been leveraged in other contexts, such as in generative adversarial training~\cite{arjovsky2017wasserstein}.

A key selling point of our result is simplicity, because simple results are easy to understand, implement, and reproduce. We obtained significant performance improvements by adding just a few lines of codes to the publicly available implementations of DQN and Rainbow~\cite{dopamine}.
\section{Related Work}

The introduction of proximal operators could be traced back to the seminal work of~\citet{moreauhal-01867195,Moreau1965}, \citet{prox_first_work} and \citet{RockafellarProxPaper1976}, and the use of the proximal operators has since expanded into many areas of science such as signal processing~\citep{proximal_signal_processing}, statistics and machine learning~\citep{beck2009fast, prox_statistics, Reddi2015DoublyRC}, and convex optimization~\citep{prox_parikh_boyd, Bertsekas2011IncrementalPM, Bertsekas15cSrv}.

In the context of RL, \citet{proximal_rl_mahadevan} introduced a proximal theory for deriving convergent off-policy algorithms with linear function approximation. One intriguing characteristic of their work is that they perform updates in primal-dual space, a property that was leveraged in sample complexity analysis~\citep{prox_finite_liu} for the proximal counterparts of the gradient temporal-difference algorithm~\citep{gtd}. Proximal operators also have appeared in the deep RL literature. For instance, \citet{fakoor2020metaqlearning} used proximal operators for meta learning, and \citet{maggipinto2020proximal} improved TD3~\citep{td3} by employing a stochastic proximal-point interpretation.

The effect of the proximal term in our work is reminiscent of the use of trust regions in policy-gradient algorithms~\citep{trpo,ppo, YuhuiNEURIPS2019, fakoor2019p3o, tomar2021mirror}. However, three factors differentiate our work: we define the proximal term using the value function, not the policy, we enforce the proximal term in the parameter space, as opposed to the function space, and we use the target network as the previous iterate in our proximal definition. 


\section{Conclusion and Future work}
We showed a clear advantage for using proximal terms to perform slower but more effective updates in approximate planning and reinforcement learning. Our results demonstrated that proximal updates lead to more robustness with respect to noise. Several improvements to proximal methods exist, such as the acceleration algorithm~\citep{nesterov,li2015accelerated}, as well as using other proximal terms~\citep{proximal_signal_processing}, which we leave for future work.
\label{s:conc}
\section{Acknowledgment}
We thank Lihong Li, Pratik Chaudhari, and Shoham Sabach for their valuable insights in different stages of this work.
\clearpage
\bibliography{reference}

\begin{thebibliography}{60}
\providecommand{\natexlab}[1]{#1}
\providecommand{\url}[1]{\texttt{#1}}
\expandafter\ifx\csname urlstyle\endcsname\relax
  \providecommand{\doi}[1]{doi: #1}\else
  \providecommand{\doi}{doi: \begingroup \urlstyle{rm}\Url}\fi

\bibitem[Arjovsky et~al.(2017)Arjovsky, Chintala, and
  Bottou]{arjovsky2017wasserstein}
Arjovsky, M., Chintala, S., and Bottou, L.
\newblock Wasserstein generative adversarial networks.
\newblock In \emph{ICML}, 2017.

\bibitem[Asadi et~al.(2018)]{asadi2018lipschitz}
Asadi et~al.
\newblock Lipschitz continuity in model-based reinforcement learning.
\newblock In \emph{ICML}, 2018.

\bibitem[Asadi et~al.(2021)Asadi, Parikh, Parr, Konidaris, and Littman]{rbfdqn}
Asadi, K., Parikh, N., Parr, R.~E., Konidaris, G.~D., and Littman, M.~L.
\newblock Deep radial-basis value functions for continuous control.
\newblock In \emph{AAAI Conference on Artificial Intelligence}, 2021.

\bibitem[Beck \& Teboulle(2009)Beck and Teboulle]{beck2009fast}
Beck, A. and Teboulle, M.
\newblock A fast iterative shrinkage-thresholding algorithm for linear inverse
  problems.
\newblock \emph{SIAM Journal on Imaging Sciences}, 2009.

\bibitem[Bellemare et~al.(2013)Bellemare, Naddaf, Veness, and
  Bowling]{bellemare_atari}
Bellemare, M.~G., Naddaf, Y., Veness, J., and Bowling, M.
\newblock The arcade learning environment: An evaluation platform for general
  agents.
\newblock \emph{Journal of Artificial Intelligence Research}, 2013.

\bibitem[Bellman(1957)]{bellamn1957}
Bellman, R.~E.
\newblock \emph{Dynamic Programming}.
\newblock 1957.

\bibitem[Bertsekas(2011{\natexlab{a}})]{Bertsekas15cSrv}
Bertsekas, D.~P.
\newblock Incremental gradient, subgradient, and proximal methods for convex
  optimization: A survey.
\newblock \emph{Optimization for Machine Learning}, 2011{\natexlab{a}}.

\bibitem[Bertsekas(2011{\natexlab{b}})]{Bertsekas2011IncrementalPM}
Bertsekas, D.~P.
\newblock Incremental proximal methods for large scale convex optimization.
\newblock \emph{Mathematical Programming}, 2011{\natexlab{b}}.

\bibitem[Bertsekas \& Tsitsiklis(1996)Bertsekas and
  Tsitsiklis]{bertsekas1996neuro}
Bertsekas, D.~P. and Tsitsiklis, J.~N.
\newblock \emph{Neuro-dynamic programming}.
\newblock Athena Scientific, 1996.

\bibitem[Brockman et~al.(2016)Brockman, Cheung, Pettersson, Schneider,
  Schulman, Tang, and Zaremba]{openaiGym}
Brockman, G., Cheung, V., Pettersson, L., Schneider, J., Schulman, J., Tang,
  J., and Zaremba, W.
\newblock Openai gym, 2016.

\bibitem[Castro et~al.(2018)Castro, Moitra, Gelada, Kumar, and
  Bellemare]{dopamine}
Castro, P.~S., Moitra, S., Gelada, C., Kumar, S., and Bellemare, M.~G.
\newblock Dopamine: {A} {R}esearch {F}ramework for {D}eep {R}einforcement
  {L}earning.
\newblock 2018.

\bibitem[Combettes \& Pesquet(2009)Combettes and
  Pesquet]{proximal_signal_processing}
Combettes, P.~L. and Pesquet, J.-C.
\newblock {Proximal Splitting Methods in Signal Processing}.
\newblock \emph{Fixed-point algorithms for inverse problems in science and
  engineering}, 2009.

\bibitem[Dabney et~al.(2018)Dabney, Ostrovski, Silver, and Munos]{iqn}
Dabney, W., Ostrovski, G., Silver, D., and Munos, R.
\newblock Implicit quantile networks for distributional reinforcement learning.
\newblock In \emph{International conference on machine learning}, pp.\
  1096--1105. PMLR, 2018.

\bibitem[Fakoor et~al.(2020{\natexlab{a}})Fakoor, Chaudhari, and
  Smola]{fakoor2019p3o}
Fakoor, R., Chaudhari, P., and Smola, A.~J.
\newblock P3{O}: Policy-on policy-off policy optimization.
\newblock In \emph{Conference on Uncertainty in Artificial Intelligence},
  2020{\natexlab{a}}.

\bibitem[Fakoor et~al.(2020{\natexlab{b}})Fakoor, Chaudhari, Soatto, and
  Smola]{fakoor2020metaqlearning}
Fakoor, R., Chaudhari, P., Soatto, S., and Smola, A.~J.
\newblock Meta-{Q}-learning.
\newblock In \emph{International Conference on Learning Representations},
  2020{\natexlab{b}}.

\bibitem[Fujimoto et~al.(2018)Fujimoto, Hoof, and Meger]{td3}
Fujimoto, S., Hoof, H., and Meger, D.
\newblock Addressing function approximation error in actor-critic methods.
\newblock In \emph{International Conference on Machine Learning}, 2018.

\bibitem[Hessel et~al.(2018)Hessel, Modayil, van Hasselt, Schaul, Ostrovski,
  Dabney, Horgan, Piot, Azar, and Silver]{rainbow}
Hessel, M., Modayil, J., van Hasselt, H., Schaul, T., Ostrovski, G., Dabney,
  W., Horgan, D., Piot, B., Azar, M., and Silver, D.
\newblock Rainbow: Combining improvements in deep reinforcement learning.
\newblock In \emph{AAAI Conference on Artificial Intelligence}, 2018.

\bibitem[Howard(1960)]{howard1960dynamic}
Howard, R.~A.
\newblock Dynamic programming and markov processes.
\newblock 1960.

\bibitem[Kim et~al.(2019)Kim, Asadi, Littman, and Konidaris]{kim2019deepmellow}
Kim, S., Asadi, K., Littman, M., and Konidaris, G.
\newblock Deepmellow: removing the need for a target network in deep
  q-learning.
\newblock In \emph{International Joint Conference on Artificial Intelligence},
  2019.

\bibitem[Kingma \& Ba(2015)Kingma and Ba]{KingmaB14}
Kingma, D.~P. and Ba, J.
\newblock Adam: {A} method for stochastic optimization.
\newblock In \emph{International Conference on Learning Representations}, 2015.

\bibitem[Knight \& Lerner(2018)Knight and Lerner]{knight2018natural}
Knight, E. and Lerner, O.
\newblock Natural gradient deep q-learning.
\newblock \emph{arXiv preprint arXiv:1803.07482}, 2018.

\bibitem[Kober et~al.(2013)Kober, Bagnell, and Peters]{kober_robotics}
Kober, J., Bagnell, J.~A., and Peters, J.
\newblock Reinforcement learning in robotics: A survey.
\newblock \emph{International Journal of Robotics Research}, 2013.

\bibitem[Lee \& He(2019)Lee and He]{target_based_TD}
Lee, D. and He, N.
\newblock Target-based temporal-difference learning.
\newblock In \emph{International Conference on Machine Learning}, 2019.

\bibitem[Li \& Lin(2015)Li and Lin]{li2015accelerated}
Li, H. and Lin, Z.
\newblock Accelerated proximal gradient methods for nonconvex programming.
\newblock \emph{Advances in neural information processing systems}, 2015.

\bibitem[Lillicrap et~al.(2015)Lillicrap, Hunt, Pritzel, Heess, Erez, Tassa,
  Silver, and Wierstra]{ddpg}
Lillicrap, T.~P., Hunt, J.~J., Pritzel, A., Heess, N., Erez, T., Tassa, Y.,
  Silver, D., and Wierstra, D.
\newblock Continuous control with deep reinforcement learning.
\newblock In \emph{International Conference on Learning Representations}, 2015.

\bibitem[Lin(1992)]{lin1992self}
Lin, L.-J.
\newblock Self-improving reactive agents based on reinforcement learning,
  planning and teaching.
\newblock \emph{Machine learning}, 1992.

\bibitem[Littman \& Szepesv{\'a}ri(1996)Littman and
  Szepesv{\'a}ri]{littman_csaba}
Littman, M.~L. and Szepesv{\'a}ri, C.
\newblock A generalized reinforcement-learning model: Convergence and
  applications.
\newblock In \emph{ICML}, volume~96, pp.\  310--318. Citeseer, 1996.

\bibitem[Liu et~al.(2020)Liu, Liu, Ghavamzadeh, Mahadevan, and
  Petrik]{prox_finite_liu}
Liu, B., Liu, J., Ghavamzadeh, M., Mahadevan, S., and Petrik, M.
\newblock Finite-sample analysis of proximal gradient {TD} algorithms.
\newblock In \emph{Conference on Uncertainty in Artificial Intelligence}, 2020.

\bibitem[Machado et~al.(2018)Machado, Bellemare, Talvitie, Veness, Hausknecht,
  and Bowling]{MachadoAracde2017}
Machado, M.~C., Bellemare, M.~G., Talvitie, E., Veness, J., Hausknecht, M., and
  Bowling, M.
\newblock Revisiting the arcade learning environment: Evaluation protocols and
  open problems for general agents.
\newblock \emph{Journal of Artificial Intelligence Research}, 2018.

\bibitem[Maggipinto et~al.(2020)Maggipinto, Susto, and
  Chaudhari]{maggipinto2020proximal}
Maggipinto, M., Susto, G.~A., and Chaudhari, P.
\newblock Proximal deterministic policy gradient.
\newblock In \emph{International Conference on Intelligent Robots and Systems},
  2020.

\bibitem[Mahadevan et~al.(2014)Mahadevan, Liu, Thomas, Dabney, Giguere, Jacek,
  Gemp, and Liu]{proximal_rl_mahadevan}
Mahadevan, S., Liu, B., Thomas, P., Dabney, W., Giguere, S., Jacek, N., Gemp,
  I., and Liu, J.
\newblock {Proximal Reinforcement Learning: A New Theory of Sequential Decision
  Making in Primal-Dual Spaces}.
\newblock \emph{arXiv}, 2014.

\bibitem[Martinet(1970)]{prox_first_work}
Martinet, B.
\newblock Regularisation, d'in{\'e}quations variationelles par approximations
  succesives.
\newblock \emph{Revue Francaise d'informatique et de Recherche operationelle},
  1970.

\bibitem[Mnih et~al.(2015)Mnih, Kavukcuoglu, Silver, Rusu, Veness, Bellemare,
  Graves, Riedmiller, Fidjeland, Ostrovski, et~al.]{mnih2015human}
Mnih, V., Kavukcuoglu, K., Silver, D., Rusu, A.~A., Veness, J., Bellemare,
  M.~G., Graves, A., Riedmiller, M., Fidjeland, A.~K., Ostrovski, G., et~al.
\newblock Human-level control through deep reinforcement learning.
\newblock \emph{Nature}, 2015.

\bibitem[Moreau(1962)]{moreauhal-01867195}
Moreau, J.~J.
\newblock {Fonctions convexes duales et points proximaux dans un espace
  hilbertien}.
\newblock \emph{Comptes rendus hebdomadaires des s{\'e}ances de l'Acad{\'e}mie
  des sciences}, 1962.

\bibitem[Moreau(1965)]{Moreau1965}
Moreau, J.~J.
\newblock Proximit\'e et dualit\'e dans un espace hilbertien.
\newblock \emph{Bulletin de la Soci\'et\'e Math\'ematique de France}, 1965.

\bibitem[Nair et~al.(2015)Nair, Srinivasan, Blackwell, Alcicek, Fearon,
  De~Maria, Panneershelvam, Suleyman, Beattie, Petersen,
  et~al.]{nair2015massively}
Nair, A., Srinivasan, P., Blackwell, S., Alcicek, C., Fearon, R., De~Maria, A.,
  Panneershelvam, V., Suleyman, M., Beattie, C., Petersen, S., et~al.
\newblock Massively parallel methods for deep reinforcement learning.
\newblock \emph{arXiv preprint arXiv:1507.04296}, 2015.

\bibitem[Nesterov(1983)]{nesterov}
Nesterov, Y.
\newblock A method for unconstrained convex minimization problem with the rate
  of convergence o (1/k\^{} 2).
\newblock In \emph{Doklady an USSR}, 1983.

\bibitem[Neyshabur et~al.(2015)]{neyshabur2015norm}
Neyshabur et~al.
\newblock Norm-based capacity control in neural networks.
\newblock In \emph{COLT}, 2015.

\bibitem[Parikh \& Boyd(2014)Parikh and Boyd]{prox_parikh_boyd}
Parikh, N. and Boyd, S.
\newblock {proximal algorithms}.
\newblock \emph{Foundations and Trends in optimization}, 2014.

\bibitem[Polson et~al.(2015)Polson, Scott, and Willard]{prox_statistics}
Polson, N.~G., Scott, J.~G., and Willard, B.~T.
\newblock Proximal algorithms in statistics and machine learning.
\newblock \emph{Statistical Science}, 2015.

\bibitem[Puterman(1994)]{PutermanMDP1994}
Puterman, M.~L.
\newblock \emph{Markov Decision Processes: Discrete Stochastic Dynamic
  Programming}.
\newblock 1994.

\bibitem[Reddi et~al.(2015)Reddi, Poczos, and Smola]{Reddi2015DoublyRC}
Reddi, S., Poczos, B., and Smola, A.
\newblock Doubly robust covariate shift correction.
\newblock In \emph{AAAI Conference on Artificial Intelligence}, 2015.

\bibitem[Rockafellar(1976)]{RockafellarProxPaper1976}
Rockafellar, R.~T.
\newblock Monotone operators and the proximal point algorithm.
\newblock \emph{SIAM Journal on Control and Optimization}, 1976.

\bibitem[Scherrer et~al.(2015)Scherrer, Ghavamzadeh, Gabillon, Lesner, and
  Geist]{ampi}
Scherrer, B., Ghavamzadeh, M., Gabillon, V., Lesner, B., and Geist, M.
\newblock Approximate modified policy iteration and its application to the game
  of tetris.
\newblock \emph{J. Mach. Learn. Res.}, 16:\penalty0 1629--1676, 2015.

\bibitem[Schulman et~al.(2015)Schulman, Levine, Abbeel, Jordan, and
  Moritz]{trpo}
Schulman, J., Levine, S., Abbeel, P., Jordan, M., and Moritz, P.
\newblock Trust region policy optimization.
\newblock In \emph{International conference on machine learning}, 2015.

\bibitem[Schulman et~al.(2017)Schulman, Wolski, Dhariwal, Radford, and
  Klimov]{ppo}
Schulman, J., Wolski, F., Dhariwal, P., Radford, A., and Klimov, O.
\newblock Proximal policy optimization algorithms.
\newblock \emph{arXiv}, 2017.

\bibitem[Silver et~al.(2017)Silver, Schrittwieser, Simonyan, Antonoglou, Huang,
  Guez, Hubert, Baker, Lai, Bolton, et~al.]{silver2017mastering}
Silver, D., Schrittwieser, J., Simonyan, K., Antonoglou, I., Huang, A., Guez,
  A., Hubert, T., Baker, L., Lai, M., Bolton, A., et~al.
\newblock Mastering the game of go without human knowledge.
\newblock \emph{Nature}, 2017.

\bibitem[Smirnova \& Dohmatob(2020)Smirnova and
  Dohmatob]{smirnova2020convergence}
Smirnova, E. and Dohmatob, E.
\newblock On the convergence of smooth regularized approximate value iteration
  schemes.
\newblock \emph{Advances in Neural Information Processing Systems}, 33, 2020.

\bibitem[Sutton(1988)]{sutton_td}
Sutton, R.~S.
\newblock Learning to predict by the methods of temporal differences.
\newblock \emph{Machine learning}, 1988.

\bibitem[Sutton \& Barto(2018)Sutton and Barto]{rl_book}
Sutton, R.~S. and Barto, A.~G.
\newblock \emph{Reinforcement learning: An introduction}.
\newblock 2018.

\bibitem[Sutton et~al.(2008)Sutton, Szepesv{\'a}ri, and Maei]{gtd}
Sutton, R.~S., Szepesv{\'a}ri, C., and Maei, H.~R.
\newblock A convergent {O}(n) temporal-difference algorithm for off-policy
  learning with linear function approximation.
\newblock In \emph{Advances in Neural Information Processing Systems}, 2008.

\bibitem[Tesauro(1994)]{td_gammon}
Tesauro, G.
\newblock T{D}-gammon, a self-teaching backgammon program, achieves
  master-level play.
\newblock \emph{Neural computation}, 1994.

\bibitem[Tomar et~al.(2021)Tomar, Shani, Efroni, and
  Ghavamzadeh]{tomar2021mirror}
Tomar, M., Shani, L., Efroni, Y., and Ghavamzadeh, M.
\newblock Mirror descent policy optimization, 2021.

\bibitem[van Hasselt et~al.(2018)van Hasselt, Doron, Strub, Hessel, Sonnerat,
  and Modayil]{van_hasselt_deadly_triad}
van Hasselt, H., Doron, Y., Strub, F., Hessel, M., Sonnerat, N., and Modayil,
  J.
\newblock Deep reinforcement learning and the deadly triad.
\newblock \emph{arXiv}, 2018.

\bibitem[van Seijen et~al.(2019)van Seijen, Fatemi, and Tavakoli]{van2019using}
van Seijen, H., Fatemi, M., and Tavakoli, A.
\newblock Using a logarithmic mapping to enable lower discount factors in
  reinforcement learning.
\newblock \emph{Advances in Neural Information Processing Systems}, 2019.

\bibitem[Wang et~al.(2019)Wang, He, Tan, and Gan]{YuhuiNEURIPS2019}
Wang, Y., He, H., Tan, X., and Gan, Y.
\newblock Trust region-guided proximal policy optimization.
\newblock In \emph{Advances in Neural Information Processing Systems}, 2019.

\bibitem[Wang et~al.(2016)Wang, Schaul, Hessel, Hasselt, Lanctot, and
  Freitas]{wang2016dueling}
Wang, Z., Schaul, T., Hessel, M., Hasselt, H., Lanctot, M., and Freitas, N.
\newblock Dueling network architectures for deep reinforcement learning.
\newblock In \emph{International conference on machine learning}, pp.\
  1995--2003. PMLR, 2016.

\bibitem[Watkins \& Dayan(1992)Watkins and Dayan]{q_learning}
Watkins, C.~J. and Dayan, P.
\newblock Q-learning.
\newblock \emph{Machine learning}, 1992.

\bibitem[Williams et~al.(2017)Williams, Asadi, and Zweig]{williams2017hybrid}
Williams, J.~D., Asadi, K., and Zweig, G.
\newblock Hybrid code networks: practical and efficient end-to-end dialog
  control with supervised and reinforcement learning.
\newblock In \emph{Association for Computational Linguistics}, 2017.

\bibitem[Zhang et~al.(2021)Zhang, Yao, and
  Whiteson]{zhang_breaking_deadly_triad}
Zhang, S., Yao, H., and Whiteson, S.
\newblock Breaking the deadly triad with a target network.
\newblock \emph{arXiv}, 2021.

\end{thebibliography}
\bibliographystyle{icml2023}
\newpage
\onecolumn
\section{Appendix}
\subsection{Pseudo-code for DQN Pro}
Below, we present the pseudo-code for DQN Pro. Notice that the difference between DQN and DQN Pro is minimal (highlighted in gray).
\begin{algorithm}[H]
\caption{DQN with Proximal Iteration (DQN Pro)}
\label{alg:DQNPro}
\begin{algorithmic}[1]
\State Initialize $\theta$, N, \textit{period}, replay buffer $\mathcal{D}, \alpha,$ and ${\boldsymbol{\tilde c}}$
\State $s \leftarrow $ env.reset(), $w\leftarrow \theta$, $\textrm{numUpdates} \leftarrow 0$
\Repeat

        \State $a \sim \epsilon$-greedy\big($ Q(s,\cdot;w$)\big)
        \State $s', r \leftarrow\ $env.step($s, a$) 
        \State  add $\langle s, a, r, s'\rangle$ to $\mathcal{D}$
        \If {$s'$ is terminal}
        \State $s \leftarrow $ env.reset()
        \EndIf        
        \For{$n$ in \{1, \dots, N\}} 
        \State sample $\mathcal{B} = \{\langle s, a, r, s'\rangle\}$, compute $\nabla_{w} h(w)$
        \State \colorbox{lightgray}{$w\!\leftarrow\!{ \big(1-(\alpha/\tilde c)\big) w +(\alpha/\tilde c) \theta}- \alpha \nabla_w h(w) $}
        \State $\textrm{numUpdates} \leftarrow \textrm{numUpdates} +1$
        \If {$ \textrm{numUpdates}\ \%\ \textit{period} = 0$} 
        \State $\theta \leftarrow w$
        \EndIf
        \EndFor{}
\Until{convergence}
\end{algorithmic}
\end{algorithm}
\subsection{Implementation Details}\label{sec:app:hyperparams}
Table \ref{tab:table-exp-hypers} and \ref{tab:table-compute-hypers} show hyper-parameters, computing infrastructure, and libraries used for the experiments in this paper for all games tested. Our training and evaluation protocols and the hyper-parameter settings closely follow those of the Dopamine baseline. To report performance results, we measured the undiscounted sum of rewards obtained by the learned policy during evaluation. 
\begin{table*}
    \centering
    \begin{tabular}{|l|c|}
        \hline
        \multicolumn{2}{|c|}{\textbf{DQN hyper-parameters (shared)}} \\ \hline
        Replay buffer size & $200000$ \\
        Target update period & $8000$ \\
        Max steps per episode & $27000$ \\
        Evaluation  frequency & $10000$ \\
        Batch size & $64$ \\
        Update period & $4$ \\
        Number of frame skip & $4$ \\
        Number of episodes to evaluate & $2$ \\
        Update horizon & $1$ \\
        $\epsilon$-greedy (training time)  & $0.01$ \\
        $\epsilon$-greedy (evaluation time)  & $0.001$ \\
        $\epsilon$-greedy decay period  & $250000$ \\
        Burn-in period / Min replay size & $20000$ \\
        Learning rate & $10^{-4}$ \\
        Discount factor ($\gamma$) & $0.99$ \\
        Total number of iterations & $3 \times 10^{7}$  \\
        Sticky actions & True \\
        Optimizer & Adam~\cite{KingmaB14}\\
        Network architecture & Nature DQN network~\cite{mnih2015human} \\
        Random seeds & $\{0,1,2,3,4\}$ 
        \\ \hline
        \multicolumn{2}{|c|}{\textbf{Rainbow hyper-parameters (shared)}} \\ \hline
        Batch size & 64 \\ 
        Other &  Config file \href{https://github.com/google/dopamine/blob/master/dopamine/agents/rainbow/configs/rainbow_aaai.gin}{rainbow\textunderscore aaai.gin} from Dopamine \\ \hline
        \multicolumn{2}{|c|}{\textbf{DQN Pro and Rainbow Pro hyper-parameter}} \\ \hline
        $\tilde c\ \textrm{(DQN Pro)}$ & $ 0.2$ \\
        \hline
        $\tilde c\ \textrm{(Rainbow Pro)}$ & $ 0.05$ \\
        \hline
   \end{tabular}
    \caption{Hyper-parameters used for all methods for all 55 games of Atari-2600 benchmarks}. All results reported in our paper are averages over repeated runs initialized with each of the random seeds listed above and run for the listed number of episodes.
    
    \label{tab:table-exp-hypers}
\end{table*}

\begin{table*}
    \centering
    \begin{tabular}{|l|c|}
        \hline
        \multicolumn{2}{|c|}{\textbf{Computing Infrastructure}} \\ \hline
        Machine Type & AWS EC2 - p2.16xlarge\\
        GPU Family & Tesla K80 \\
        CPU Family & Intel Xeon 2.30GHz \\
        CUDA Version & $11.0$ \\
        NVIDIA-Driver & $450.80.02$ \\  \hline
       \multicolumn{2}{|c|}{\textbf{Library Version}} \\ \hline
        Python & $3.8.5$ \\
        Numpy &  $1.20.1$ \\
        Gym & $0.18.0$ \\
        Pytorch & $1.8.0$ \\
        \hline
   \end{tabular}
    \caption{Computing infrastructure and software libraries used in all experiments in this paper.}
    \label{tab:table-compute-hypers}
\end{table*}
\newpage
\subsection{Proofs}
\contraction*
We make two assumptions:
\begin{enumerate}
    \item $f$ is smooth, or more specifically that its gradient is 1-Lipschitz: $||\nabla f(v_1) -\nabla f(v_2)||\leq ||v_1-v_2||\ \forall v_1,\forall v_2$ .
    \item the value of the parameter $c$ is large, in particular $c>\frac{2}{1-\gamma}\ $.
\end{enumerate}
\begin{proof}
Both terms are convex and differentiable, therefore by setting the gradient to zero, we have:
$$\mathcal{T}^{\star}_{c, f}v = \mathcal{T}^{\star}v + \frac{1}{c}\big(\nabla f(v) - \nabla f(\mathcal{T}^{\star}_{c, f}v)\big), $$
We can then show:
\begin{eqnarray*}
    || \mathcal{T}^{\star}_{c, f}v_1 - \mathcal{T}^{\star}_{c, f}v_2|| &=& || T^{\star} v_1 + \frac{1}{c}\big(\nabla f(v_1) - \nabla f(\mathcal{T}^{\star}_{c, f}v_1)\big) - T^{\star}v_2 - \frac{1}{c}\big(\nabla f(v_2) -\nabla f(\mathcal{T}^{\star}_{c, f}v_2)\big) ||\\
    & \leq & || T^{\star} v_1 - T^{\star}v_2|| + \frac{1}{c}|| \nabla f(v_1)- \nabla f(v_2)|| + \frac{1}{c}|| \nabla f(\mathcal{T}^{\star}_{c, f}v_1)  - \nabla f(\mathcal{T}^{\star}_{c, f}v_2) ||\\
    && \textrm{(first assumption)}\\
    &\leq& || T^{\star}v_1 - T^{\star}v_2|| + \frac{1}{c}|| \nabla f(v_1)- \nabla f(v_2)|| + \frac{1}{c}||\mathcal{T}^{\star}_{c, f}v_1 - \mathcal{T}^{\star}_{c, f}v_2 ||
\end{eqnarray*}
This implies:
\begin{eqnarray*}
\frac{c-1}{c} ||\mathcal{T}^{\star}_{c, f}v_1 - \mathcal{T}^{\star}_{c, f}v_2|| &\leq& || T^{\star}v_1 - T^{\star}v_2|| + \frac{1}{c}|| \nabla f(v_1)- \nabla f(v_2)||\\
&& \textrm{(first assumption)}\\ 
&\leq & || T^{\star}v_1 - T^{\star}v_2|| + \frac{1}{c}|| v_1- v_2||\\
&\leq&\frac{\gamma c+1}{c} || v_1 - v_2||
\end{eqnarray*}
Therefore,
\begin{eqnarray*}
|| \mathcal{T}^{\star}_{c, f}v_1 - \mathcal{T}^{\star}_{c, f}v_2|| &\leq & 
 \frac{\gamma c + 1}{c-1} ||v_1 - v_2 ||\ ,
\end{eqnarray*}
Allowing us to conclude that $\mathcal{T}^{\star}_{c, f}$ is a contraction~(second assumption).

Further, to show that $v^{\star}$ is indeed the fixed point of $\mathcal{T}^{\star}_{c, f}$, notice from the original formulation:
\begin{equation*}
    \mathcal{T}^{\star}_{c, f} v \myeq\arg\min_{v'} ||v'- \mathcal{T}^{\star} v||^{2}_{2} +\frac{1}{c} D_{f}(v',v)\ ,
\end{equation*}
that, at point $v^{*}$ setting $v'=v^{\star}$ jointly minimizes the first term, because $v^{\star} = \mathcal{T}^{\star}v^{\star}$ due to fixed-point defintion, and it also minimizes the second term because $D_{f}(v^{\star},v^{\star}) = 0$ and that Bregman divergence is non-negative. Therefore, $\mathcal{T}^{\star}_{c, f} v^{\star} = v^{\star}$; $v^{\star}$ is the fixed-point of $\mathcal{T}^{\star}_{c, f} v^{\star}$. Since, $\mathcal{T}^{\star}_{c, f}$ is a contraction, this fixed point is unique.
\end{proof}
\error*
We make two assumptions:
\begin{enumerate}
    \item we assume $\epsilon$ error in policy evaluation step, as already stated in equation (4).
    \item we assume $\epsilon'$ error in policy greedification step $\pi_{k} \leftarrow \mathcal{G}_{\epsilon'_{k}}v_{k-1}\forall k$. This means $\forall \pi\ \mathcal{T}^{\pi} v_k - \mathcal{T}^{\pi_{k+1}} v_k\leq \epsilon'_{k+1}$. Note that this assumption is orthogonal to the thesis of our paper, but we kept it for generality.
\end{enumerate}
\begin{proof}Step 0: bound the Bellman residual: $b_{k} \myeq v_{k} - \mathcal{T}^{\pi_{k+1}} v_k \ .$
\begin{eqnarray*}
b_{k} &=& v_{k} -  \mathcal{T}^{\pi_{k+1}} v_k \\
&=& v_{k} - \mathcal{T}^{\pi_{k}} v_k+ \mathcal{T}^{\pi_{k}} v_k - \mathcal{T}^{\pi_{k+1}} v_k\\
&& \textrm{(from our assumption $\quad \forall \pi\ \mathcal{T}^{\pi} v_k - \mathcal{T}^{\pi_{k+1}} v_k\leq \epsilon'_{k+1}$)}\\
& \leq & v_{k} - \mathcal{T}^{\pi_{k}} v_k+ \epsilon'_{k+1}\\
& = & v_{k} - (1-\beta) \epsilon_k - \mathcal{T}^{\pi_{k}} v_k+(1-\beta)\gamma P^{\pi_{k}}\epsilon_{k} +(1-\beta) \epsilon_k -(1-\beta)\gamma P^{\pi_{k}}\epsilon_{k} +\epsilon'_{k+1}\\
&& \textrm{\Big(from $\mathcal{T}^{\pi_{k}}v_k+(1-\beta)\gamma P^{\pi_{k}}\epsilon_{k}=\mathcal{T}^{\pi_{k}}(v_k-(1-\beta)\epsilon_k\big)$\Big)}\\
& = & v_{k} - (1-\beta) \epsilon_k - \mathcal{T}^{\pi_{k}}(v_k-(1-\beta)\epsilon_k)+(1-\beta) \underbrace{(I-\gamma P_{\pi_{k}})\epsilon_k }_{x_k}+\epsilon'_{k+1}\\
& = & v_{k} - (1-\beta) \epsilon_k - \mathcal{T}^{\pi_{k}}(v_k-(1-\beta)\epsilon_k)+(1-\beta)x_k + \epsilon'_{k+1}\\
&& \textrm{(from $v_k - (1-\beta)\epsilon_k = (\mathcal{T}^{\pi_k}_{\beta})^{n} v_{k-1}$)}\\
& = & (1-\beta) (\mathcal{T}^{\pi_{k}})^n v_{k-1}+\beta v_{k-1} - \mathcal{T}^{\pi_{k}}\big(  (1-\beta) (\mathcal{T}^{\pi_{k}})^n v_{k-1}+\beta v_{k-1} \big) +(1-\beta)x_k + \epsilon'_{k+1}\\
&& \textrm{(from linearity of $\mathcal{T}^{\pi_{k}}$)}\\
& = & (1-\beta) (\mathcal{T}^{\pi_{k}})^n v_{k-1} - \mathcal{T}^{\pi_{k}}\big(  (1-\beta) (\mathcal{T}^{\pi_{k}})^n v_{k-1}\big)+\beta \big(v_{k-1} - \mathcal{T}^{\pi_{k}} v_{k-1}\big) +(1-\beta)x_k + \epsilon'_{k+1}\\
& = & (1-\beta) \Big((\mathcal{T}^{\pi_{k}})^n v_{k-1} - \mathcal{T}^{\pi_{k}}\big( (T ^{\pi_{k}})^n v_{k-1}\big)\Big)+\beta \big(v_{k-1} - \mathcal{T}^{\pi_{k}}v_{k-1}\big) +(1-\beta)x_k + \epsilon'_{k+1}\\
& = & (1-\beta) \Big((\mathcal{T}^{\pi_{k}})^n v_{k-1} - (\mathcal{T}^{\pi_{k}})^n\big( \mathcal{T}^{\pi_{k}} v_{k-1}\big)\Big)+\beta \big(v_{k-1} - \mathcal{T}^{\pi_{k}} v_{k-1}\big) +(1-\beta)x_k + \epsilon'_{k+1}\\
& = & (1-\beta)(\gamma P^{\pi_k})^{n} \big(\underbrace{v_{k-1} - \mathcal{T}^{\pi_{k}}(v_{k-1}}_{=b_{k-1}})\big)+\beta \big(\underbrace{v_{k-1} - \mathcal{T}^{\pi_{k}}v_{k-1}}_{=b_{k-1}}\big) +(1-\beta)x_k + \epsilon'_{k+1} \ ,
\end{eqnarray*}
allowing us to conclude:
$$b_{k} = \big((1-\beta)(\gamma P^{\pi_k})^{n} + \beta I\big) b_{k-1}+ (1-\beta) x_{k} + \epsilon'_{k+1} \ . $$
\\
Step 1: bound the distance to the optimal value: $d_{k+1} \myeq v^{*}-(\mathcal{T}^{\pi_{k+1}}_{\beta})^n v_k\ $.
\begin{eqnarray*}
d_{k+1} &=& v^{*}-(\mathcal{T}^{\pi_{k+1}}_{\beta})^n v_k\\
&=& \mathcal{T}^{\pi^{*}}v^{*}-\mathcal{T}^{\pi^{*}}v_{k}+\underbrace{\mathcal{T}^{\pi^{*}}v_{k}-\mathcal{T}^{\pi_{k+1}}v_{k}}_{\leq \epsilon'_{k+1}}+\underbrace{\mathcal{T}^{\pi_{k+1}}v_{k}-(\mathcal{T}^{\pi_{k+1}}_{\beta})^n v_k}_{= g_{k+1}}\\
&\leq & \gamma P^{\pi^{*}}(v^{*}-v_{k})+\epsilon'_{k+1}+g_{k+1}\\
&=& \gamma P^{\pi^{*}}(v^{*}-v_{k})+(1-\beta)\gamma P^{\pi^*} \epsilon_{k}-(1-\beta)\gamma P^{\pi^*} \epsilon_{k}+\epsilon'_{k+1}+g_{k+1}\\
&=&\gamma P^{\pi^{*}}\Big(v^{*}-\big(v_{k}-(1-\beta)\epsilon_k\big)\Big)-(1-\beta)\underbrace{\gamma P^{\pi^*}\epsilon_{k}}_{y_k} +\epsilon'_{k+1}+g_{k+1}\\
&=& \gamma P^{\pi^{*}}\big(\underbrace{v^{*}-(\mathcal{T}^{\pi_{k}}_{\beta})^n v_{k-1}}_{=d_{k}}\big)-(1-\beta)y_k+\epsilon'_{k+1}+g_{k+1}\\
&=& \gamma P^{\pi^{*}}d_{k}-(1-\beta)y_k + \epsilon'_{k+1}+g_{k+1}
\end{eqnarray*}
Additionally we can bound $g_{k+1}$ as follows:
\begin{eqnarray*}
g_{k+1} &=& \mathcal{T}^{\pi_{k+1}} v_{k}-(\mathcal{T}^{\pi_{k+1}}_{\beta})^n v_k\\
&=& (1-\beta)\big(\mathcal{T}^{\pi_{k+1}}v_{k}-(\mathcal{T}^{\pi^{k+1}})^n v_k \big) + \beta (T^{\pi_{k+1}} v_{k} - v_{k})\\
&=& (1-\beta)\sum_{j=1}^{n-1}(\gamma P^{\pi_{k+1}})^j b_{k} + \beta(-b_{k})\\
\end{eqnarray*}
Allowing us to conclude that:
$$d_{k+1} \leq \gamma P^{\pi^{*}}d_{k} -\big((1-\beta)y_{k} + \beta b_k\big) + (1-\beta)\sum_{j=1}^{n-1}(\gamma P^{\pi_{k+1}})^j b_{k} + \epsilon'_{k+1}$$
Step 2: bound the distance between the approximate value and the value of the policy: $s_k\myeq (T^{\pi_{k}}_{\beta})^n v_{k-1} - v^{\pi_k}\ $.
\begin{eqnarray*}
s_k &=&  (T^{\pi_{k}}_{\beta})^n v_{k-1} - v^{\pi_k} \\ 
&=& (T^{\pi_{k}}_{\beta})^n v_{k-1} - (\mathcal{T}^{\pi_{k}})^{\infty} v_{k-1}\\
&=& (1-\beta)(\mathcal{T}^{\pi_{k}})^n v_{k-1} + \beta v_{k-1} -(1-\beta)(\mathcal{T}^{\pi_{k}})^{\infty} v_{k-1} -\beta (\mathcal{T}^{\pi_{k}})^{\infty}v_{k-1}\\
&=& (1-\beta)\big((\mathcal{T}^{\pi_{k}})^n v_{k-1}-(\mathcal{T}^{\pi_{k}})^{\infty}v_{k-1}\big) + \beta\big(v_{k-1}-(\mathcal{T}^{\pi_{k}})^{\infty}v_{k-1}\big)\\
&=& (1-\beta)(\gamma P^{\pi_k})^n\big(v_{k-1}-(\mathcal{T}^{\pi_{k}})^{\infty} v_{k-1}\big) + \beta\big(v_{k-1}-(\mathcal{T}^{\pi_{k}})^{\infty} v_{k-1}\big)\\
&=&\big((1-\beta)(\gamma P^{\pi_k})^n + \beta I\big)\big(v_{k-1}-(\mathcal{T}^{\pi_{k}})^{\infty}v_{k-1}\big)\\
&=&\big((1-\beta)(\gamma P^{\pi_k})^n + \beta I\big)(I-\gamma P^{\pi_k})^{-1}\big(\underbrace{v_{k-1} - \mathcal{T}^{\pi_k}v_{k-1}}_{b_{k-1}}\big)\ .
\end{eqnarray*}
Allowing us to conclude that:
$$s_k = \big((1-\beta)(\gamma P^{\pi_k})^n + \beta I\big)(I-\gamma P^{\pi_k})^{-1}b_{k-1} \ .$$
\end{proof}
\newpage
\section{Learning curves}
We present full learning curves of DQN, DQN Pro, Rainbow, and Rainbow Pro for the 55 Atari games. All results are averaged over 5 independent seeds.
\begin{figure}
\centering\captionsetup[subfigure]{justification=centering}
\begin{subfigure}[t]{ 0.2\textwidth} 
\centering 
\includegraphics[width=\textwidth]{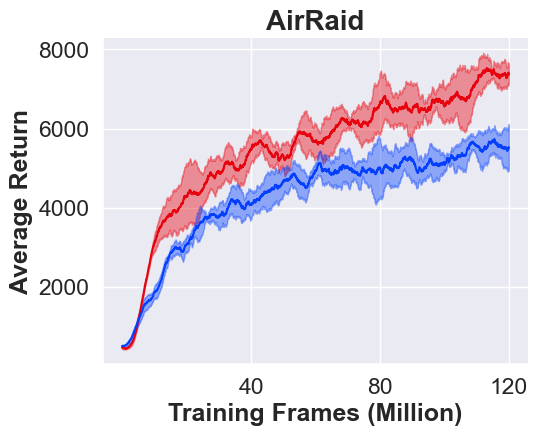} 
\label{fig:4_learning_curves_appendix/AirRaidNoFrameskip-v0_4_learning_curves_appendix.png} 
\end{subfigure}%
~ 
\begin{subfigure}[t]{ 0.2\textwidth} 
\centering 
\includegraphics[width=\textwidth]{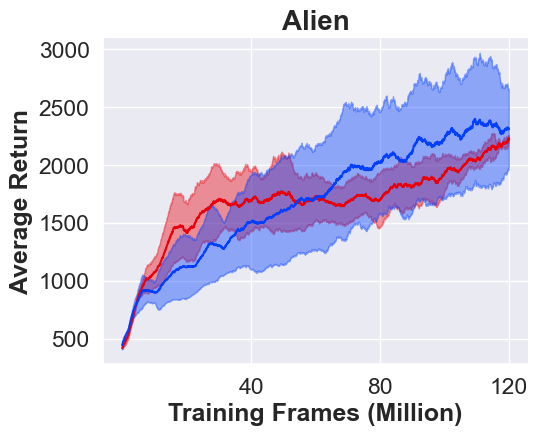} 
\label{fig:4_learning_curves_appendix/AlienNoFrameskip-v0_4_learning_curves_appendix.png} 
\end{subfigure}%
~ 
\begin{subfigure}[t]{ 0.2\textwidth} 
\centering 
\includegraphics[width=\textwidth]{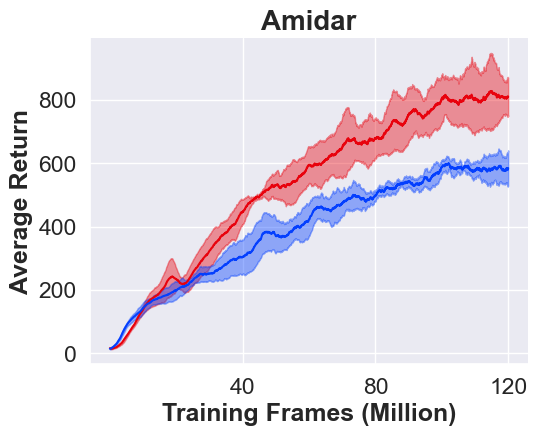} 
\label{fig:4_learning_curves_appendix/AmidarNoFrameskip-v0_4_learning_curves_appendix.png} 
\end{subfigure}%
~ 
\begin{subfigure}[t]{ 0.2\textwidth} 
\centering 
\includegraphics[width=\textwidth]{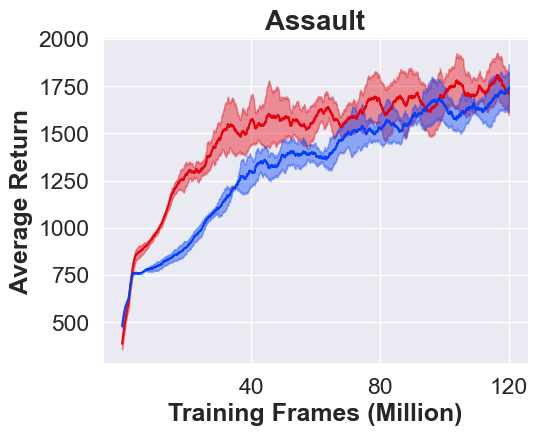} 
\label{fig:4_learning_curves_appendix/AssaultNoFrameskip-v0_4_learning_curves_appendix.png} 
\end{subfigure}%
~ 
\begin{subfigure}[t]{ 0.2\textwidth} 
\centering 
\includegraphics[width=\textwidth]{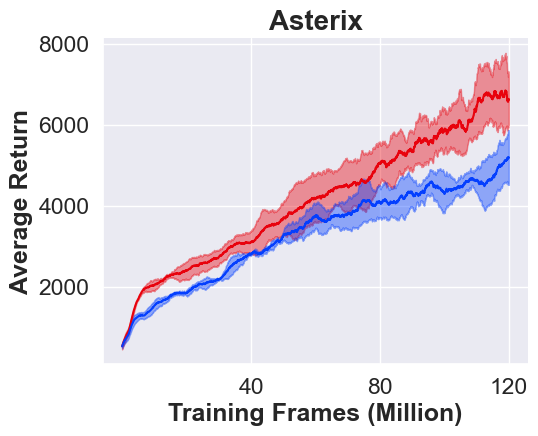} 
\label{fig:4_learning_curves_appendix/AsterixNoFrameskip-v0_4_learning_curves_appendix.png} 
\end{subfigure}%

\begin{subfigure}[t]{ 0.2\textwidth} 
\centering 
\includegraphics[width=\textwidth]{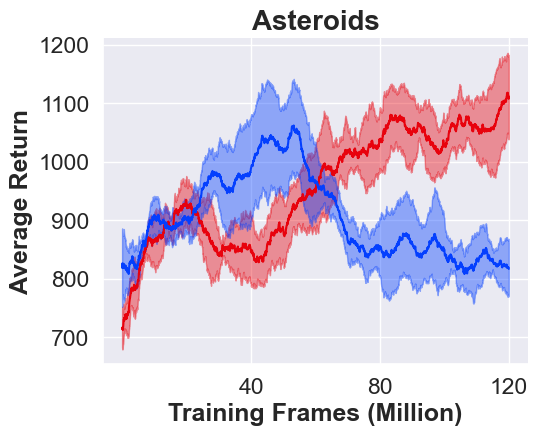} 
\label{fig:4_learning_curves_appendix/AsteroidsNoFrameskip-v0_4_learning_curves_appendix.png} 
\end{subfigure}%
~ 
\begin{subfigure}[t]{ 0.2\textwidth} 
\centering 
\includegraphics[width=\textwidth]{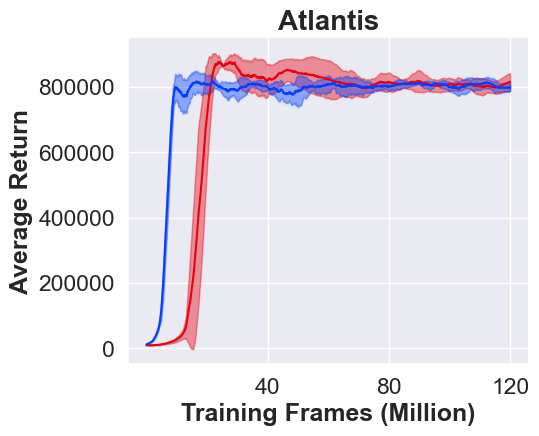} 
\label{fig:4_learning_curves_appendix/AtlantisNoFrameskip-v0_4_learning_curves_appendix.png} 
\end{subfigure}%
~ 
\begin{subfigure}[t]{ 0.2\textwidth} 
\centering 
\includegraphics[width=\textwidth]{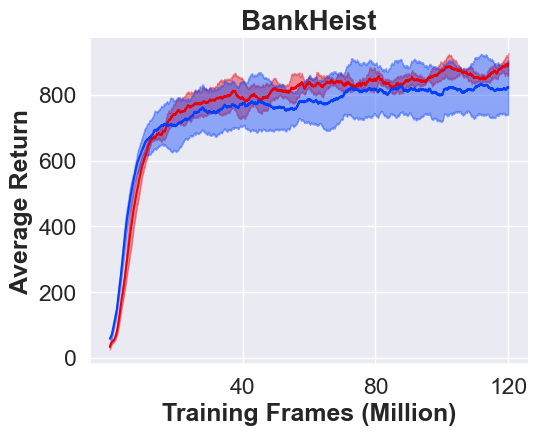} 
\label{fig:4_learning_curves_appendix/BankHeistNoFrameskip-v0_4_learning_curves_appendix.png} 
\end{subfigure}%
~ 
\begin{subfigure}[t]{ 0.2\textwidth} 
\centering 
\includegraphics[width=\textwidth]{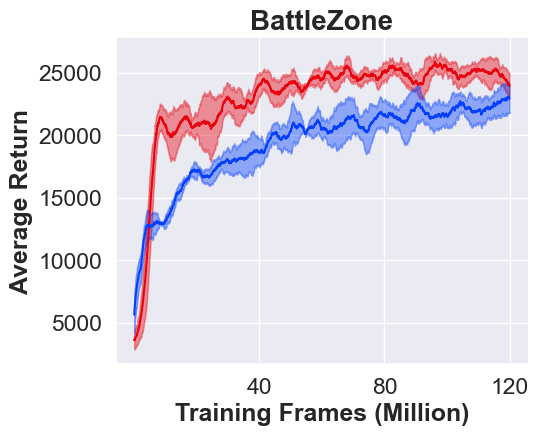} 
\label{fig:4_learning_curves_appendix/BattleZoneNoFrameskip-v0_4_learning_curves_appendix.png} 
\end{subfigure}%
~ 
\begin{subfigure}[t]{ 0.2\textwidth} 
\centering 
\includegraphics[width=\textwidth]{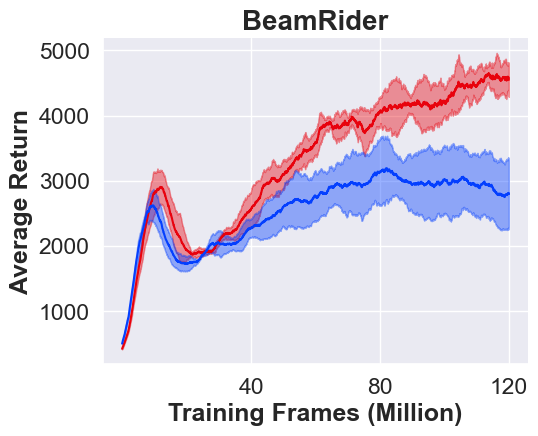} 
\label{fig:4_learning_curves_appendix/BeamRiderNoFrameskip-v0_4_learning_curves_appendix.png} 
\end{subfigure}%

\begin{subfigure}[t]{ 0.2\textwidth} 
\centering 
\includegraphics[width=\textwidth]{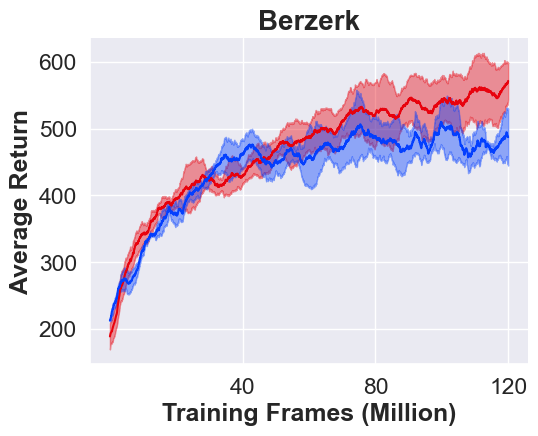} 
\label{fig:4_learning_curves_appendix/BerzerkNoFrameskip-v0_4_learning_curves_appendix.png} 
\end{subfigure}%
~ 
\begin{subfigure}[t]{ 0.2\textwidth} 
\centering 
\includegraphics[width=\textwidth]{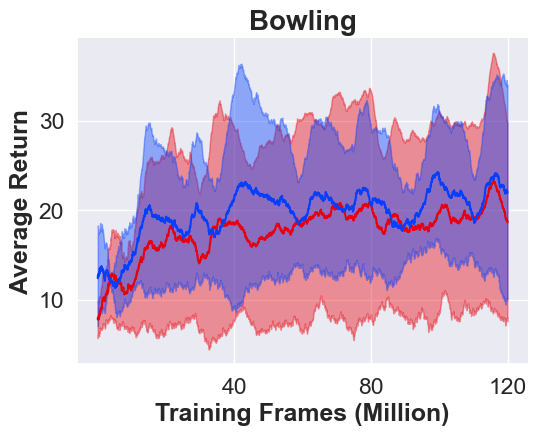} 
\label{fig:4_learning_curves_appendix/BowlingNoFrameskip-v0_4_learning_curves_appendix.png} 
\end{subfigure}%
~ 
\begin{subfigure}[t]{ 0.2\textwidth} 
\centering 
\includegraphics[width=\textwidth]{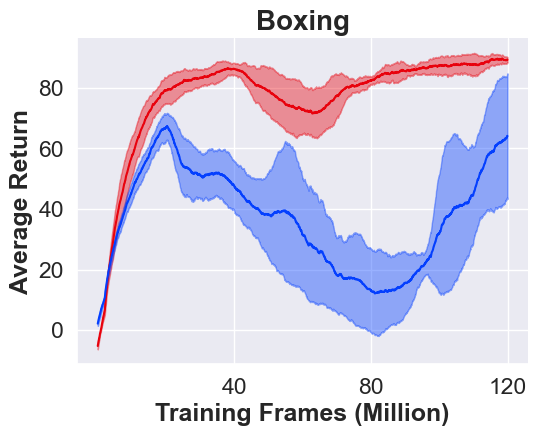} 
\label{fig:4_learning_curves_appendix/BoxingNoFrameskip-v0_4_learning_curves_appendix.png} 
\end{subfigure}%
~ 
\begin{subfigure}[t]{ 0.2\textwidth} 
\centering 
\includegraphics[width=\textwidth]{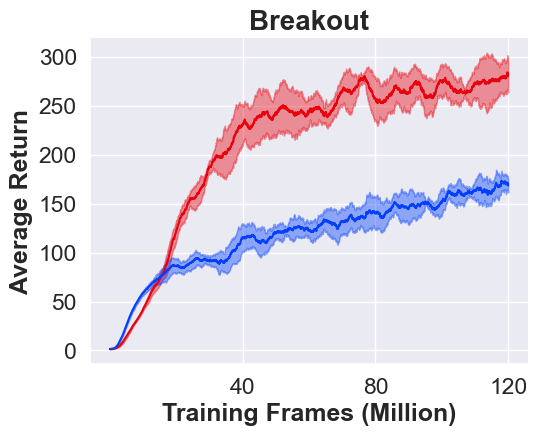} 
\label{fig:4_learning_curves_appendix/BreakoutNoFrameskip-v0_4_learning_curves_appendix.png} 
\end{subfigure}%
~ 
\begin{subfigure}[t]{ 0.2\textwidth} 
\centering 
\includegraphics[width=\textwidth]{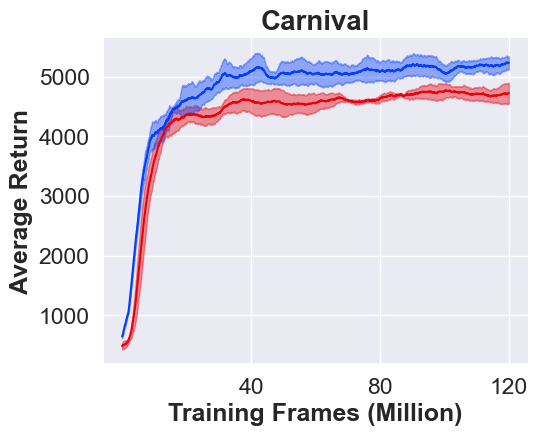} 
\label{fig:4_learning_curves_appendix/CarnivalNoFrameskip-v0_4_learning_curves_appendix.png} 
\end{subfigure}%

\begin{subfigure}[t]{ 0.2\textwidth} 
\centering 
\includegraphics[width=\textwidth]{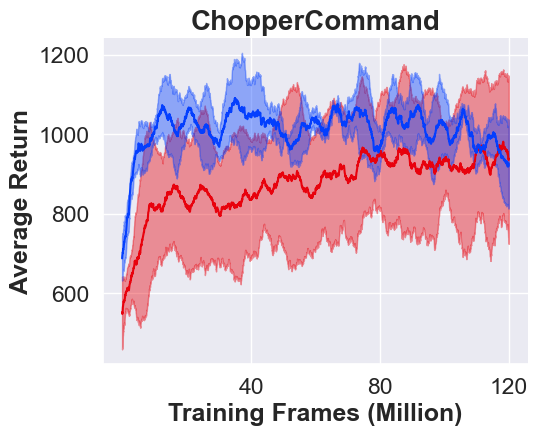} 
\label{fig:4_learning_curves_appendix/ChopperCommandNoFrameskip-v0_4_learning_curves_appendix.png} 
\end{subfigure}%
~ 
\begin{subfigure}[t]{ 0.2\textwidth} 
\centering 
\includegraphics[width=\textwidth]{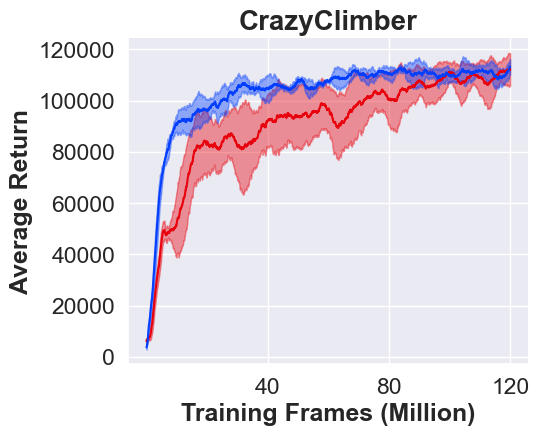} 
\label{fig:4_learning_curves_appendix/CrazyClimberNoFrameskip-v0_4_learning_curves_appendix.png} 
\end{subfigure}%
~ 
\begin{subfigure}[t]{ 0.2\textwidth} 
\centering 
\includegraphics[width=\textwidth]{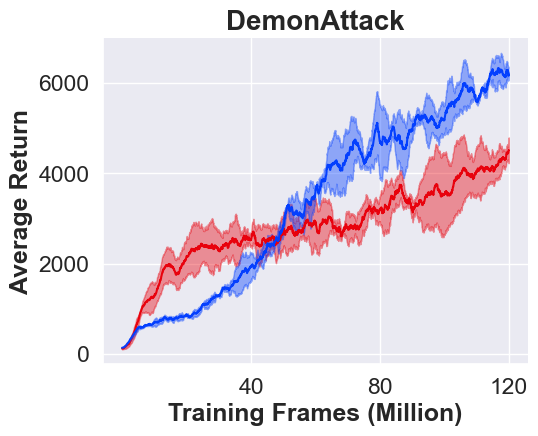} 
\label{fig:4_learning_curves_appendix/DemonAttackNoFrameskip-v0_4_learning_curves_appendix.png} 
\end{subfigure}%
~ 
\begin{subfigure}[t]{ 0.2\textwidth} 
\centering 
\includegraphics[width=\textwidth]{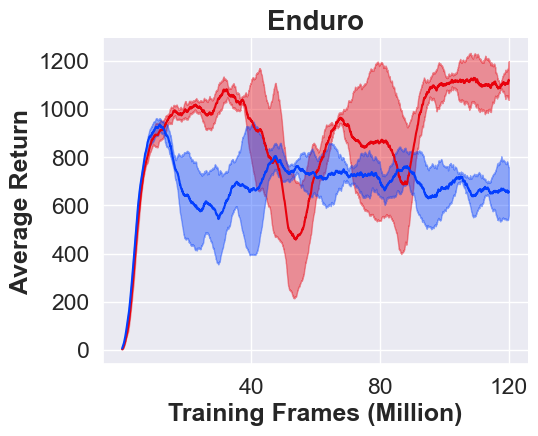} 
\label{fig:4_learning_curves_appendix/EnduroNoFrameskip-v0_4_learning_curves_appendix.png} 
\end{subfigure}%
~ 
\begin{subfigure}[t]{ 0.2\textwidth} 
\centering 
\includegraphics[width=\textwidth]{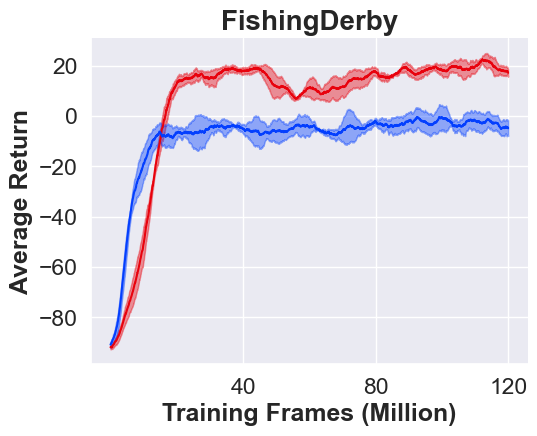} 
\label{fig:4_learning_curves_appendix/FishingDerbyNoFrameskip-v0_4_learning_curves_appendix.png} 
\end{subfigure}%

\begin{subfigure}[t]{ 0.2\textwidth} 
\centering 
\includegraphics[width=\textwidth]{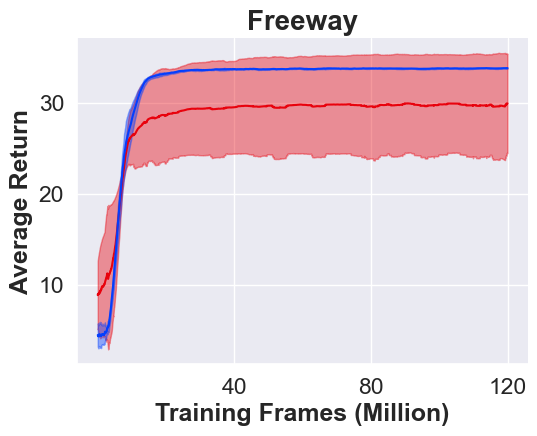} 
\label{fig:4_learning_curves_appendix/FreewayNoFrameskip-v0_4_learning_curves_appendix.png} 
\end{subfigure}%
~ 
\begin{subfigure}[t]{ 0.2\textwidth} 
\centering 
\includegraphics[width=\textwidth]{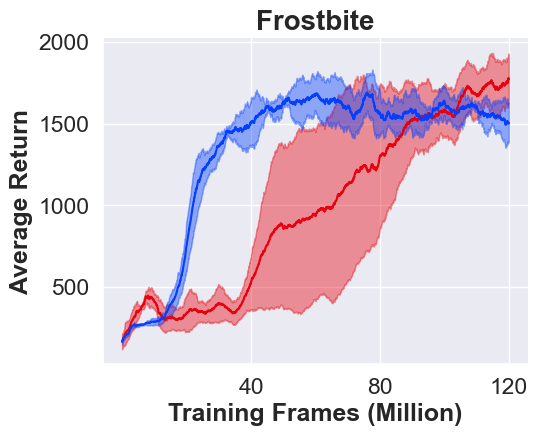} 
\label{fig:4_learning_curves_appendix/FrostbiteNoFrameskip-v0_4_learning_curves_appendix.png} 
\end{subfigure}%
~ 
\begin{subfigure}[t]{ 0.2\textwidth} 
\centering 
\includegraphics[width=\textwidth]{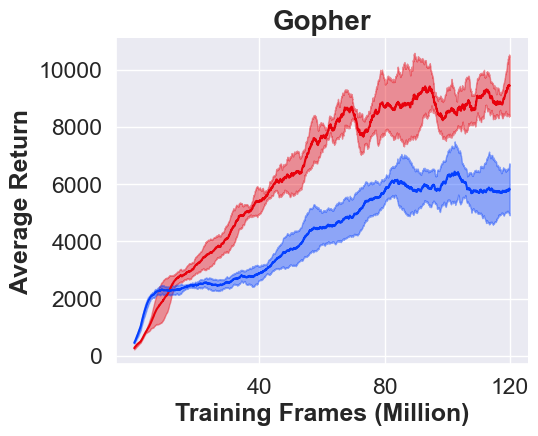} 
\label{fig:4_learning_curves_appendix/GopherNoFrameskip-v0_4_learning_curves_appendix.png} 
\end{subfigure}%
~ 
\begin{subfigure}[t]{ 0.2\textwidth} 
\centering 
\includegraphics[width=\textwidth]{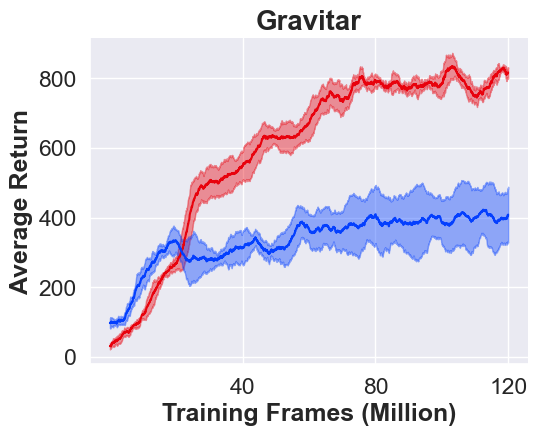} 
\label{fig:4_learning_curves_appendix/GravitarNoFrameskip-v0_4_learning_curves_appendix.png} 
\end{subfigure}%
~ 
\begin{subfigure}[t]{ 0.2\textwidth} 
\centering 
\includegraphics[width=\textwidth]{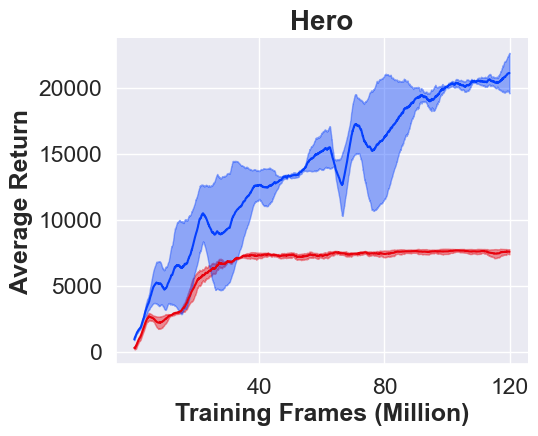} 
\label{fig:4_learning_curves_appendix/HeroNoFrameskip-v0_4_learning_curves_appendix.png} 
\end{subfigure}%

\begin{subfigure}[t]{ 0.2\textwidth} 
\centering 
\includegraphics[width=\textwidth]{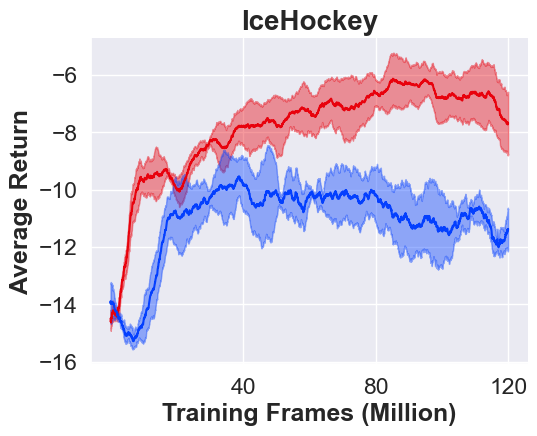} 
\label{fig:4_learning_curves_appendix/IceHockeyNoFrameskip-v0_4_learning_curves_appendix.png} 
\end{subfigure}%
~ 
\begin{subfigure}[t]{ 0.2\textwidth} 
\centering 
\includegraphics[width=\textwidth]{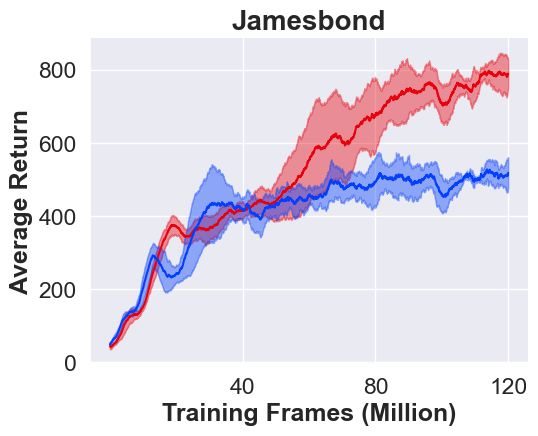} 
\label{fig:4_learning_curves_appendix/JamesbondNoFrameskip-v0_4_learning_curves_appendix.png} 
\end{subfigure}%
~ 
\begin{subfigure}[t]{ 0.2\textwidth} 
\centering 
\includegraphics[width=\textwidth]{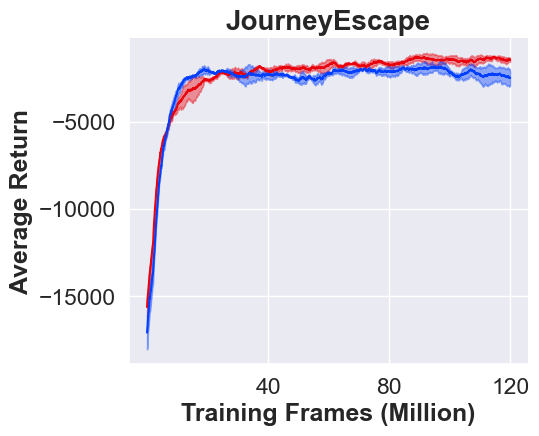} 
\label{fig:4_learning_curves_appendix/JourneyEscapeNoFrameskip-v0_4_learning_curves_appendix.png} 
\end{subfigure}%
~ 
\begin{subfigure}[t]{ 0.2\textwidth} 
\centering 
\includegraphics[width=\textwidth]{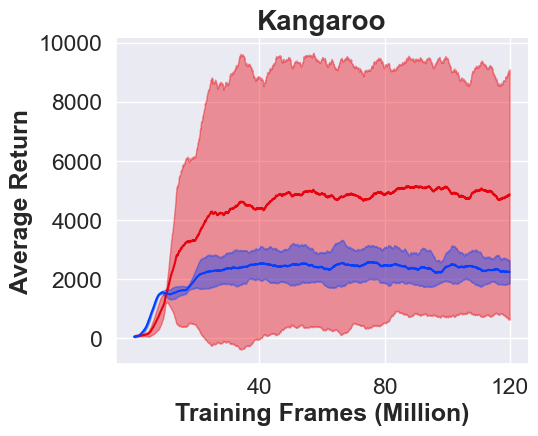} 
\label{fig:4_learning_curves_appendix/KangarooNoFrameskip-v0_4_learning_curves_appendix.png} 
\end{subfigure}%
~ 
\begin{subfigure}[t]{ 0.2\textwidth} 
\centering 
\includegraphics[width=\textwidth]{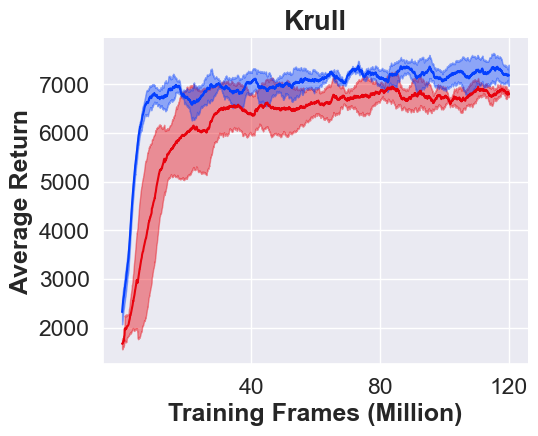} 
\label{fig:4_learning_curves_appendix/KrullNoFrameskip-v0_4_learning_curves_appendix.png} 
\end{subfigure}%
\caption{Comparison between DQN Pro (red) and DQN (blue) over 55 Atari games (Part I).}
\end{figure}
\begin{figure}
\begin{subfigure}[t]{ 0.2\textwidth} 
\centering 
\includegraphics[width=\textwidth]{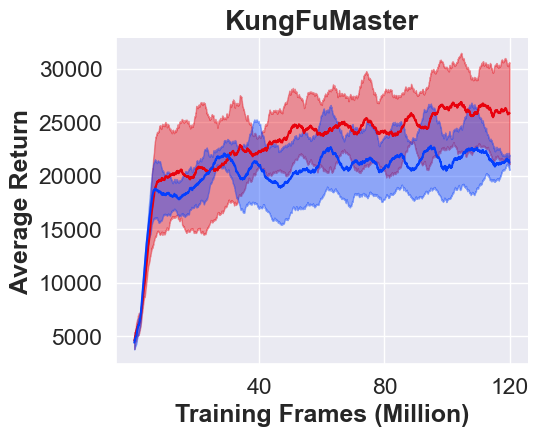} 
\label{fig:4_learning_curves_appendix/KungFuMasterNoFrameskip-v0_4_learning_curves_appendix.png} 
\end{subfigure}%
~ 
\begin{subfigure}[t]{ 0.2\textwidth} 
\centering 
\includegraphics[width=\textwidth]{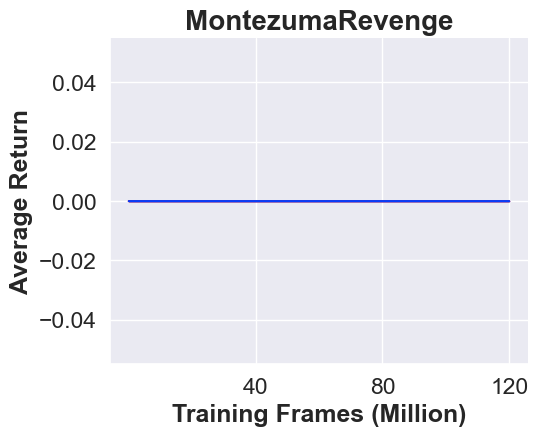} 
\label{fig:4_learning_curves_appendix/MontezumaRevengeNoFrameskip-v0_4_learning_curves_appendix.png} 
\end{subfigure}%
~ 
\begin{subfigure}[t]{ 0.2\textwidth} 
\centering 
\includegraphics[width=\textwidth]{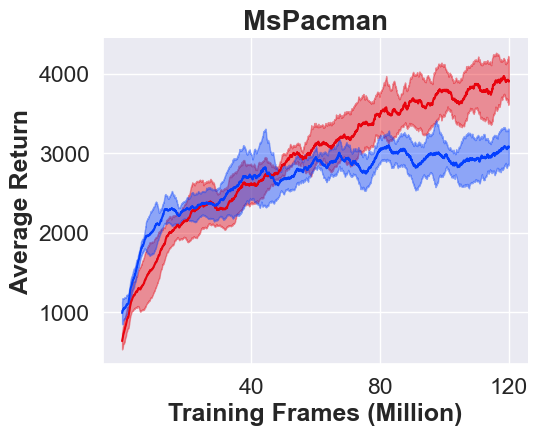} 
\label{fig:4_learning_curves_appendix/MsPacmanNoFrameskip-v0_4_learning_curves_appendix.png} 
\end{subfigure}%
~ 
\begin{subfigure}[t]{ 0.2\textwidth} 
\centering 
\includegraphics[width=\textwidth]{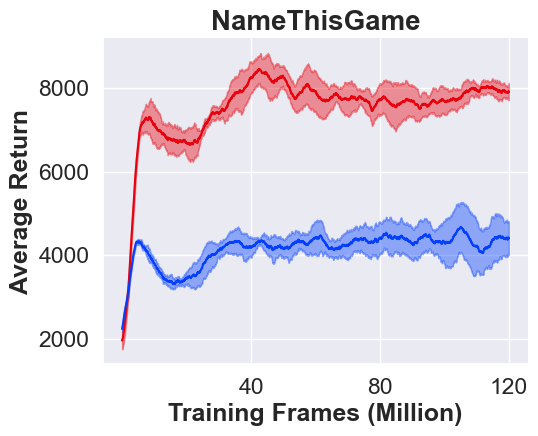} 
\label{fig:4_learning_curves_appendix/NameThisGameNoFrameskip-v0_4_learning_curves_appendix.png} 
\end{subfigure}%
~ 
\begin{subfigure}[t]{ 0.2\textwidth} 
\centering 
\includegraphics[width=\textwidth]{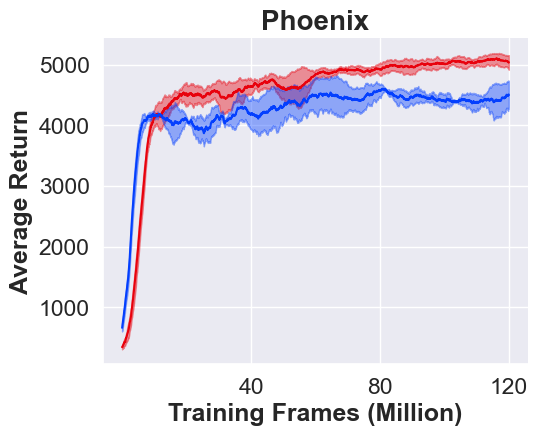} 
\label{fig:4_learning_curves_appendix/PhoenixNoFrameskip-v0_4_learning_curves_appendix.png} 
\end{subfigure}%

\begin{subfigure}[t]{ 0.2\textwidth} 
\centering 
\includegraphics[width=\textwidth]{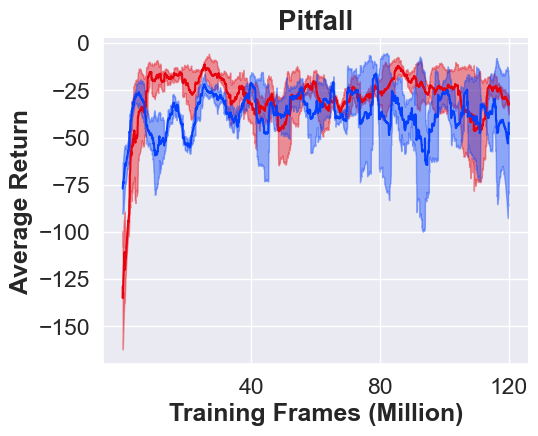} 
\label{fig:4_learning_curves_appendix/PitfallNoFrameskip-v0_4_learning_curves_appendix.png} 
\end{subfigure}%
~ 
\begin{subfigure}[t]{ 0.2\textwidth} 
\centering 
\includegraphics[width=\textwidth]{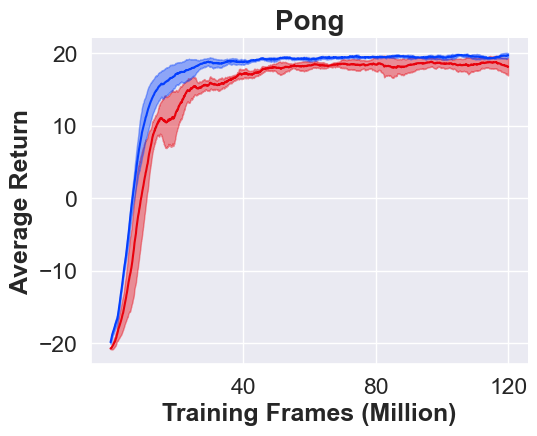} 
\label{fig:4_learning_curves_appendix/PongNoFrameskip-v0_4_learning_curves_appendix.png} 
\end{subfigure}%
~ 
\begin{subfigure}[t]{ 0.2\textwidth} 
\centering 
\includegraphics[width=\textwidth]{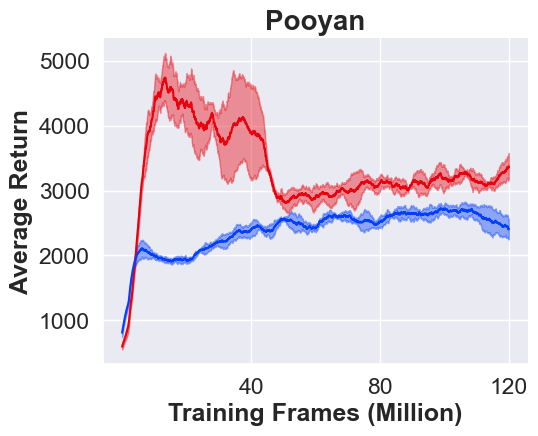} 
\label{fig:4_learning_curves_appendix/PooyanNoFrameskip-v0_4_learning_curves_appendix.png} 
\end{subfigure}%
~ 
\begin{subfigure}[t]{ 0.2\textwidth} 
\centering 
\includegraphics[width=\textwidth]{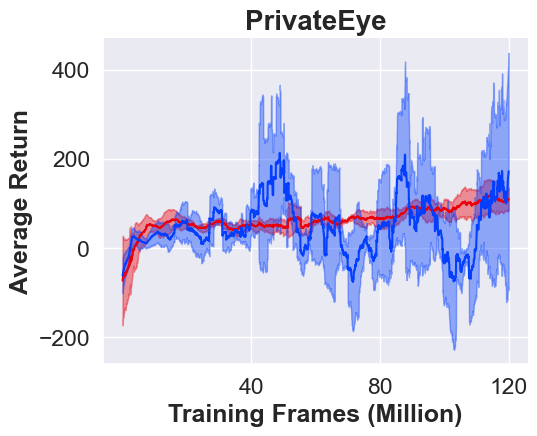} 
\label{fig:4_learning_curves_appendix/PrivateEyeNoFrameskip-v0_4_learning_curves_appendix.png} 
\end{subfigure}%
~ 
\begin{subfigure}[t]{ 0.2\textwidth} 
\centering 
\includegraphics[width=\textwidth]{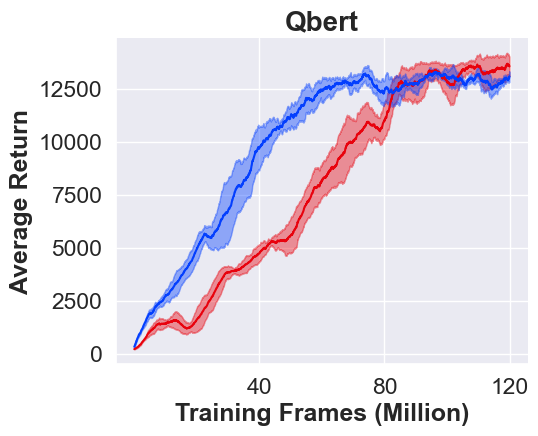} 
\label{fig:4_learning_curves_appendix/QbertNoFrameskip-v0_4_learning_curves_appendix.png} 
\end{subfigure}%

\begin{subfigure}[t]{ 0.2\textwidth} 
\centering 
\includegraphics[width=\textwidth]{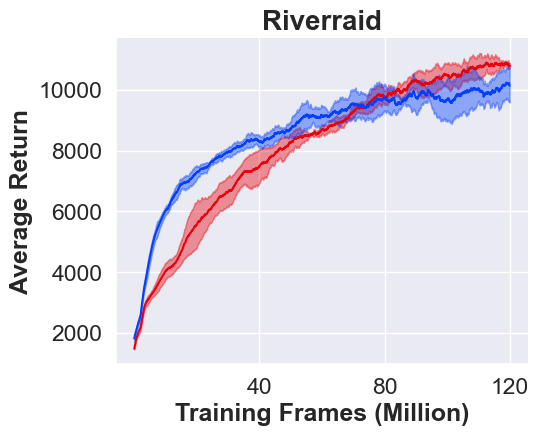} 
\label{fig:4_learning_curves_appendix/RiverraidNoFrameskip-v0_4_learning_curves_appendix.png} 
\end{subfigure}%
~ 
\begin{subfigure}[t]{ 0.2\textwidth} 
\centering 
\includegraphics[width=\textwidth]{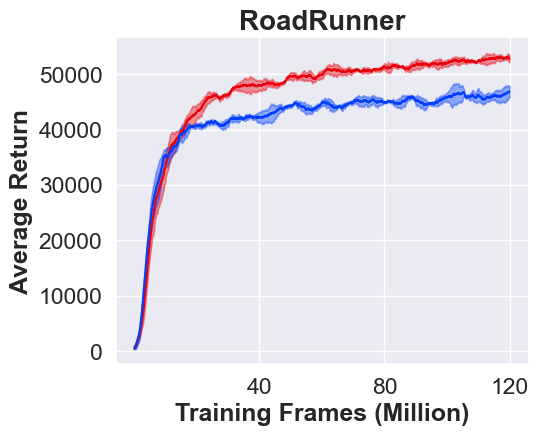} 
\label{fig:4_learning_curves_appendix/RoadRunnerNoFrameskip-v0_4_learning_curves_appendix.png} 
\end{subfigure}%
~ 
\begin{subfigure}[t]{ 0.2\textwidth} 
\centering 
\includegraphics[width=\textwidth]{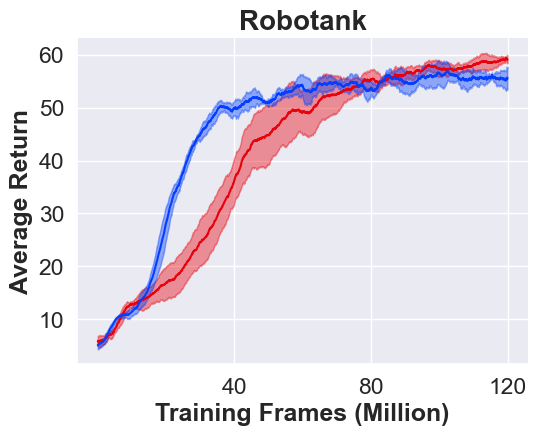} 
\label{fig:4_learning_curves_appendix/RobotankNoFrameskip-v0_4_learning_curves_appendix.png} 
\end{subfigure}%
~ 
\begin{subfigure}[t]{ 0.2\textwidth} 
\centering 
\includegraphics[width=\textwidth]{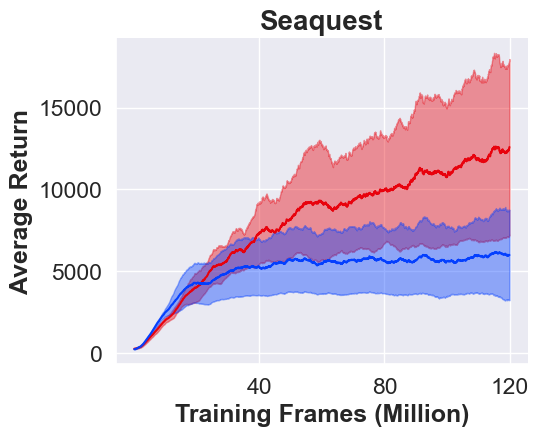} 
\label{fig:4_learning_curves_appendix/SeaquestNoFrameskip-v0_4_learning_curves_appendix.png} 
\end{subfigure}%
~ 
\begin{subfigure}[t]{ 0.2\textwidth} 
\centering 
\includegraphics[width=\textwidth]{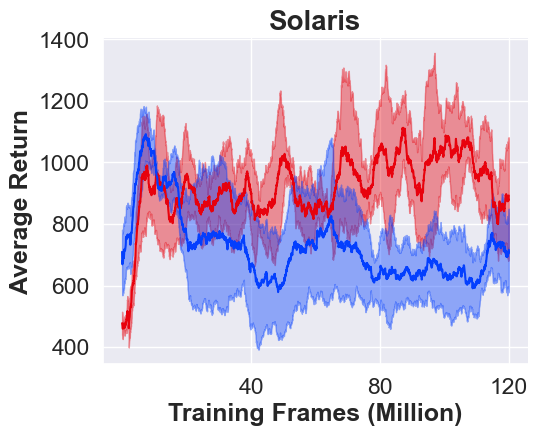} 
\label{fig:4_learning_curves_appendix/SolarisNoFrameskip-v0_4_learning_curves_appendix.png} 
\end{subfigure}%

\begin{subfigure}[t]{ 0.2\textwidth} 
\centering 
\includegraphics[width=\textwidth]{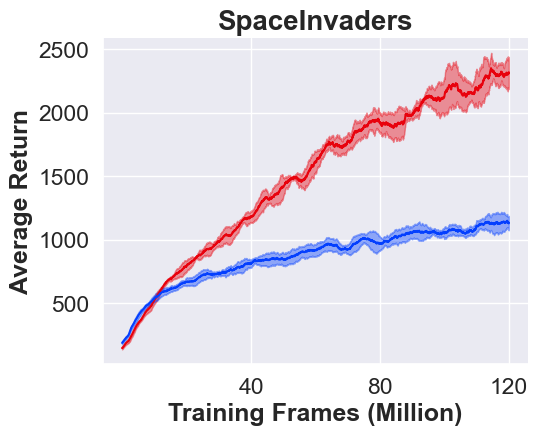} 
\label{fig:4_learning_curves_appendix/SpaceInvadersNoFrameskip-v0_4_learning_curves_appendix.png} 
\end{subfigure}%
~ 
\begin{subfigure}[t]{ 0.2\textwidth} 
\centering 
\includegraphics[width=\textwidth]{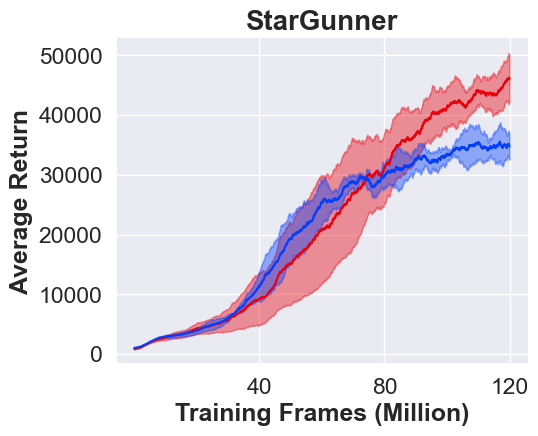} 
\label{fig:4_learning_curves_appendix/StarGunnerNoFrameskip-v0_4_learning_curves_appendix.png} 
\end{subfigure}%
~ 
\begin{subfigure}[t]{ 0.2\textwidth} 
\centering 
\includegraphics[width=\textwidth]{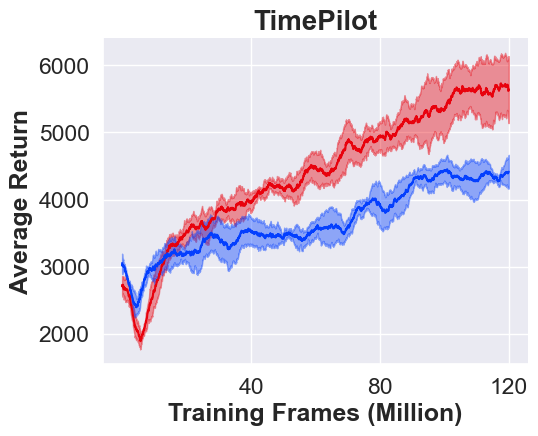} 
\label{fig:4_learning_curves_appendix/TimePilotNoFrameskip-v0_4_learning_curves_appendix.png} 
\end{subfigure}%
~ 
\begin{subfigure}[t]{ 0.2\textwidth} 
\centering 
\includegraphics[width=\textwidth]{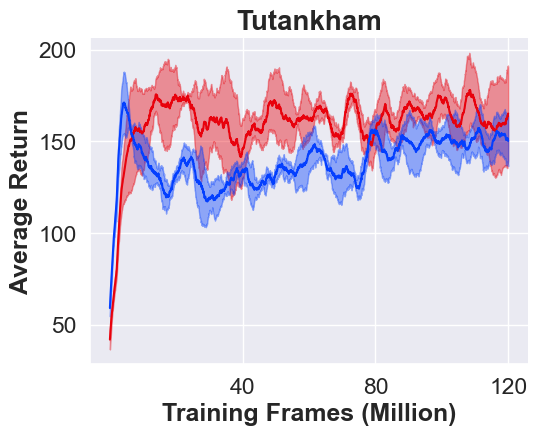} 
\label{fig:4_learning_curves_appendix/TutankhamNoFrameskip-v0_4_learning_curves_appendix.png} 
\end{subfigure}%
~ 
\begin{subfigure}[t]{ 0.2\textwidth} 
\centering 
\includegraphics[width=\textwidth]{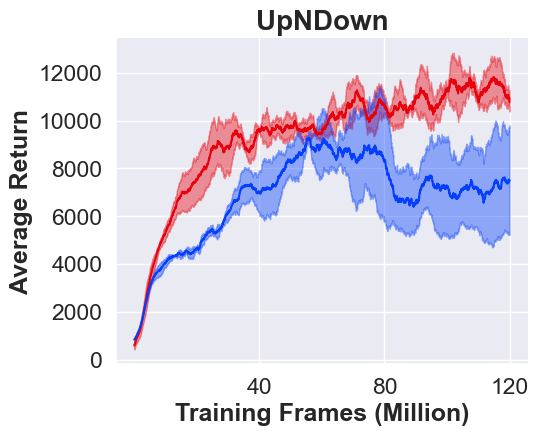} 
\label{fig:4_learning_curves_appendix/UpNDownNoFrameskip-v0_4_learning_curves_appendix.png} 
\end{subfigure}%

\begin{subfigure}[t]{ 0.2\textwidth} 
\centering 
\includegraphics[width=\textwidth]{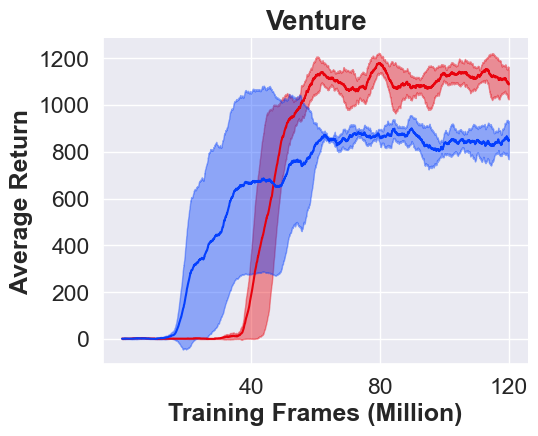} 
\label{fig:4_learning_curves_appendix/VentureNoFrameskip-v0_4_learning_curves_appendix.png} 
\end{subfigure}%
~ 
\begin{subfigure}[t]{ 0.2\textwidth} 
\centering 
\includegraphics[width=\textwidth]{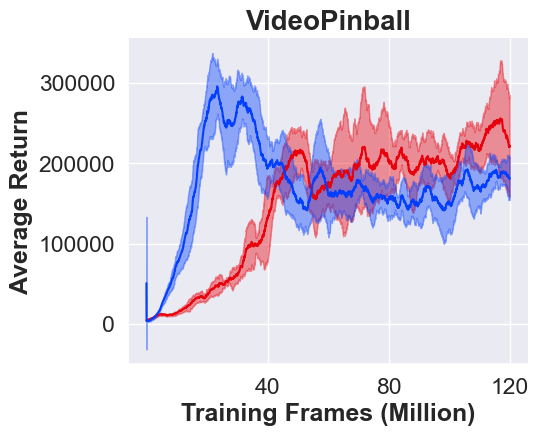} 
\label{fig:4_learning_curves_appendix/VideoPinballNoFrameskip-v0_4_learning_curves_appendix.png} 
\end{subfigure}%
~ 
\begin{subfigure}[t]{ 0.2\textwidth} 
\centering 
\includegraphics[width=\textwidth]{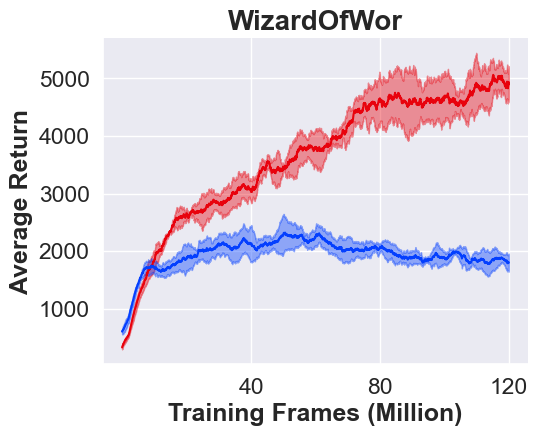} 
\label{fig:4_learning_curves_appendix/WizardOfWorNoFrameskip-v0_4_learning_curves_appendix.png} 
\end{subfigure}%
~ 
\begin{subfigure}[t]{ 0.2\textwidth} 
\centering 
\includegraphics[width=\textwidth]{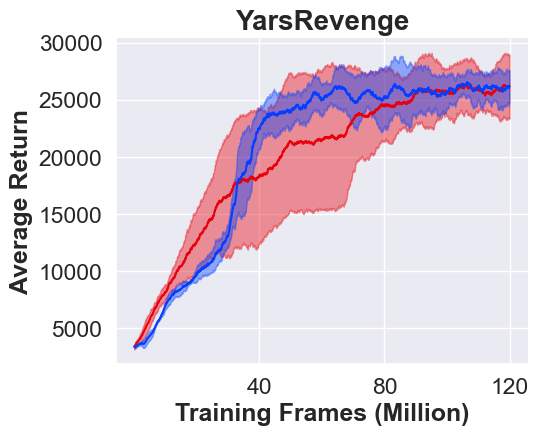} 
\label{fig:4_learning_curves_appendix/YarsRevengeNoFrameskip-v0_4_learning_curves_appendix.png} 
\end{subfigure}%
~ 
\begin{subfigure}[t]{ 0.2\textwidth} 
\centering 
\includegraphics[width=\textwidth]{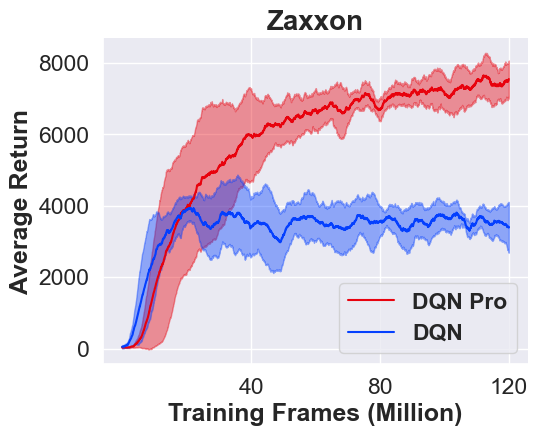} 
\label{fig:4_learning_curves_appendix/ZaxxonNoFrameskip-v0_4_learning_curves_appendix.png} 
\end{subfigure}%

\caption{Comparison between DQN Pro (red) and DQN (blue) over 55 Atari games (Part II).} 
\label{fig:dqn_Pro_vs_dqn_learning_curve} 
\end{figure} 

\begin{figure}
\centering\captionsetup[subfigure]{justification=centering}
\begin{subfigure}[t]{ 0.2\textwidth} 
\centering 
\includegraphics[width=\textwidth]{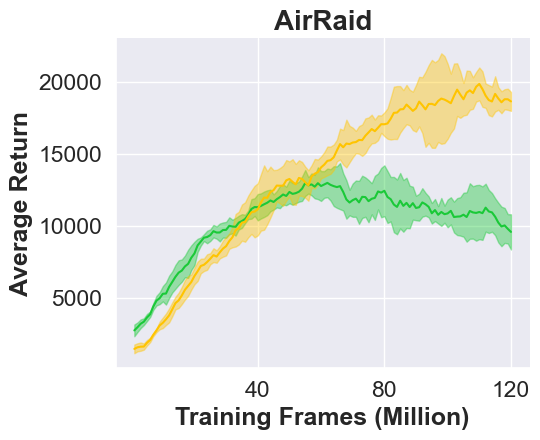} 
\label{fig:4_learning_curves_appendix_rainbow/AirRaid.png} 
\end{subfigure}%
~ 
\begin{subfigure}[t]{ 0.2\textwidth} 
\centering 
\includegraphics[width=\textwidth]{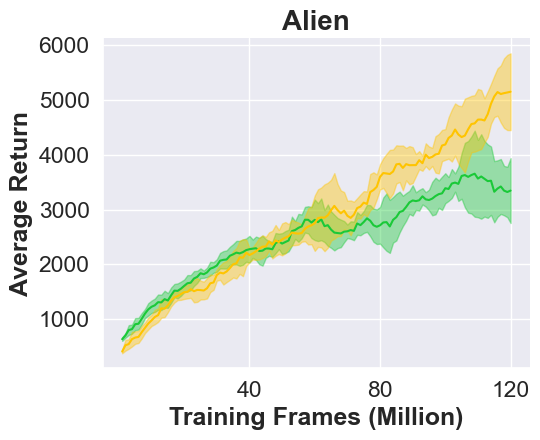} 
\label{fig:4_learning_curves_appendix_rainbow/Alien.png} 
\end{subfigure}%
~ 
\begin{subfigure}[t]{ 0.2\textwidth} 
\centering 
\includegraphics[width=\textwidth]{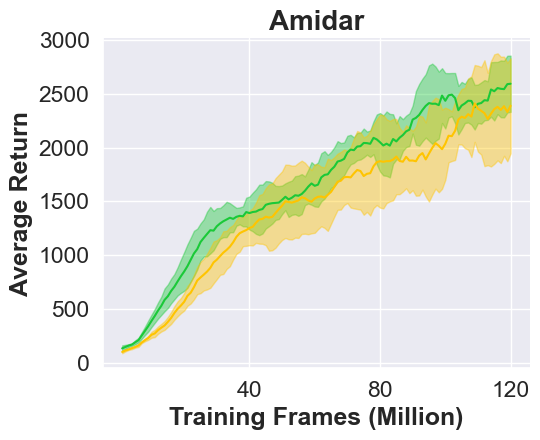} 
\label{fig:4_learning_curves_appendix_rainbow/Amidar.png} 
\end{subfigure}%
~ 
\begin{subfigure}[t]{ 0.2\textwidth} 
\centering 
\includegraphics[width=\textwidth]{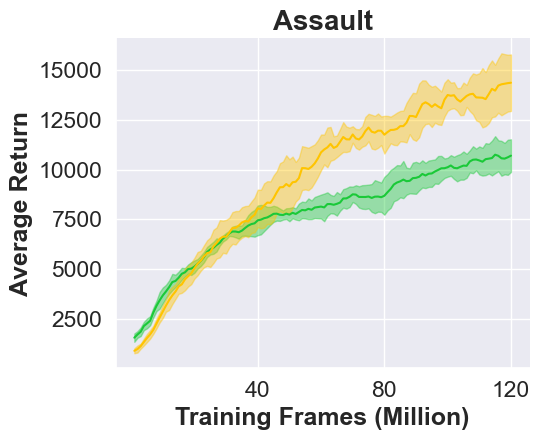} 
\label{fig:4_learning_curves_appendix_rainbow/Assault.png} 
\end{subfigure}%
~ 
\begin{subfigure}[t]{ 0.2\textwidth} 
\centering 
\includegraphics[width=\textwidth]{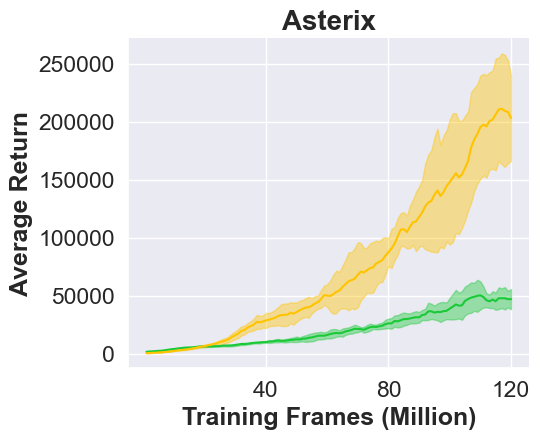} 
\label{fig:4_learning_curves_appendix_rainbow/Asterix.png} 
\end{subfigure}%

\begin{subfigure}[t]{ 0.2\textwidth} 
\centering 
\includegraphics[width=\textwidth]{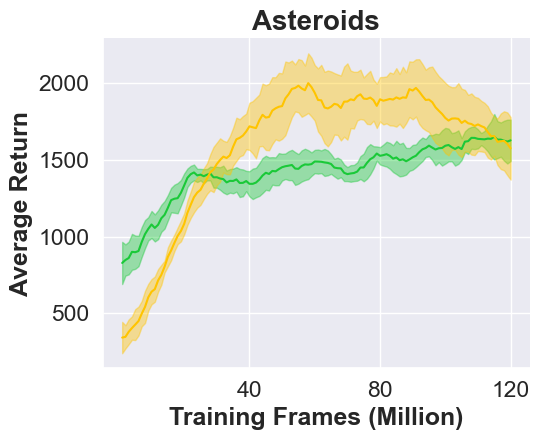} 
\label{fig:4_learning_curves_appendix_rainbow/Asteroids.png} 
\end{subfigure}%
~ 
\begin{subfigure}[t]{ 0.2\textwidth} 
\centering 
\includegraphics[width=\textwidth]{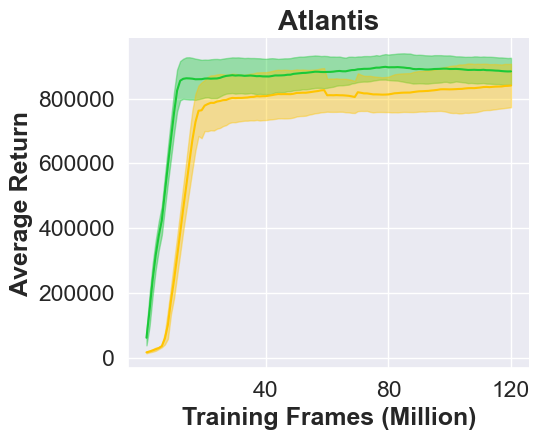} 
\label{fig:4_learning_curves_appendix_rainbow/Atlantis.png} 
\end{subfigure}%
~ 
\begin{subfigure}[t]{ 0.2\textwidth} 
\centering 
\includegraphics[width=\textwidth]{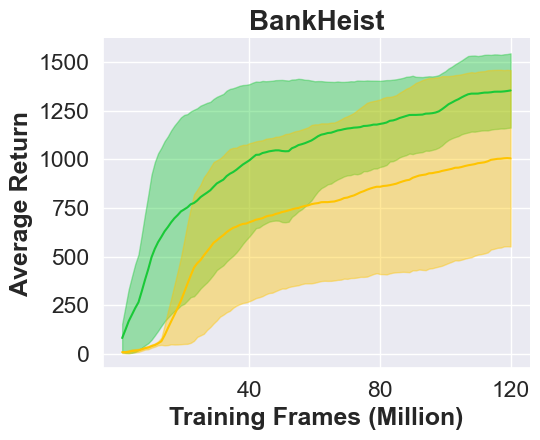} 
\label{fig:4_learning_curves_appendix_rainbow/BankHeist.png} 
\end{subfigure}%
~ 
\begin{subfigure}[t]{ 0.2\textwidth} 
\centering 
\includegraphics[width=\textwidth]{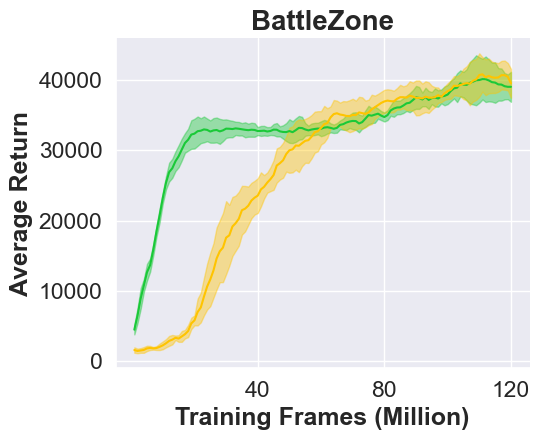} 
\label{fig:4_learning_curves_appendix_rainbow/BattleZone.png} 
\end{subfigure}%
~ 
\begin{subfigure}[t]{ 0.2\textwidth} 
\centering 
\includegraphics[width=\textwidth]{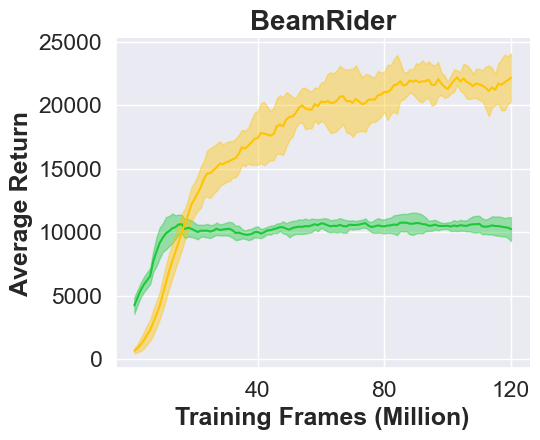} 
\label{fig:4_learning_curves_appendix_rainbow/BeamRider.png} 
\end{subfigure}%

\begin{subfigure}[t]{ 0.2\textwidth} 
\centering 
\includegraphics[width=\textwidth]{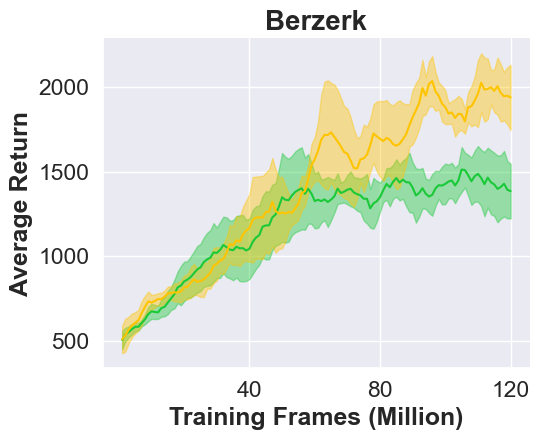} 
\label{fig:4_learning_curves_appendix_rainbow/Berzerk.png} 
\end{subfigure}%
~ 
\begin{subfigure}[t]{ 0.2\textwidth} 
\centering 
\includegraphics[width=\textwidth]{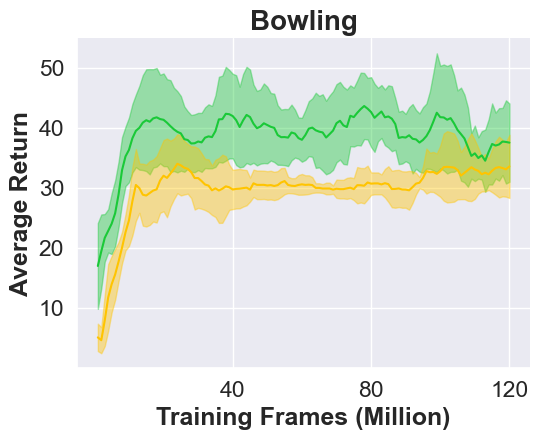} 
\label{fig:4_learning_curves_appendix_rainbow/Bowling.png} 
\end{subfigure}%
~ 
\begin{subfigure}[t]{ 0.2\textwidth} 
\centering 
\includegraphics[width=\textwidth]{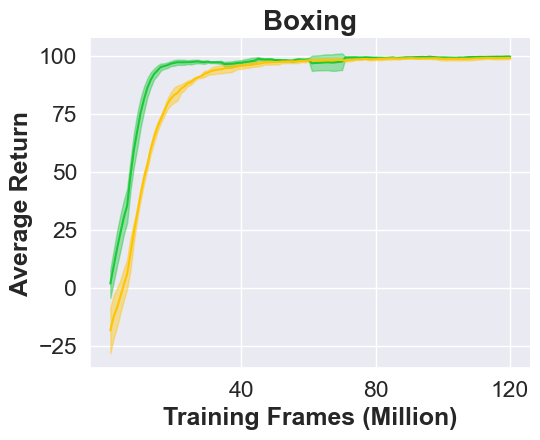} 
\label{fig:4_learning_curves_appendix_rainbow/Boxing.png} 
\end{subfigure}%
~ 
\begin{subfigure}[t]{ 0.2\textwidth} 
\centering 
\includegraphics[width=\textwidth]{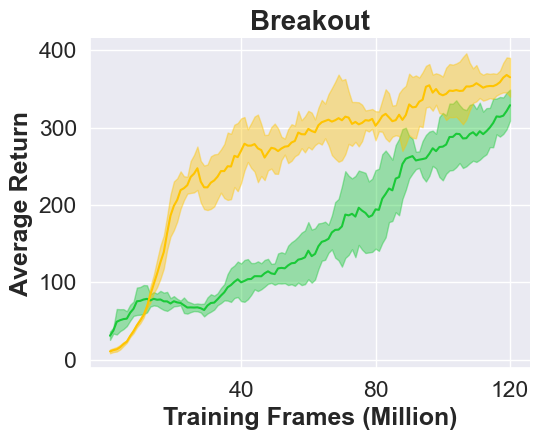} 
\label{fig:4_learning_curves_appendix_rainbow/Breakout.png} 
\end{subfigure}%
~ 
\begin{subfigure}[t]{ 0.2\textwidth} 
\centering 
\includegraphics[width=\textwidth]{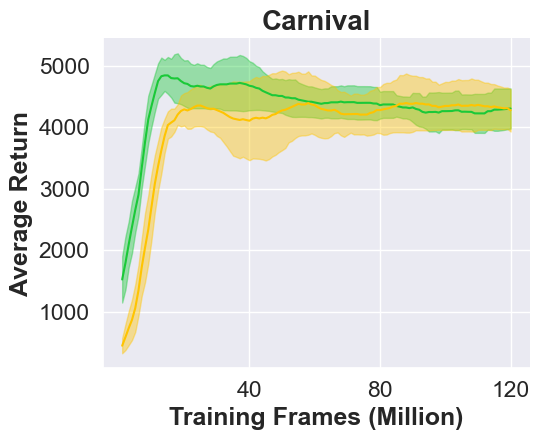} 
\label{fig:4_learning_curves_appendix_rainbow/Carnival.png} 
\end{subfigure}%

\begin{subfigure}[t]{ 0.2\textwidth} 
\centering 
\includegraphics[width=\textwidth]{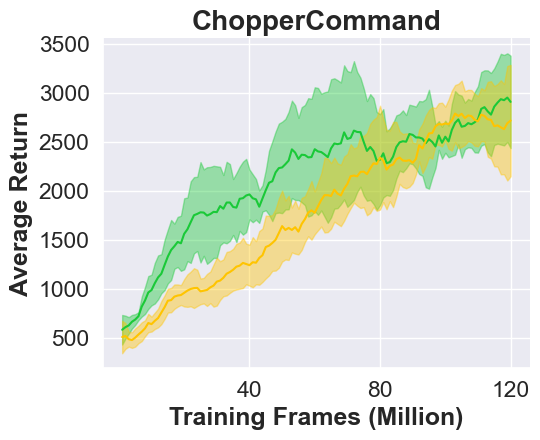} 
\label{fig:4_learning_curves_appendix_rainbow/ChopperCommand.png} 
\end{subfigure}%
~ 
\begin{subfigure}[t]{ 0.2\textwidth} 
\centering 
\includegraphics[width=\textwidth]{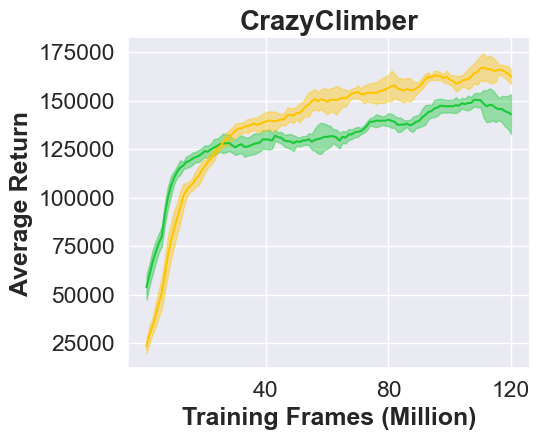} 
\label{fig:4_learning_curves_appendix_rainbow/CrazyClimber.png} 
\end{subfigure}%
~ 
\begin{subfigure}[t]{ 0.2\textwidth} 
\centering 
\includegraphics[width=\textwidth]{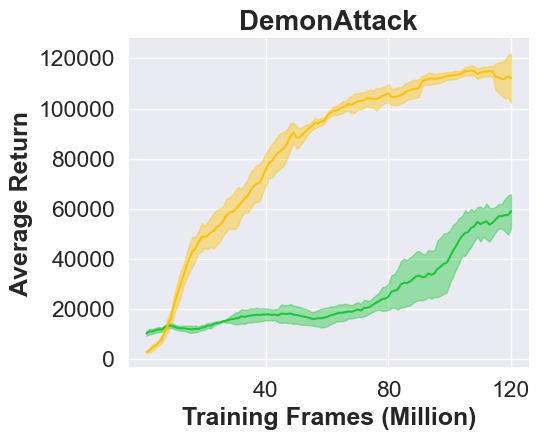} 
\label{fig:4_learning_curves_appendix_rainbow/DemonAttack.png} 
\end{subfigure}%
~ 
\begin{subfigure}[t]{ 0.2\textwidth} 
\centering 
\includegraphics[width=\textwidth]{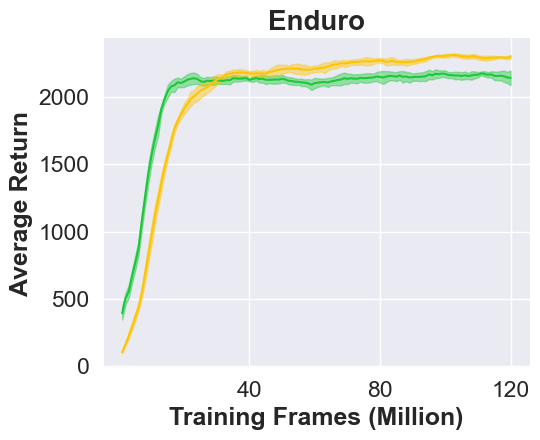} 
\label{fig:4_learning_curves_appendix_rainbow/Enduro.png} 
\end{subfigure}%
~ 
\begin{subfigure}[t]{ 0.2\textwidth} 
\centering 
\includegraphics[width=\textwidth]{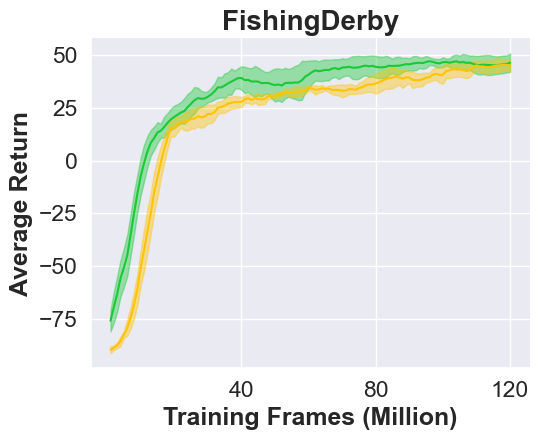} 
\label{fig:4_learning_curves_appendix_rainbow/FishingDerby.png} 
\end{subfigure}%

\begin{subfigure}[t]{ 0.2\textwidth} 
\centering 
\includegraphics[width=\textwidth]{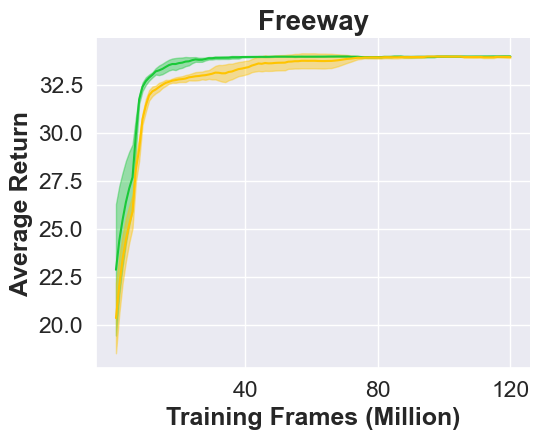} 
\label{fig:4_learning_curves_appendix_rainbow/Freeway.png} 
\end{subfigure}%
~ 
\begin{subfigure}[t]{ 0.2\textwidth} 
\centering 
\includegraphics[width=\textwidth]{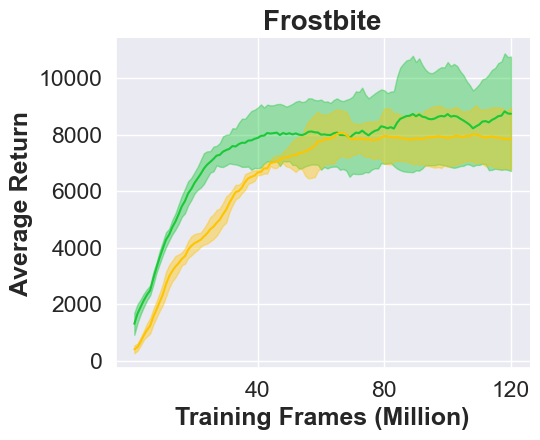} 
\label{fig:4_learning_curves_appendix_rainbow/Frostbite.png} 
\end{subfigure}%
~ 
\begin{subfigure}[t]{ 0.2\textwidth} 
\centering 
\includegraphics[width=\textwidth]{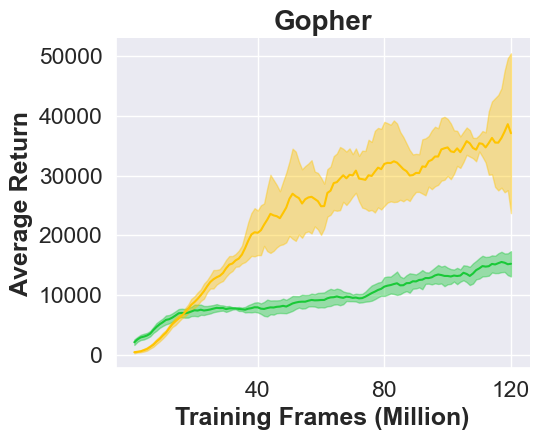} 
\label{fig:4_learning_curves_appendix_rainbow/Gopher.png} 
\end{subfigure}%
~ 
\begin{subfigure}[t]{ 0.2\textwidth} 
\centering 
\includegraphics[width=\textwidth]{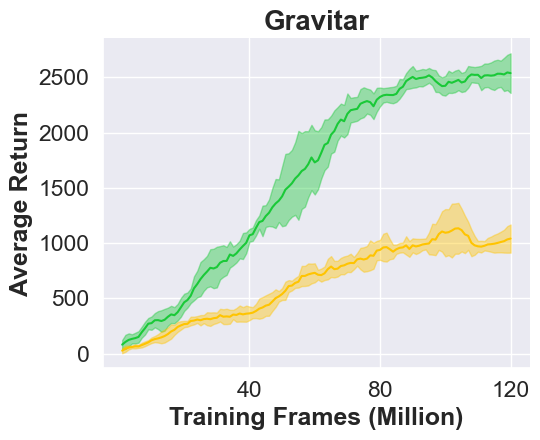} 
\label{fig:4_learning_curves_appendix_rainbow/Gravitar.png} 
\end{subfigure}%
~ 
\begin{subfigure}[t]{ 0.2\textwidth} 
\centering 
\includegraphics[width=\textwidth]{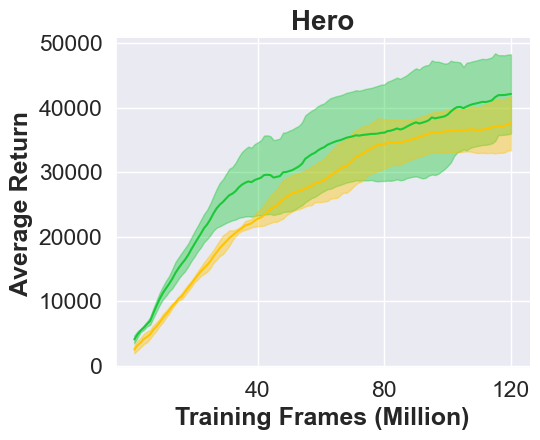} 
\label{fig:4_learning_curves_appendix_rainbow/Hero.png} 
\end{subfigure}%

\begin{subfigure}[t]{ 0.2\textwidth} 
\centering 
\includegraphics[width=\textwidth]{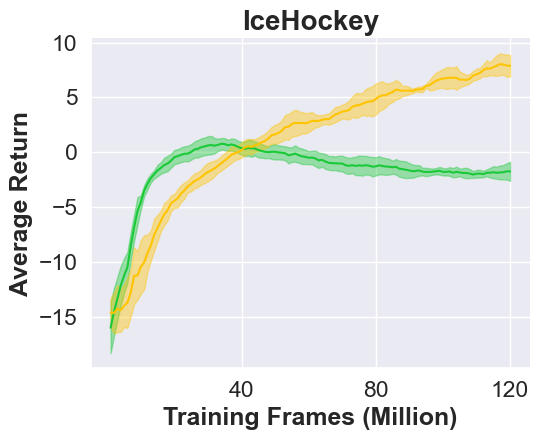} 
\label{fig:4_learning_curves_appendix_rainbow/IceHockey.png} 
\end{subfigure}%
~ 
\begin{subfigure}[t]{ 0.2\textwidth} 
\centering 
\includegraphics[width=\textwidth]{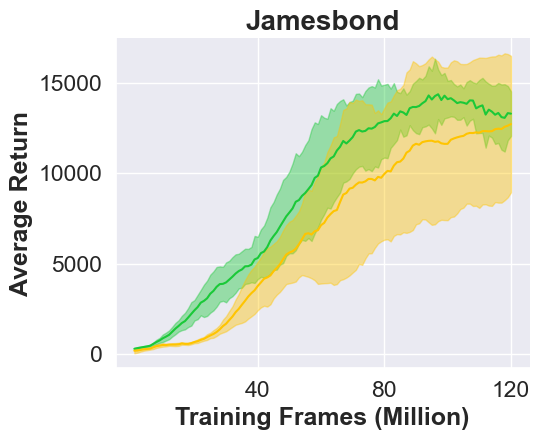} 
\label{fig:4_learning_curves_appendix_rainbow/Jamesbond.png} 
\end{subfigure}%
~ 
\begin{subfigure}[t]{ 0.2\textwidth} 
\centering 
\includegraphics[width=\textwidth]{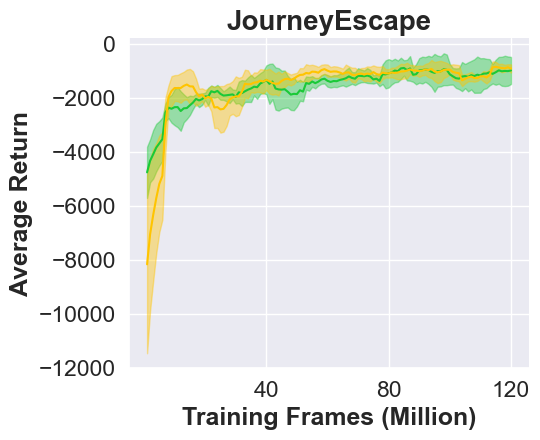} 
\label{fig:4_learning_curves_appendix_rainbow/JourneyEscape.png} 
\end{subfigure}%
~ 
\begin{subfigure}[t]{ 0.2\textwidth} 
\centering 
\includegraphics[width=\textwidth]{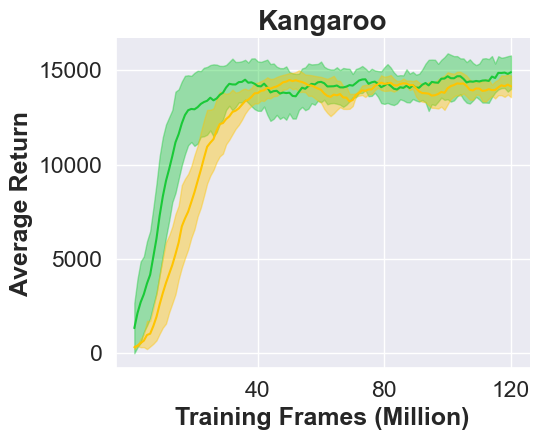} 
\label{fig:4_learning_curves_appendix_rainbow/Kangaroo.png} 
\end{subfigure}%
~ 
\begin{subfigure}[t]{ 0.2\textwidth} 
\centering 
\includegraphics[width=\textwidth]{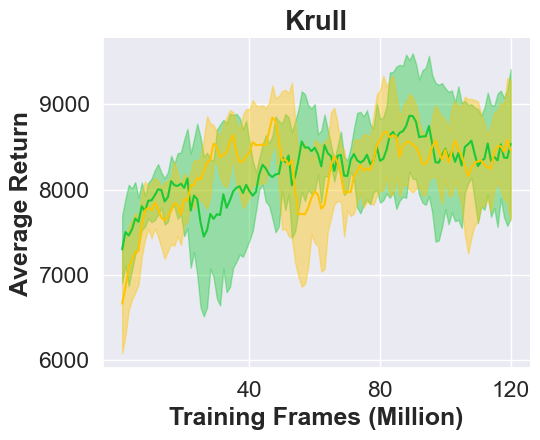} 
\label{fig:4_learning_curves_appendix_rainbow/Krull.png} 
\end{subfigure}%
\caption{Comparison between Rainbow Pro (yellow) and Rainbow (green) over 55 Atari games (Part I).} 
\end{figure}
\begin{figure}
\begin{subfigure}[t]{ 0.2\textwidth} 
\centering 
\includegraphics[width=\textwidth]{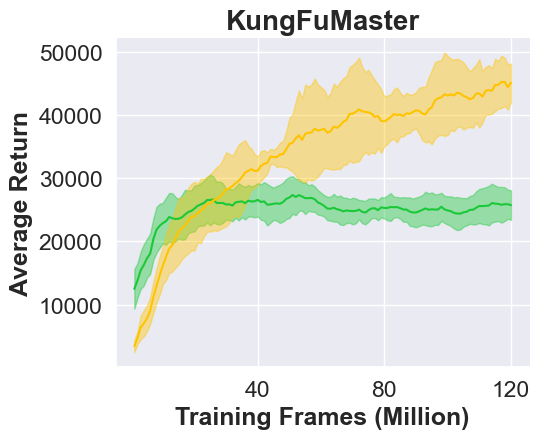} 
\label{fig:4_learning_curves_appendix_rainbow/KungFuMaster.png} 
\end{subfigure}%
~ 
\begin{subfigure}[t]{ 0.2\textwidth} 
\centering 
\includegraphics[width=\textwidth]{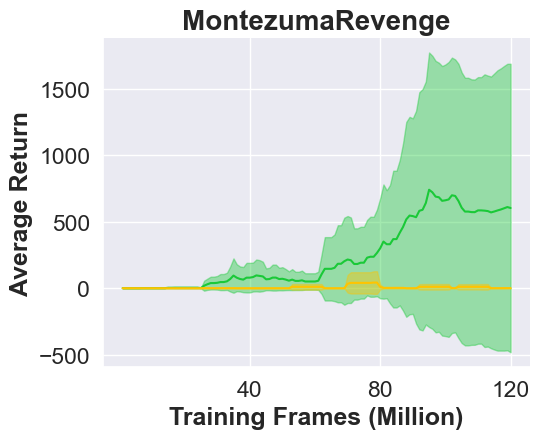} 
\label{fig:4_learning_curves_appendix_rainbow/MontezumaRevenge.png} 
\end{subfigure}%
~ 
\begin{subfigure}[t]{ 0.2\textwidth} 
\centering 
\includegraphics[width=\textwidth]{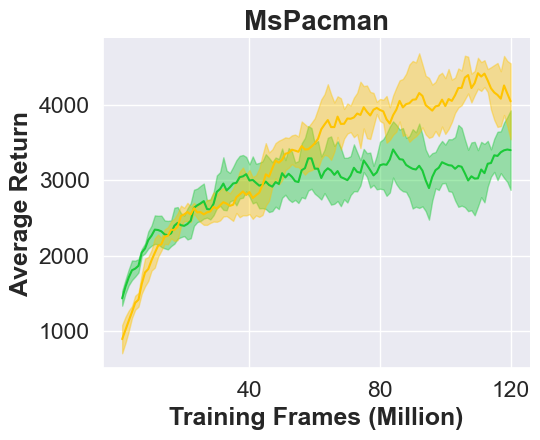} 
\label{fig:4_learning_curves_appendix_rainbow/MsPacman.png} 
\end{subfigure}%
~ 
\begin{subfigure}[t]{ 0.2\textwidth} 
\centering 
\includegraphics[width=\textwidth]{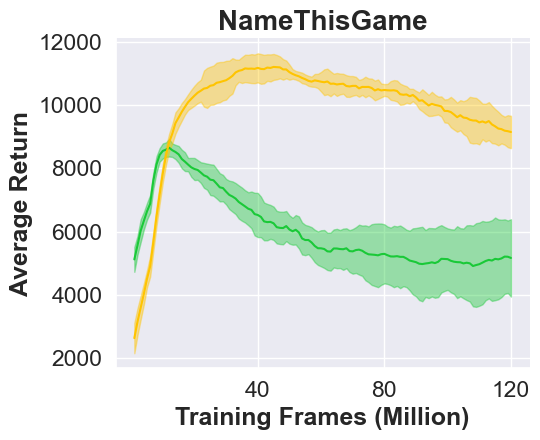} 
\label{fig:4_learning_curves_appendix_rainbow/NameThisGame.png} 
\end{subfigure}%
~ 
\begin{subfigure}[t]{ 0.2\textwidth} 
\centering 
\includegraphics[width=\textwidth]{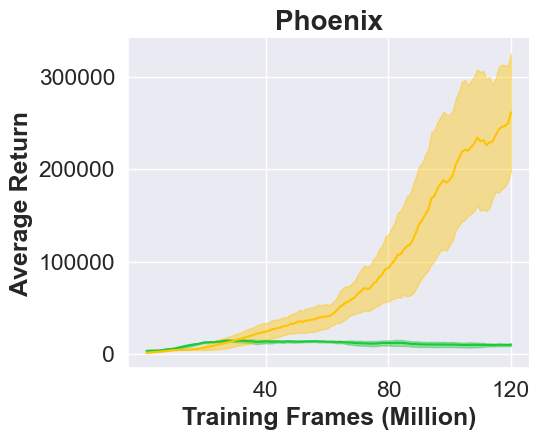} 
\label{fig:4_learning_curves_appendix_rainbow/Phoenix.png} 
\end{subfigure}%

\begin{subfigure}[t]{ 0.2\textwidth} 
\centering 
\includegraphics[width=\textwidth]{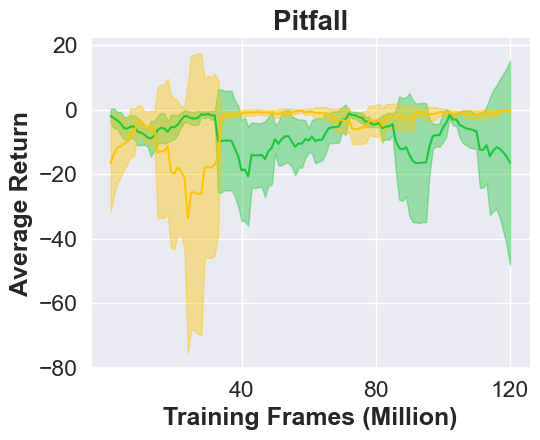} 
\label{fig:4_learning_curves_appendix_rainbow/Pitfall.png} 
\end{subfigure}%
~ 
\begin{subfigure}[t]{ 0.2\textwidth} 
\centering 
\includegraphics[width=\textwidth]{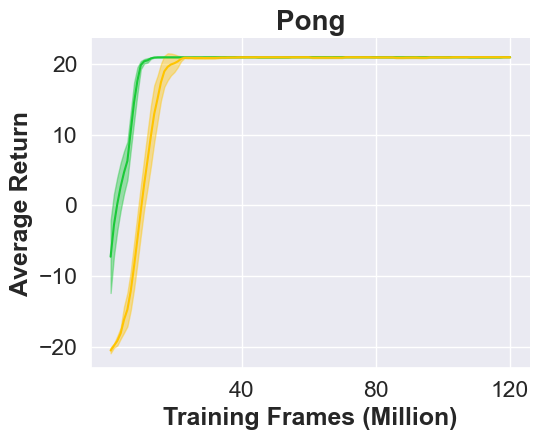} 
\label{fig:4_learning_curves_appendix_rainbow/Pong.png} 
\end{subfigure}%
~ 
\begin{subfigure}[t]{ 0.2\textwidth} 
\centering 
\includegraphics[width=\textwidth]{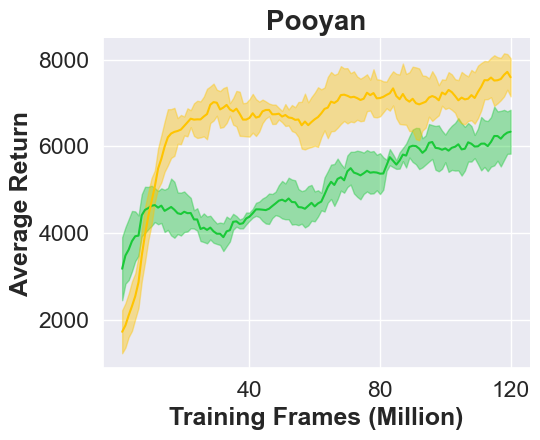} 
\label{fig:4_learning_curves_appendix_rainbow/Pooyan.png} 
\end{subfigure}%
~ 
\begin{subfigure}[t]{ 0.2\textwidth} 
\centering 
\includegraphics[width=\textwidth]{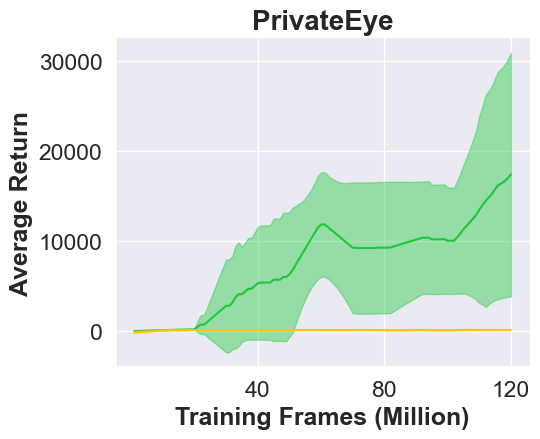} 
\label{fig:4_learning_curves_appendix_rainbow/PrivateEye.png} 
\end{subfigure}%
~ 
\begin{subfigure}[t]{ 0.2\textwidth} 
\centering 
\includegraphics[width=\textwidth]{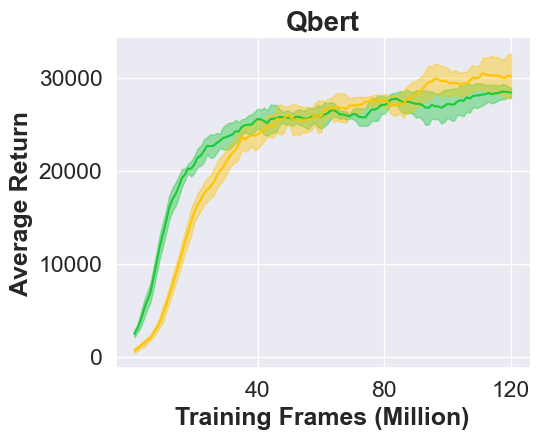} 
\label{fig:4_learning_curves_appendix_rainbow/Qbert.png} 
\end{subfigure}%

\begin{subfigure}[t]{ 0.2\textwidth} 
\centering 
\includegraphics[width=\textwidth]{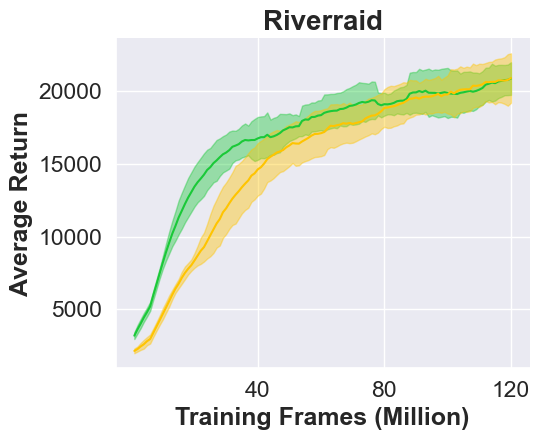} 
\label{fig:4_learning_curves_appendix_rainbow/Riverraid.png} 
\end{subfigure}%
~ 
\begin{subfigure}[t]{ 0.2\textwidth} 
\centering 
\includegraphics[width=\textwidth]{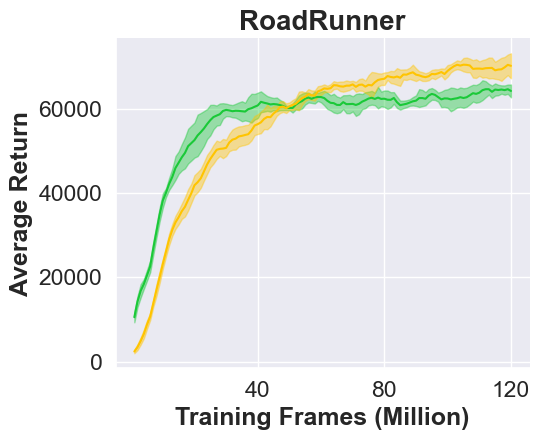} 
\label{fig:4_learning_curves_appendix_rainbow/RoadRunner.png} 
\end{subfigure}%
~ 
\begin{subfigure}[t]{ 0.2\textwidth} 
\centering 
\includegraphics[width=\textwidth]{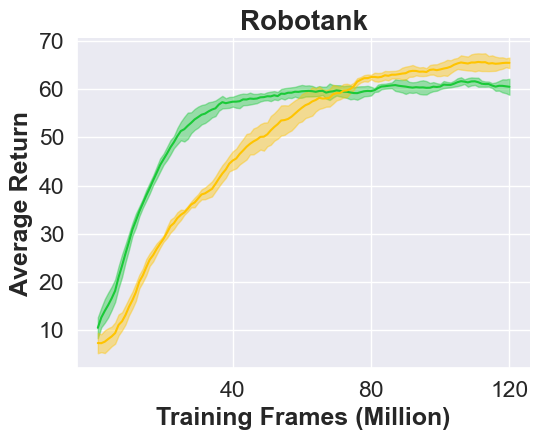} 
\label{fig:4_learning_curves_appendix_rainbow/Robotank.png} 
\end{subfigure}%
~ 
\begin{subfigure}[t]{ 0.2\textwidth} 
\centering 
\includegraphics[width=\textwidth]{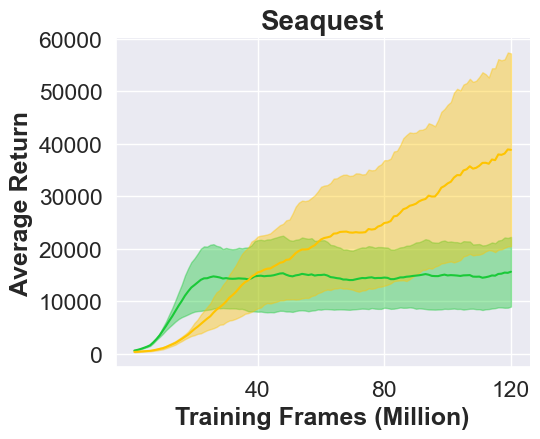} 
\label{fig:4_learning_curves_appendix_rainbow/Seaquest.png} 
\end{subfigure}%
~ 
\begin{subfigure}[t]{ 0.2\textwidth} 
\centering 
\includegraphics[width=\textwidth]{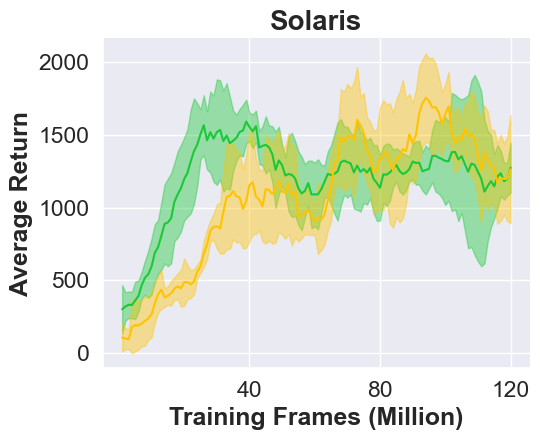} 
\label{fig:4_learning_curves_appendix_rainbow/Solaris.png} 
\end{subfigure}%

\begin{subfigure}[t]{ 0.2\textwidth} 
\centering 
\includegraphics[width=\textwidth]{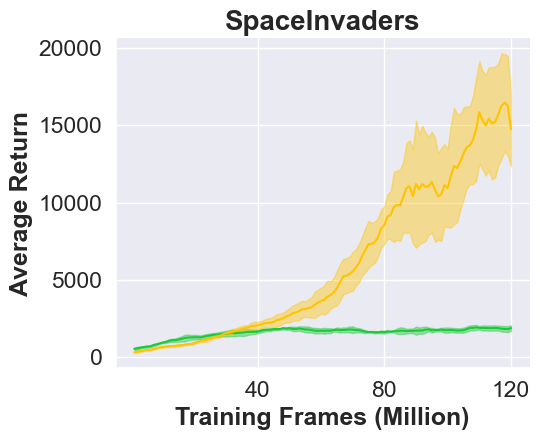} 
\label{fig:4_learning_curves_appendix_rainbow/SpaceInvaders.png} 
\end{subfigure}%
~ 
\begin{subfigure}[t]{ 0.2\textwidth} 
\centering 
\includegraphics[width=\textwidth]{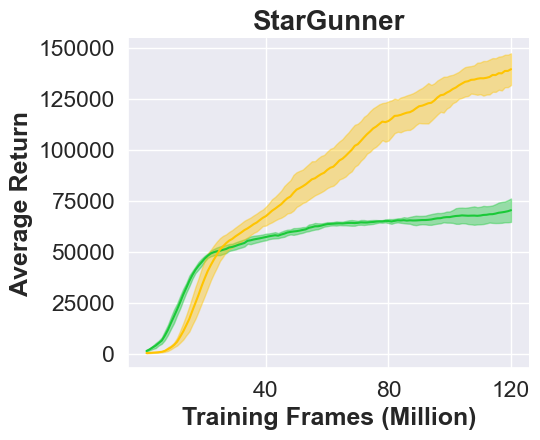} 
\label{fig:4_learning_curves_appendix_rainbow/StarGunner.png} 
\end{subfigure}%
~ 
\begin{subfigure}[t]{ 0.2\textwidth} 
\centering 
\includegraphics[width=\textwidth]{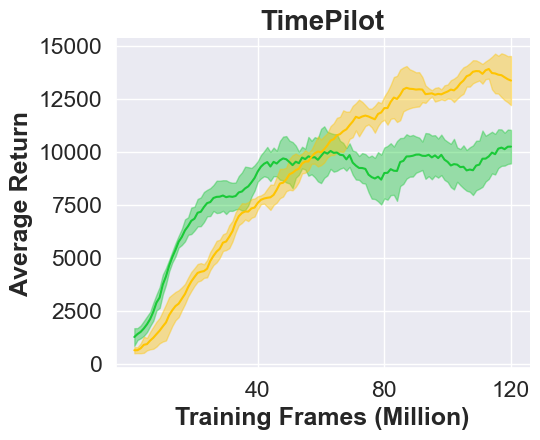} 
\label{fig:4_learning_curves_appendix_rainbow/TimePilot.png} 
\end{subfigure}%
~ 
\begin{subfigure}[t]{ 0.2\textwidth} 
\centering 
\includegraphics[width=\textwidth]{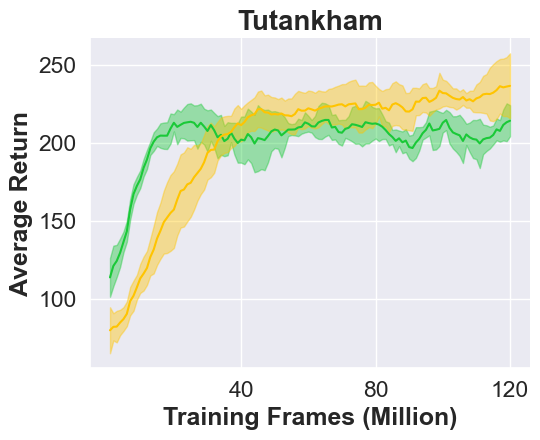} 
\label{fig:4_learning_curves_appendix_rainbow/Tutankham.png} 
\end{subfigure}%
~ 
\begin{subfigure}[t]{ 0.2\textwidth} 
\centering 
\includegraphics[width=\textwidth]{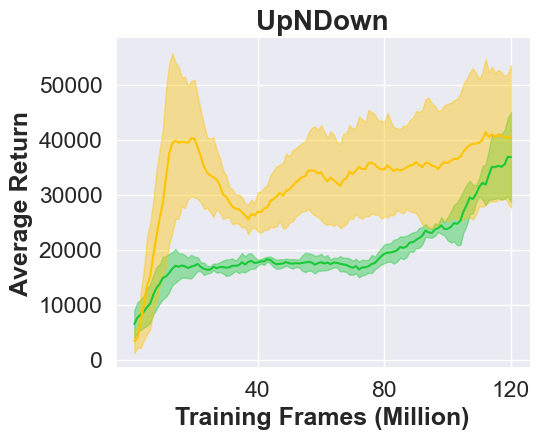} 
\label{fig:4_learning_curves_appendix_rainbow/UpNDown.png} 
\end{subfigure}%

\begin{subfigure}[t]{ 0.2\textwidth} 
\centering 
\includegraphics[width=\textwidth]{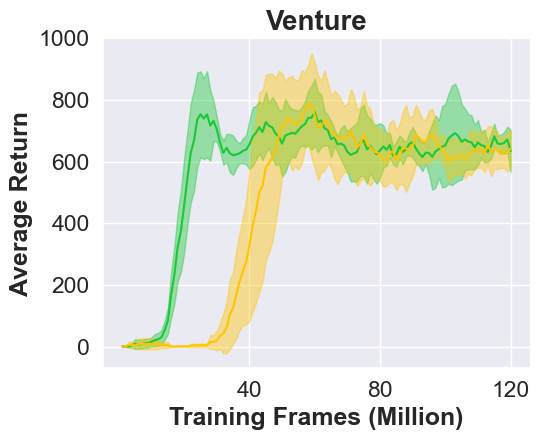} 
\label{fig:4_learning_curves_appendix_rainbow/Venture.png} 
\end{subfigure}%
~ 
\begin{subfigure}[t]{ 0.2\textwidth} 
\centering 
\includegraphics[width=\textwidth]{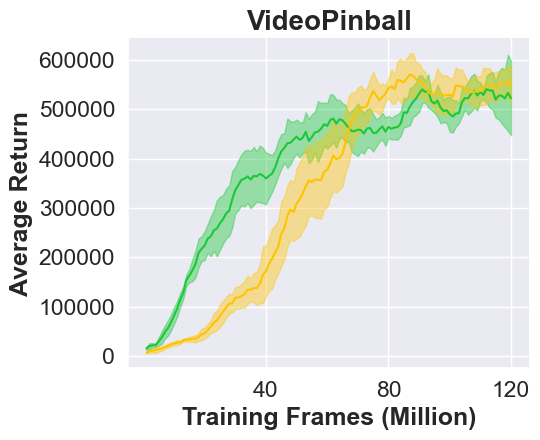} 
\label{fig:4_learning_curves_appendix_rainbow/VideoPinball.png} 
\end{subfigure}%
~ 
\begin{subfigure}[t]{ 0.2\textwidth} 
\centering 
\includegraphics[width=\textwidth]{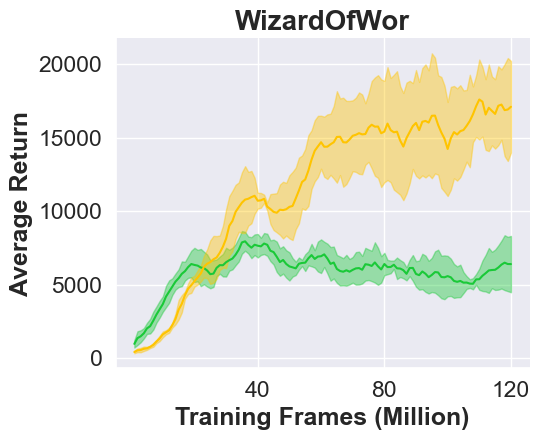} 
\label{fig:4_learning_curves_appendix_rainbow/WizardOfWor.png} 
\end{subfigure}%
~ 
\begin{subfigure}[t]{ 0.2\textwidth} 
\centering 
\includegraphics[width=\textwidth]{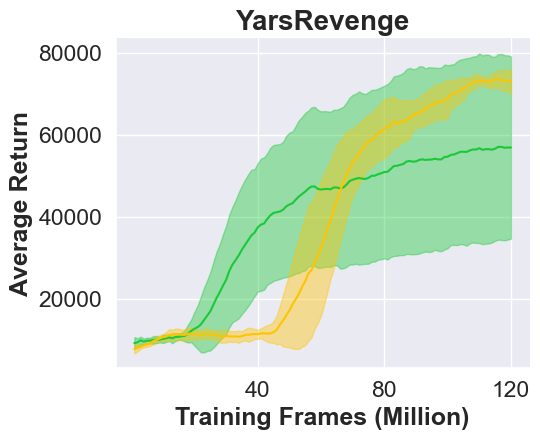} 
\label{fig:4_learning_curves_appendix_rainbow/YarsRevenge.png} 
\end{subfigure}%
~ 
\begin{subfigure}[t]{ 0.2\textwidth} 
\centering 
\includegraphics[width=\textwidth]{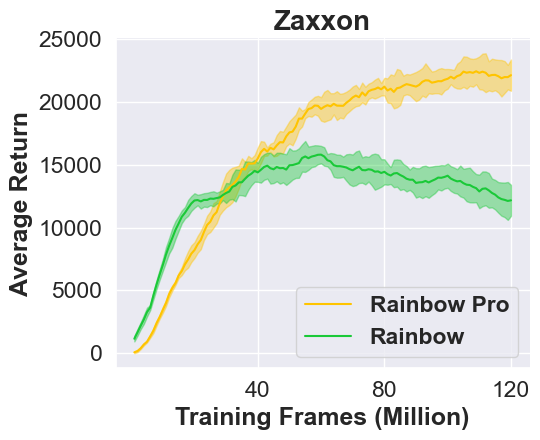} 
\label{fig:4_learning_curves_appendix_rainbow/Zaxxon.png} 
\end{subfigure}%

\caption{Comparison between Rainbow Pro (yellow) and Rainbow (green) over 55 Atari games (Part II).} 
\label{fig:rainbow_Pro_vs_rainbow_learning_curve} 
\end{figure}

\clearpage
\newpage

\section{Motivating Example for Adaptive Updates}
In this section we specifically look at 4 domains in which Rainbow Pro did significantly better than the original Rainbow, as well 4 domains where Rainbow Pro is underperforming Rainbow. Note, again, that it is uncommon for Rainbow Pro with the original $\tilde c$ to underperform, but here we have a deeper dive into these cases for a better understanding.

\begin{figure}
\centering\captionsetup[subfigure]{justification=centering,skip=0pt}
\begin{subfigure}[t]{0.245\textwidth} 
\centering 
\includegraphics[width=\textwidth]{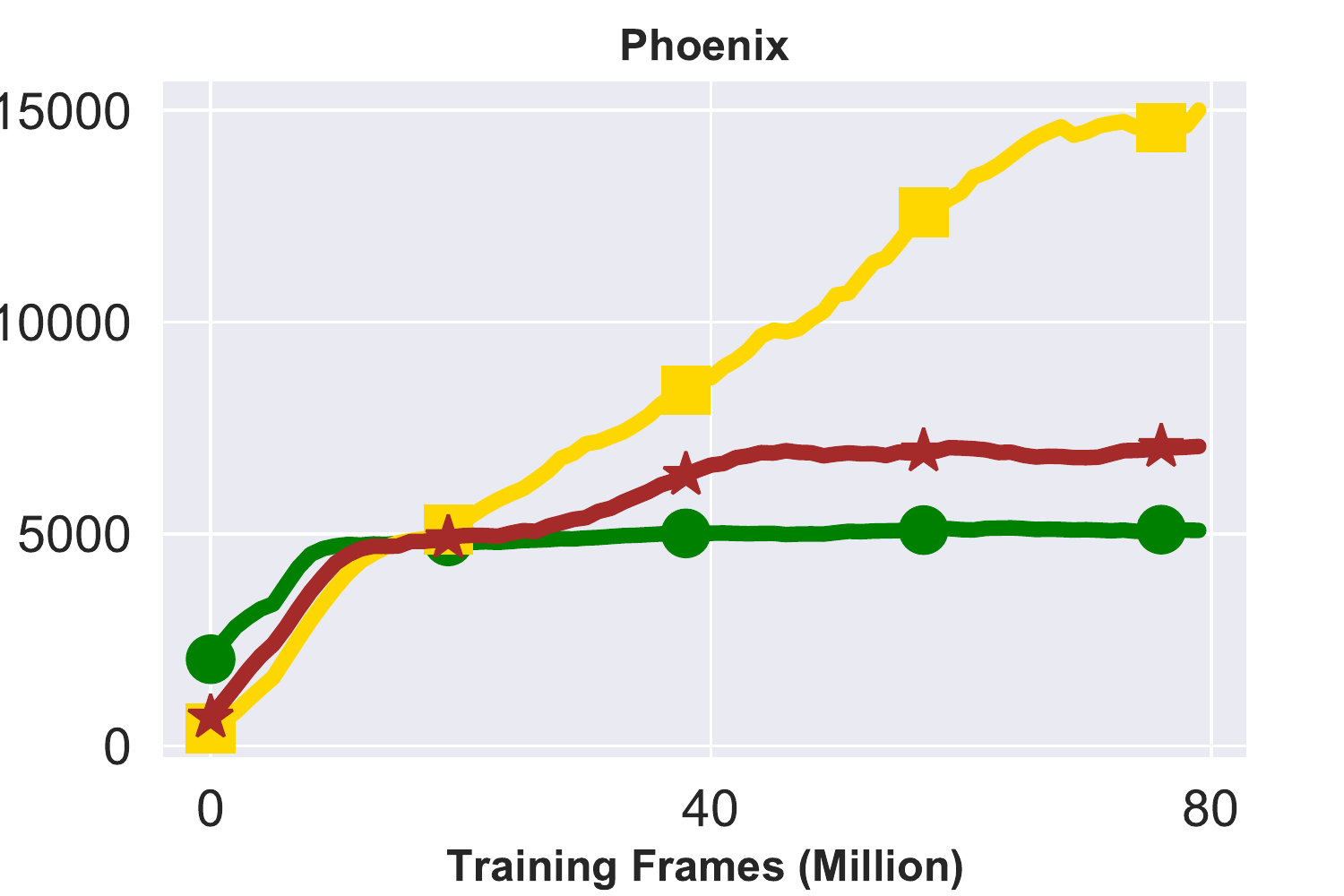} 
\end{subfigure}%
~ 
\begin{subfigure}[t]{ 0.245\textwidth} 
\centering 
\includegraphics[width=\textwidth]{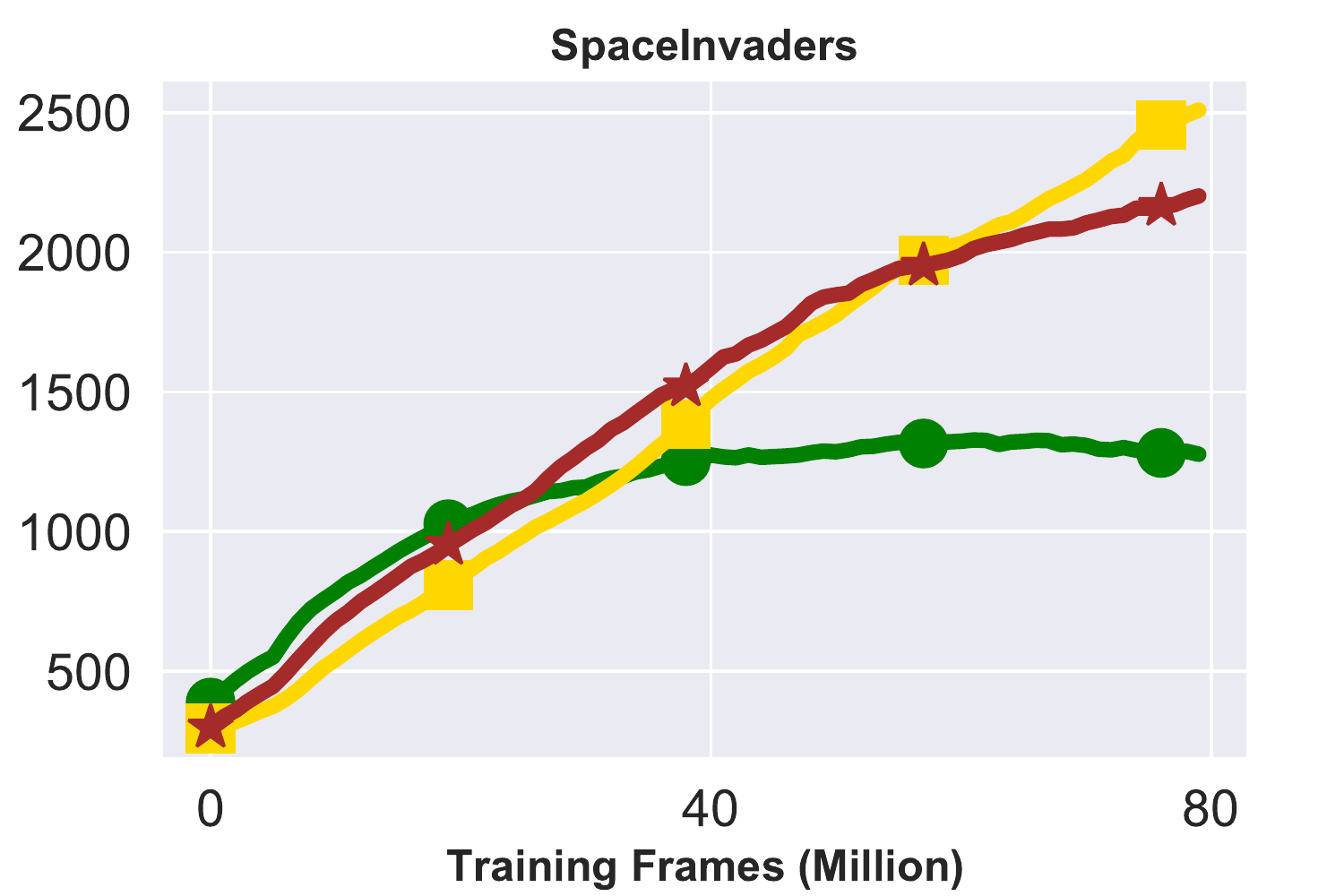} 
\end{subfigure}%
~ 
\begin{subfigure}[t]{ 0.245\textwidth} 
\centering 
\includegraphics[width=\textwidth]{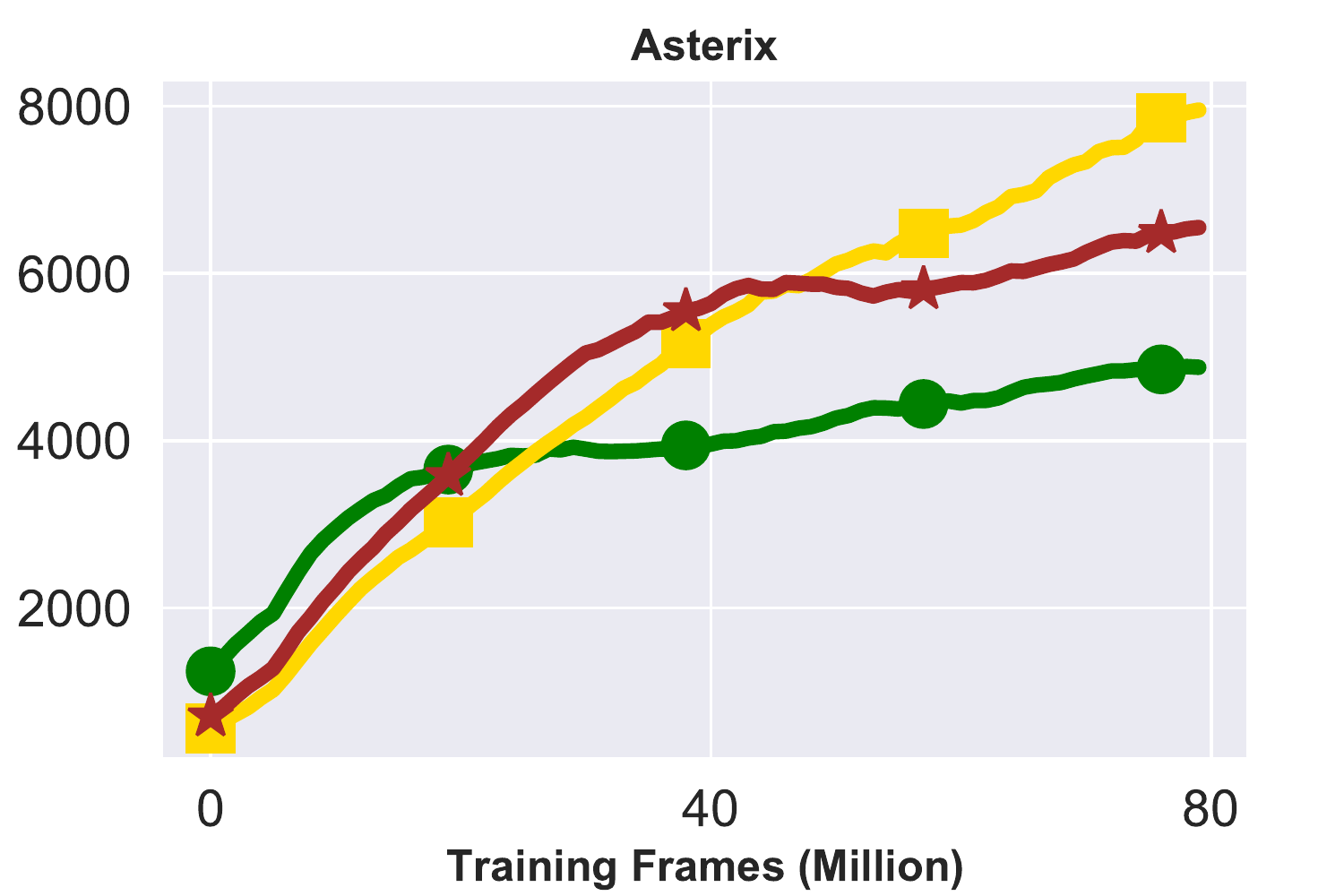} 
\end{subfigure}%
~ 
\begin{subfigure}[t]{ 0.245\textwidth} 
\centering 
\includegraphics[width=\textwidth]{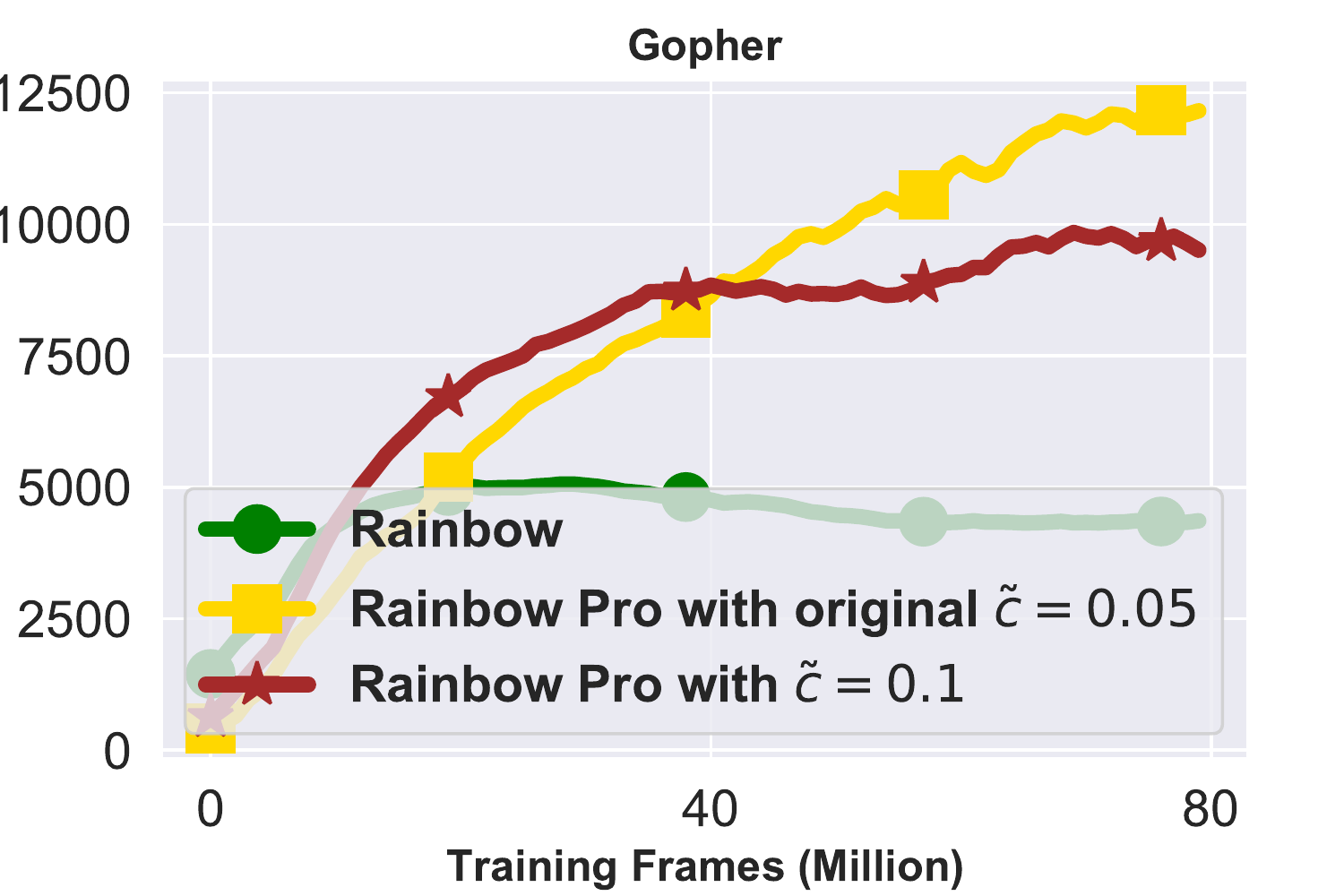} 
\end{subfigure}

\begin{subfigure}[t]{0.245\textwidth} 
\centering 
\includegraphics[width=\textwidth]{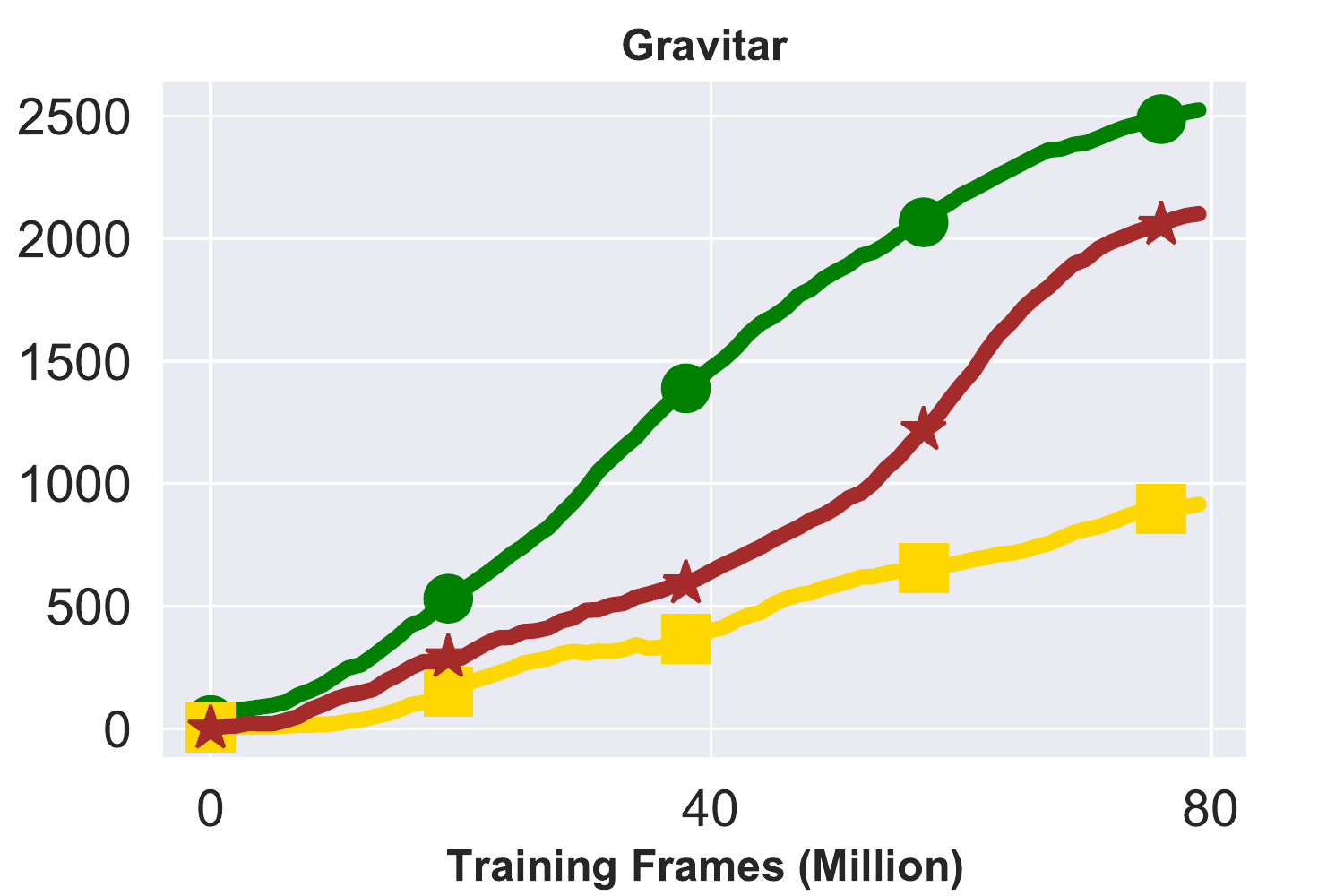} 
\end{subfigure}%
~ 
\begin{subfigure}[t]{ 0.245\textwidth} 
\centering 
\includegraphics[width=\textwidth]{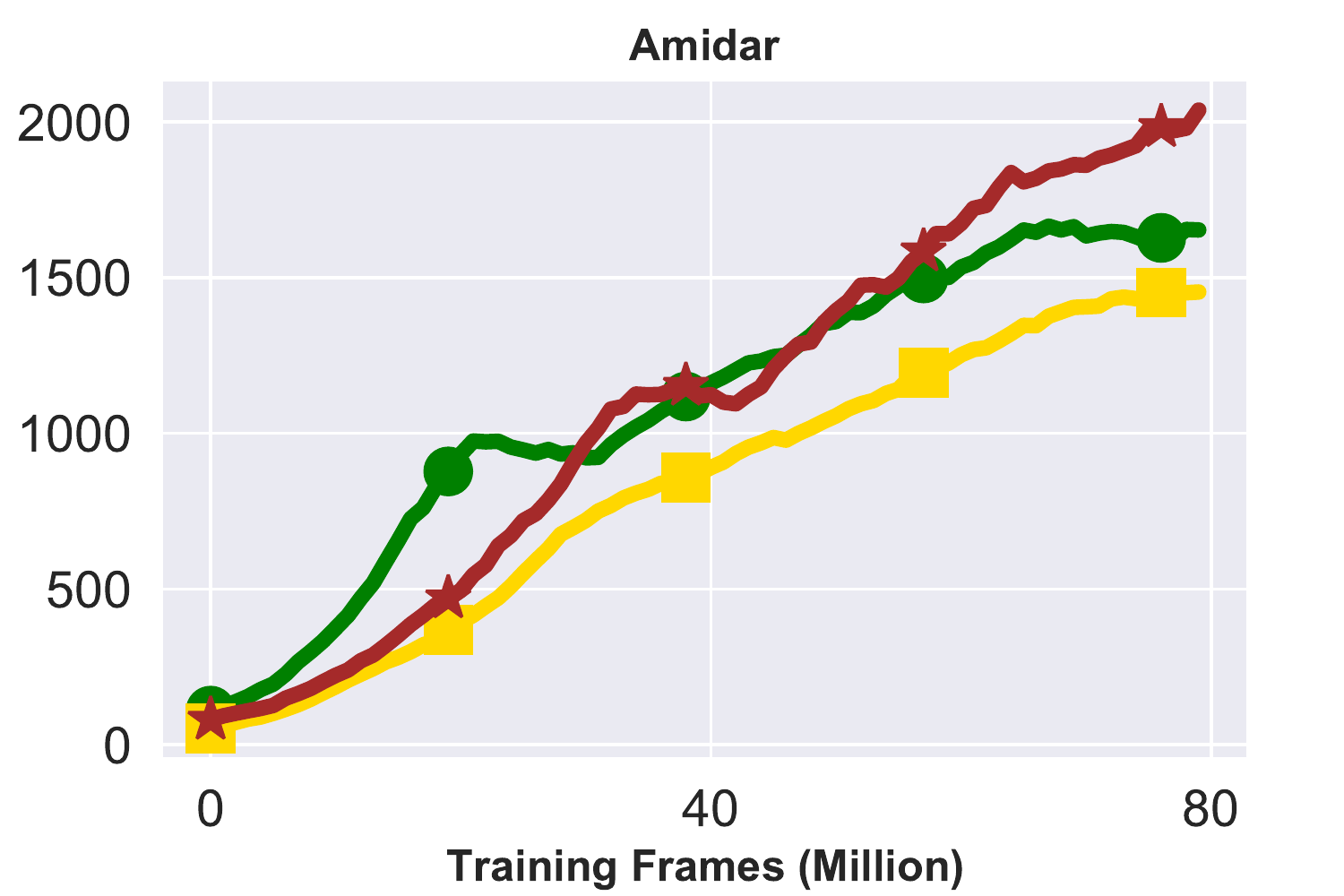} 
\end{subfigure}%
~ 
\begin{subfigure}[t]{ 0.245\textwidth} 
\centering 
\includegraphics[width=\textwidth]{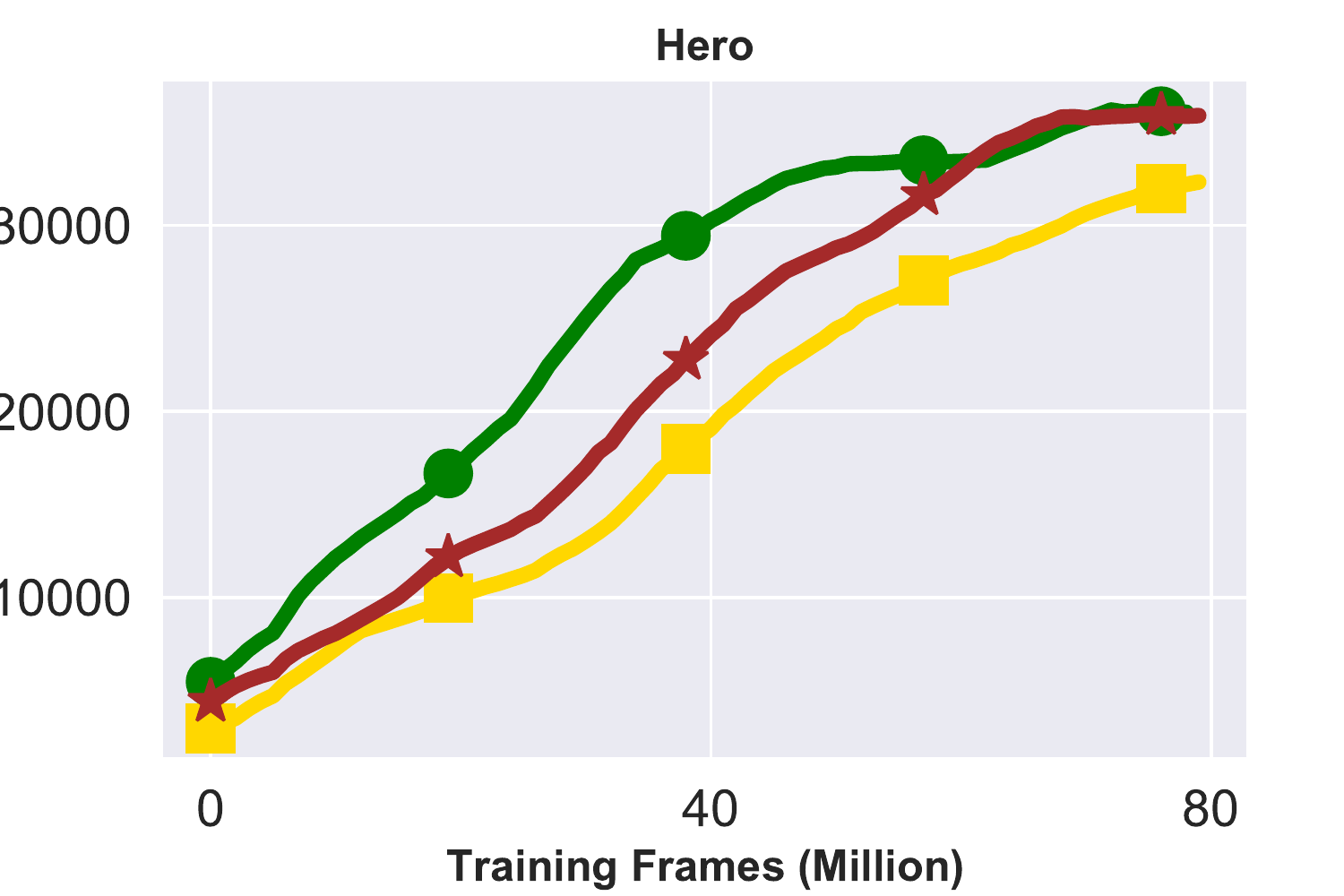} 
\end{subfigure}%
~ 
\begin{subfigure}[t]{ 0.245\textwidth} 
\centering 
\includegraphics[width=\textwidth]{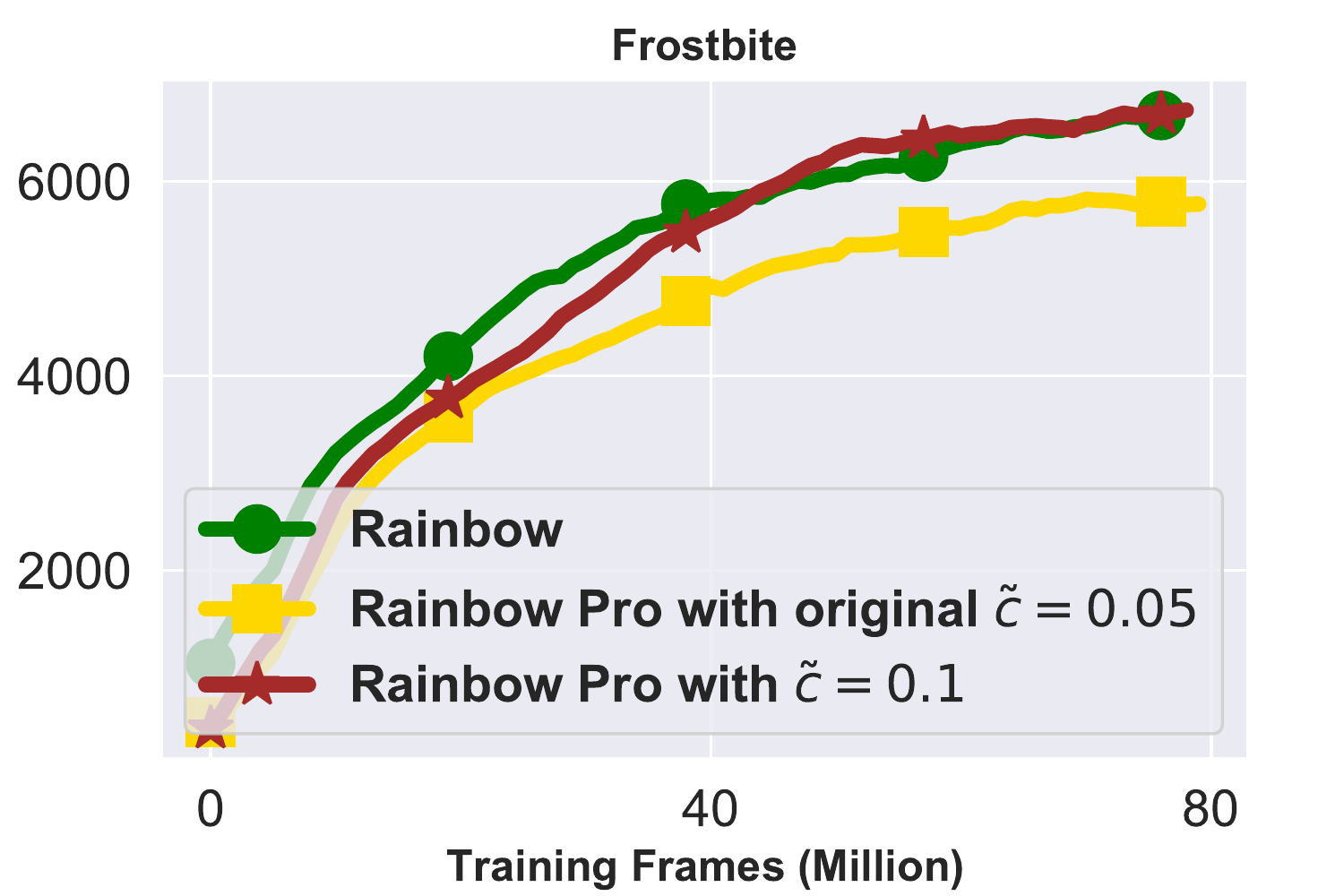} 
\end{subfigure}%
\caption{A study on games with the strongest (top) and weakest performance for Rainbow Pro with the original $\tilde c=0.05$. Using a slightly less powerful proximal term (corresponding to larger $\tilde c=0.1$) enables us to recover the downside (bottom) while still providing benefits on games that are more conducive to using the proximal updates (top).} 
\label{fig:11_adaptive} 
\end{figure}

From Figure~\ref{fig:11_adaptive}, we observe that by using a slightly larger value of $\tilde c$, which slightly decreases the incentive for online-target proximity, we can recover from the downside, while still maintaining superior performance on games that are conducive to proximal updates. This suggest that, while using a fixed $\tilde c$ value is enough to obtain significant performance improvement, adaptively choosing $\tilde c$ would provide us with even more reliable improvements when performing proximal updates. In this context, a promising idea would be to hinge on the variance of our gradient updates when setting $\tilde c$. We leave this promising direction for future work.

\clearpage
\newpage
\section{Proximal Updates in the Value Space}
Our primary contribution was to show the usefulness of performing proximal updates in the parameter space. That said, we also implemented a version of proximal updates that operated in the value space. More specifically, in this case we updated the parameters of the online network as follows: 
\begin{equation*}
 w\leftarrow\empiricalExpectation{\langle s,a,r,s'\rangle}\Big[\big(r+\gamma \max_{a'} \widehat Q(s',a';\theta)- \widehat Q(s,a;w)\big)^2\Big] + \frac{1}{\tilde c} \empiricalExpectation{\langle s,a\rangle}\Big[\big(\widehat Q(s,a;w) - \widehat Q(s,a;\theta)\big)^2]. 
\end{equation*}
We conducted numerous experiments using variants of this idea (such as using separate replay buffer for each term, performing the update for all actions in buffered states, etc) but we generally found the value-space updates to be ineffective. As mentioned in the main paper, we believe this is because the parameter-space definition can enforce the proximity globally, while in the value space one can only hope to obtain proximity locally and on a batch of samples. To perform global updates we may need to compute the natural gradient, which typically requires matrix invasion~\cite{knight2018natural}, and thus adding significant computational burden to the original algorithm. In comparison, the parameter-space version is effective, simple to implement, and capable of enforcing proximity globally due to the Lipschitz property of neural networks.
\begin{figure}
\centering\captionsetup[subfigure]{justification=centering}
\begin{subfigure}[t]{ 0.24\textwidth} 
\centering 
\includegraphics[width=\textwidth]{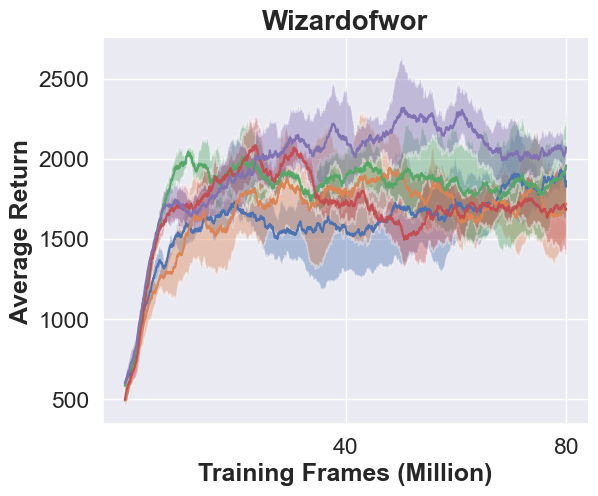} 
\label{fig:WizardOfWorNoFrameskip-v0_learning_curves_9_domains.png} 
\end{subfigure}%
~ 
\begin{subfigure}[t]{ 0.24\textwidth} 
\centering 
\includegraphics[width=\textwidth]{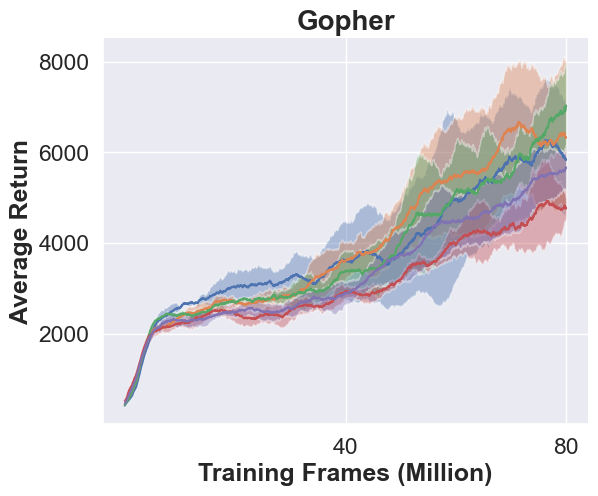} 
\label{fig:GopherNoFrameskip-v0_learning_curves_9_domains.png} 
\end{subfigure}%
~ 
\begin{subfigure}[t]{ 0.24\textwidth} 
\centering 
\includegraphics[width=\textwidth]{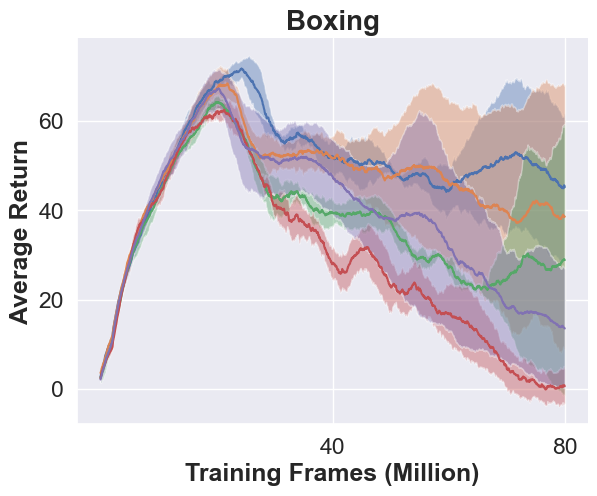} 
\label{fig:BoxingNoFrameskip-v0_learning_curves_9_domains.png} 
\end{subfigure}%
~ 
\begin{subfigure}[t]{ 0.24\textwidth} 
\centering 
\includegraphics[width=\textwidth]{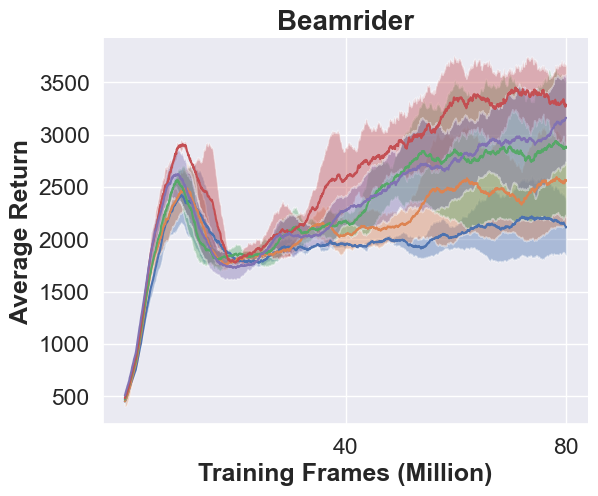} 
\label{fig:BeamRiderNoFrameskip-v0_learning_curves_9_domains.png} 
\end{subfigure}%

\begin{subfigure}[t]{ 0.24\textwidth} 
\centering 
\includegraphics[width=\textwidth]{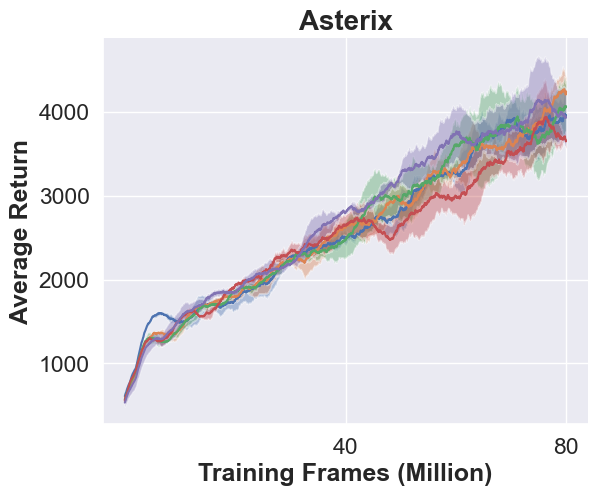} 
\label{fig:AsterixNoFrameskip-v0_learning_curves_9_domains.png} 
\end{subfigure}%
~ 
\begin{subfigure}[t]{ 0.24\textwidth} 
\centering 
\includegraphics[width=\textwidth]{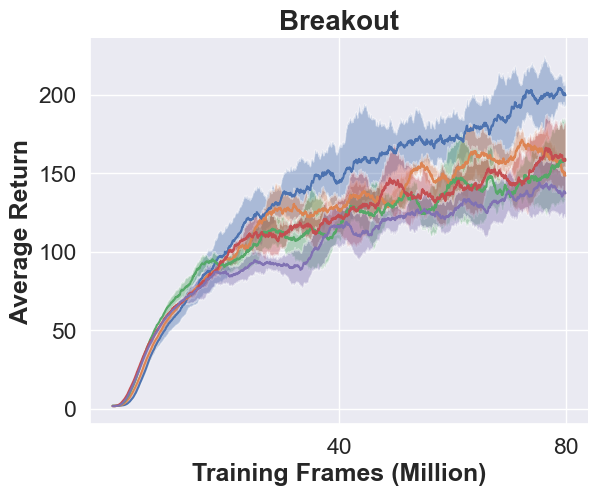} 
\label{fig:BreakoutNoFrameskip-v0_learning_curves_9_domains.png} 
\end{subfigure}%
~ 
\begin{subfigure}[t]{ 0.24\textwidth} 
\centering 
\includegraphics[width=\textwidth]{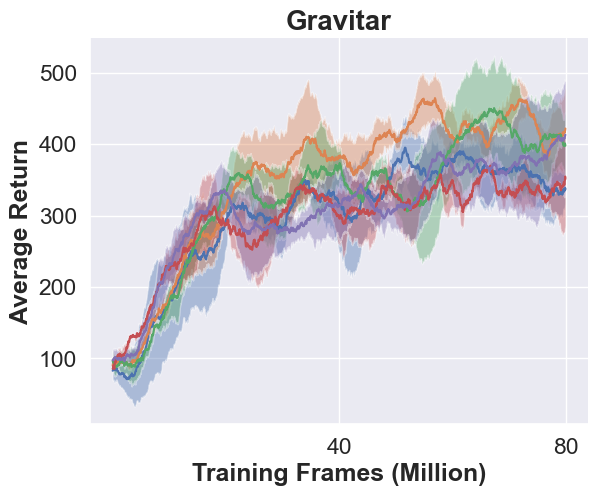} 
\label{fig:GravitarNoFrameskip-v0_learning_curves_9_domains.png} 
\end{subfigure}%
~ 
\begin{subfigure}[t]{ 0.24\textwidth} 
\centering 
\includegraphics[width=\textwidth]{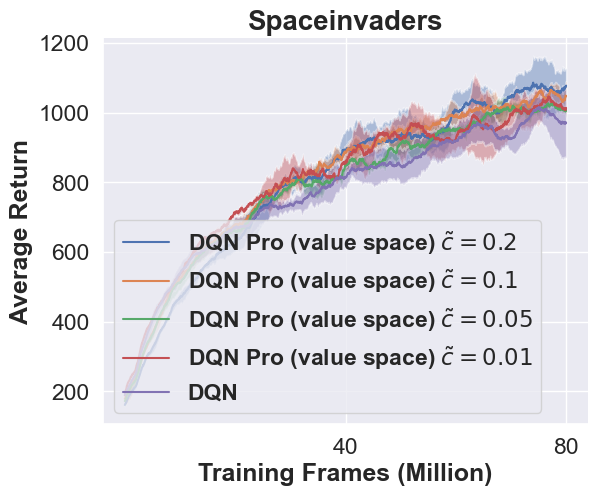} 
\label{fig:SpaceInvadersNoFrameskip-v0_learning_curves_9_domains.png} 
\end{subfigure}%

\caption{Performing proximal updates in the value space has a limited positive impact.} 
\label{fig:value_space} 
\end{figure}

\end{document}